\documentclass[letterpaper,twocolumn,10pt]{article}
\usepackage{sty/usenix-2020-09}

\usepackage{tikz}
\usepackage{amsmath}

\newcount\draft\draft=1 
\usepackage{hyperref}
\usepackage{sty/xxx} 
\usepackage{threeparttable}
\usepackage{multirow}
\usepackage{tabularx}
\usepackage{makecell}
\usepackage{subfigure}
\usepackage{booktabs} 
\usepackage{float} 
\usepackage{rotating} 
\usepackage{enumitem}
\usepackage{pifont} 
\usepackage{color}

\graphicspath{{./figures/}}

\usepackage{url}
\def\UrlBreaks{\do\/\do-\do\_}
\usepackage{breakurl}

\begin{document}

\date{}

\title{\Large \bf Rolling Colors: Adversarial Laser Exploits against Traffic Light Recognition}

\author{
{\rm Chen Yan$^1$, Zhijian Xu$^{1,2}$\thanks{This research was conducted when the author was at Zhejiang University.}, Zhanyuan Yin$^3$, Xiaoyu Ji$^1$\thanks{Xiaoyu Ji is the corresponding author.}, Wenyuan Xu$^1$}\\
$^1$Zhejiang University, $^2$The Chinese University of Hong Kong, $^3$The University of Chicago
} 

\setlist[itemize]{leftmargin=8pt,topsep=0pt,itemsep=.5ex,parsep=0pt,partopsep=0pt}
\setlist[enumerate]{leftmargin=14pt,topsep=0pt,itemsep=.5ex,parsep=0pt,partopsep=0pt}

\maketitle
\pagenumbering{gobble}


\begin{abstract}
Traffic light recognition is essential for fully autonomous driving in urban areas.
In this paper, we investigate the feasibility of fooling traffic light recognition mechanisms by shedding laser interference on the camera.
By exploiting the rolling shutter of CMOS sensors, we manage to inject a color stripe overlapped on the traffic light in the image, which can cause a red light to be recognized as a green light or vice versa. To increase the success rate, we design an optimization method to search for effective laser parameters based on empirical models of laser interference. Our evaluation in emulated and real-world setups on 2 state-of-the-art recognition systems and 5 cameras reports a maximum success rate of 30\% and 86.25\% for Red-to-Green and Green-to-Red attacks. 
We observe that the attack is effective in continuous frames from more than 40~meters away against a moving vehicle, which may cause end-to-end impacts on self-driving such as running a red light or emergency stop. 
To mitigate the threat, we propose redesigning the rolling shutter mechanism.

\end{abstract}


\section{Introduction}

A full autonomous driving requires the vehicle to behave independently not just on highways but also in urban settings, where traffic light recognition is essential.
Such systems enable the vehicle to visually detect and recognize the traffic light signals so that it can respond properly.
According to NHTSA~\cite{nhtsa}, intersection accidents are the second leading cause of vehicle collisions, led only by rear-end crashes. Reliable traffic light recognition can effectively prevent such accidents by provoking automatic responses of the vehicle. For this reason, car manufacturers such as BMW~\cite{bmw} and parts suppliers such as Mobileye~\cite{mobileye} are actively developing and testing their own traffic light recognition systems.
Tesla~\cite{tesla} has recently deployed traffic light recognition on their vehicles in the U.S.
Looking forward, we will not be surprised to see a prevalence of such systems in the near future.

Before that day, a critical question is how safe the vision-based traffic light recognition systems are.
The consequence of erroneous recognition is severe: recognizing a red light as green may cause intersection collisions; recognizing a green light as red may force stop the vehicle and jam the traffic.
Previous studies have shown the insecurity of similar vision-based object detection and recognition systems.
For example, a strong light can over-expose the camera and cause denial-of-service of the collision avoidance system~\cite{petit2015remote,yan2016can}; a projector can inject adversarial ghost images on an object~\cite{nassi2020phantom,lovisotto2020slap} or into the camera~\cite{man2020ghostimage} to fool the advanced driver-assistance system.
However, the security of traffic light recognition has not been studied yet. In addition, previous attacks generally used light that may raise stealthiness concerns and are constricted to an attack distance of a few meters. 

We are motivated to investigate the feasibility of fooling vision-based traffic light recognition in the real world. To achieve a stealthier effect and longer attack distance, we choose laser as the attack signal, which is a very narrow and energy-concentrated beam of light.
We envision that a laser shed into the camera's lens may modify the captured images and affect any subsequent recognition systems.
However, such an attack is non-trivial due to the following reasons.
(1) Though cameras are known to be susceptible to optical interference, a laser can easily saturate the CMOS sensor in the camera and render the entire image unrecognizable. It is unknown whether a laser can modify the image in a controllable way, especially only around the traffic light's area.
(2) Traffic light detection is normally performed before recognition. The attacker needs to ensure that the laser interference does not disrupt the traffic light detection so that it can fool the recognition.
(3) The red, green, and yellow lamps are on different parts of a traffic light. It is unknown whether and how an attacker can cause a red light to be recognized as a green light (or vice versa) without physically altering the lamps.

In this paper, we overcome these challenges and validate the attack's feasibility.
After studying all potential effects a laser may impose on camera imaging, we manage to create a color stripe overlapped on the traffic light in the image by exploiting the inherent vulnerability of the camera's rolling shutter.
Our experiments show that color stripes of specific cases will not disrupt the traffic light detection and can cause desired color recognition results.
To achieve a higher success rate, we empirically model the laser attack process and search for effective laser parameters based on simulated images under attack. We find that the attack can succeed with a wide range of laser parameters, suggesting high robustness to the uncertainties in real attack scenarios. 
We evaluate the attack in both emulated and real-world setups using real traffic light, laser diodes, and 5 cameras on 2 state-of-the-art open-source traffic light recognition systems. To demonstrate the potential real-world impacts, we experiment in practical settings and demonstrate successful attacks against a moving vehicle from more than 40 meters away in continuous frames.
We propose rolling shutter redesign to mitigate the threat. We hope this work can help build more security into traffic light recognition for future autonomous vehicles.

Our contributions are summarized as follows:
\begin{itemize}
	\item We discovered an inherent vulnerability of the rolling shutters in CMOS cameras that can be exploited by laser to create adjustable color stripes in an image, introducing a new approach to inject adversarial image patterns.
	\item We experimentally validated the feasibility of fooling traffic light recognition using laser, and proposed an optimization method to search for the effective laser parameters by simulating the attack with empirical models.
	\item We evaluated the attack in emulated and real-world setups on 2 traffic light recognition systems and 5 cameras, including one used on Tesla vehicles. The highest average success rates are 30\% and 86.25\% for Red-to-Green and Green-to-Red attacks. We built an attack equipment and demonstrated the feasibility of attacks in practical settings against a moving vehicle.
\end{itemize}

\section{Background}
\label{sec:background}

\subsection{Traffic Light Detection and Recognition}
\par Traffic light (TL) recognition is generally performed after traffic light detection. They are differentiated by the target used when taking traffic lights as input~\cite{mu2015traffic}. Traffic light detection's target is to box out the traffic lights' positions in a given image. Traffic light recognition, usually taking the result of traffic light detection as input, will further distinguish the traffic light's color. Usually, the autonomous vehicle system will combine the detection and recognition step to provide instructions to the vehicle. We briefly introduce a few state-of-the-art methods and open-source systems.
\par Apollo~\cite{apollo} is an open autonomous driving platform developed by Baidu. It uses YOLO~\cite{yolov3} to detect the traffic lights and treats the recognition as a typical CNN classification task. The RetinaNet receives an image with a list of bounding boxes with traffic lights as inputs and outputs a list with four elements, representing four possibilities for each box to be black, red, green, and yellow. Within red, yellow, and green, the class with the highest possibility will be regarded as the light's status when the possibility is over a certain threshold. Otherwise, the light's status will be set to black, meaning uncertain status. 
\par Other than separating the detection and recognition, there is another approach to combine these steps and treat traffic lights with different colors as different objects when training a classification model such as DNN. Therefore, systems~\cite{nexar} developed in such an approach will directly detect and classify red/yellow/green/black traffic lights in one step.

\subsection{Camera Imaging}





\begin{figure}
    \centering
    \includegraphics[width=0.45\textwidth]{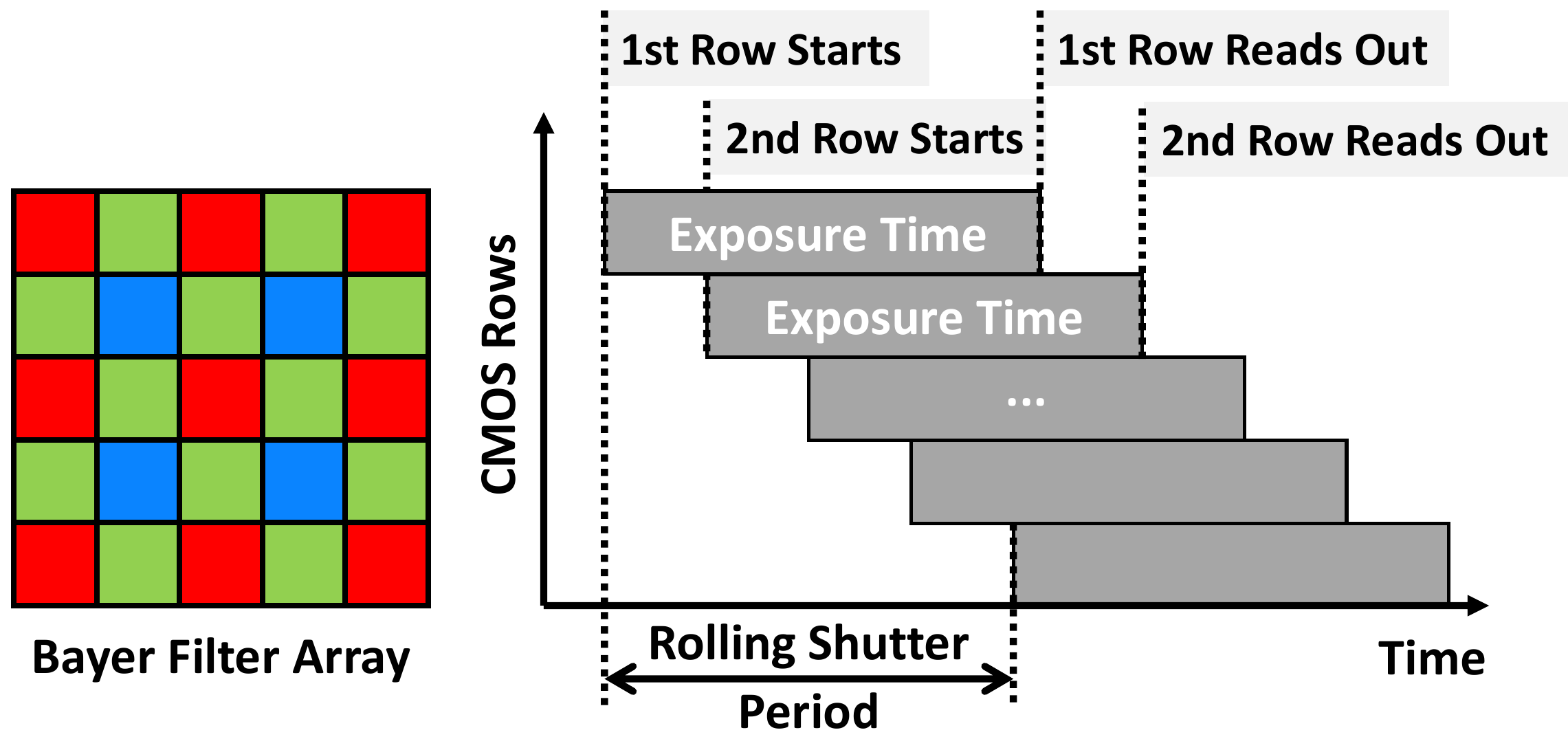}
    \vspace{-0.05in}
    \caption{Bayer filter array and rolling shutter.}
    \vspace{-0.2in}
    \label{fig: rolling shutter}
\end{figure}

A camera is an optical system that can capture images. Basically, a camera uses a lens to gather  light and a light-sensitive surface (usually photographic film or a digital sensor) placed on a focal plane to capture the image. A camera can control the luminous flux that reaches the surface by adjusting the aperture's size or exposure time. The latter corresponds to the amount of time the photosensitive surface is exposed to light, which is controlled by a shutter.
Digital cameras use digital image sensors as the photosensitive surface that can transfer light into digital signals. There are two major types of electronic image sensors, i.e., charge-couple device (CCD) and complementary metal-oxide-semiconductor (CMOS) sensors. Cameras in consumer products generally use CMOS sensors as they are cheaper and have lower power consumption. 

Image sensors capture the light intensity and can only create grayscale images. To acquire the color information, image sensors are typically overlaid with a ``color filter array,'' which consists of tiny color filters that only allow the light within specific wavelength ranges to pass through. Thus, the image processing algorithm can restore the color information by checking each pixel's color filter arrangement. The most commonly used color filter array is called ``Bayer Filter Array,'' which consists of alternating rows of red-green and green-blue filters, as shown in the left of Fig.~\ref{fig: rolling shutter}.

Most digital cameras adjust the exposure time by controlling the ``on-off'' of the photodiodes on the image sensor, i.e., via electronic shutters. There are two types of electronic shutters: global shutter and rolling shutter.
Global shutters enable the whole sensor to turn on and off simultaneously, while cameras with rolling shutters turn on and off the photodiodes row-by-row as if it is ``rolling.'' Therefore, cameras with rolling shutters do not capture the entire scene at a single time instant.
Due to the readout bottleneck of CMOS image sensors, most of them use rolling shutters. Fig.~\ref{fig: rolling shutter} demonstrates the timing diagram of a simplified rolling shutter sensor. Rows of pixels start exposure successively and get read out when the row exceeds exposure time. Meanwhile, the pixels are cleared for the next exposure. Since image rows are captured at different time instances, taking a picture of rapidly moving objects or flashing light can cause image distortion and skewness. 
We find that most cameras on vehicles (e.g., Tesla) use rolling shutters. Apollo Auto recommends the LI-USB30-AZ023WDRB camera, which is also a rolling shutter sensor.


\section{Threat Model}

\subsection{Attacker's Goal and Attack Scenarios}
The attacker's goal is to make self-driving vehicles make incorrect decisions near traffic lights by fooling the traffic light recognition system. In particular, there may be two scenarios:
\textit{1) Make a vehicle move while it should stop:} a vehicle should stop at red or yellow lights. Failing to stop accordingly may cause intersection accidents. 
\textit{2) Make a vehicle stop while it should move:} a vehicle should move at green lights when there is no obstacle. Such a case may trap the vehicle and even jam the traffic.

We envision that denial-of-service attacks and various types of color spoofing attacks against the traffic light detection and recognition systems may cause the above cases.
Table~\ref{tab:goal} summarizes the potential attack scenarios. 

\begin{table}[t]
	\centering
	\small
	\caption{Attacker's goals, attack scenarios, and outcomes.}
	\label{tab:goal}
	\begin{tabular}{cccc}
		\toprule
		Goal & DoS & Color Spoofing & Outcome \\
		\midrule
		Stop$\rightarrow$Move & N/A & R$\rightarrow$G & Collision \\
		Move$\rightarrow$Stop & Probable & G$\rightarrow$R, G$\rightarrow$Y, G$\rightarrow$B & Trap \\
		\bottomrule
	\end{tabular}
	\vspace{-0.15in}
\end{table}

\textbf{Denial-of-Service (DoS) Attacks.}  DoS attacks aim to disable traffic light detection, i.e., making a vehicle fail to detect existing traffic lights. To fail safe, a vehicle is supposed to slow down or stop at an intersection when no traffic light is detected. However, a vehicle may easily notice from the map if a traffic light is missing and take safety precautions. 

\textbf{Color Spoofing Attacks.} Such attacks aim to alter the results of traffic light recognition, i.e., causing the traffic light's color status to be interpreted incorrectly. 
A traffic light has four color statuses: red (R), green (G), yellow (Y), and black (B), where black means that the light is off. We consider the most common traffic light status to be either red or green because black is rare in real life and yellow only lasts for a very short time. An attacker may make a vehicle move by making a red light recognized as green, and force a vehicle to stop by making a green light recognized as red, yellow, or black.
In this paper, we focus on color spoofing attacks as they are stealthier and pose a more significant threat.

\subsection{Attacker's Capabilities}
We consider an attacker with the following capabilities.

\textbf{Laser Interference to the On-board Camera.} 
The attacker can only alter the results of traffic light detection and recognition by remotely shedding a laser interference to the camera on the target vehicle. She cannot physically alter existing traffic lights or build a spurious traffic light.

\textbf{No Direct Access to the Target Vehicle.}
The attacker has no direct physical or digital access to the target vehicle before or during the attack. She cannot make any changes to the on-board camera or the traffic light detection and recognition system. The target vehicle remains enclosed to the attacker.

\textbf{Camera and System Awareness.}
The attacker may infer the model and structure of the target camera and system based on public information or reverse engineering.
She may obtain the same camera or vehicle to assess the attack beforehand in setups similar to the planned attack scenario. For example, she could drive a vehicle of the same model through the target intersection and record images of the target traffic light. However, she can only treat the target camera as a black box.

\begin{figure}[t]
    \centering
    \includegraphics[width=0.48\textwidth]{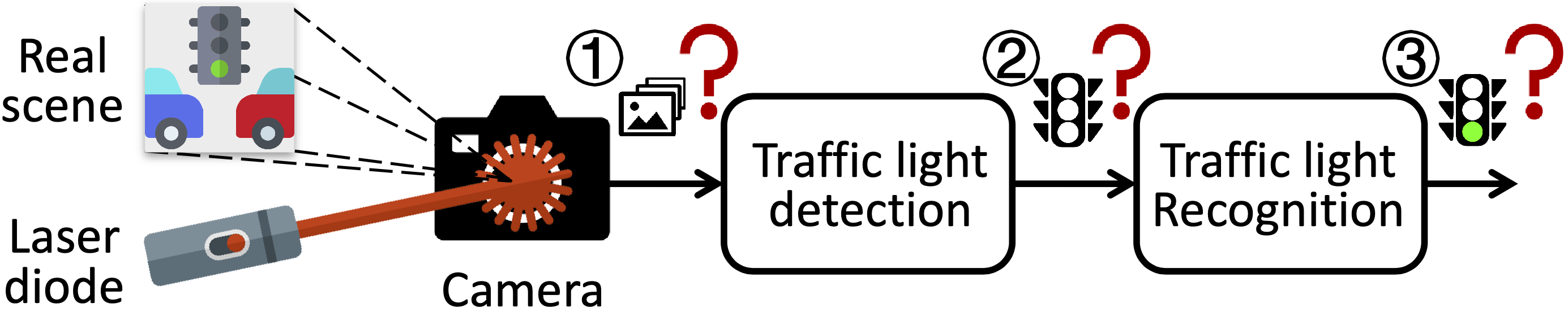}
    \vspace{-0.2in}
    \caption{Illustration of the three attack stages.}
    \vspace{-0.2in}
    \label{fig: Attack Model}
\end{figure}

\section{Feasibility Study}
In this section, we study the feasibility of attacking vision-based traffic light detection and recognition mechanisms using laser interference. We consider the three attack stages shown in Fig.~\ref{fig: Attack Model}.

\begin{figure*}[t]
    \centering
    \subfigure[Original]{
            \includegraphics[width=0.152\textwidth]{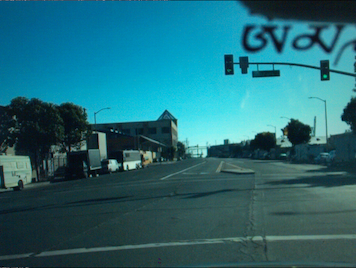}
    }
    \subfigure[Overexposure]{
            \includegraphics[width=0.152\textwidth]{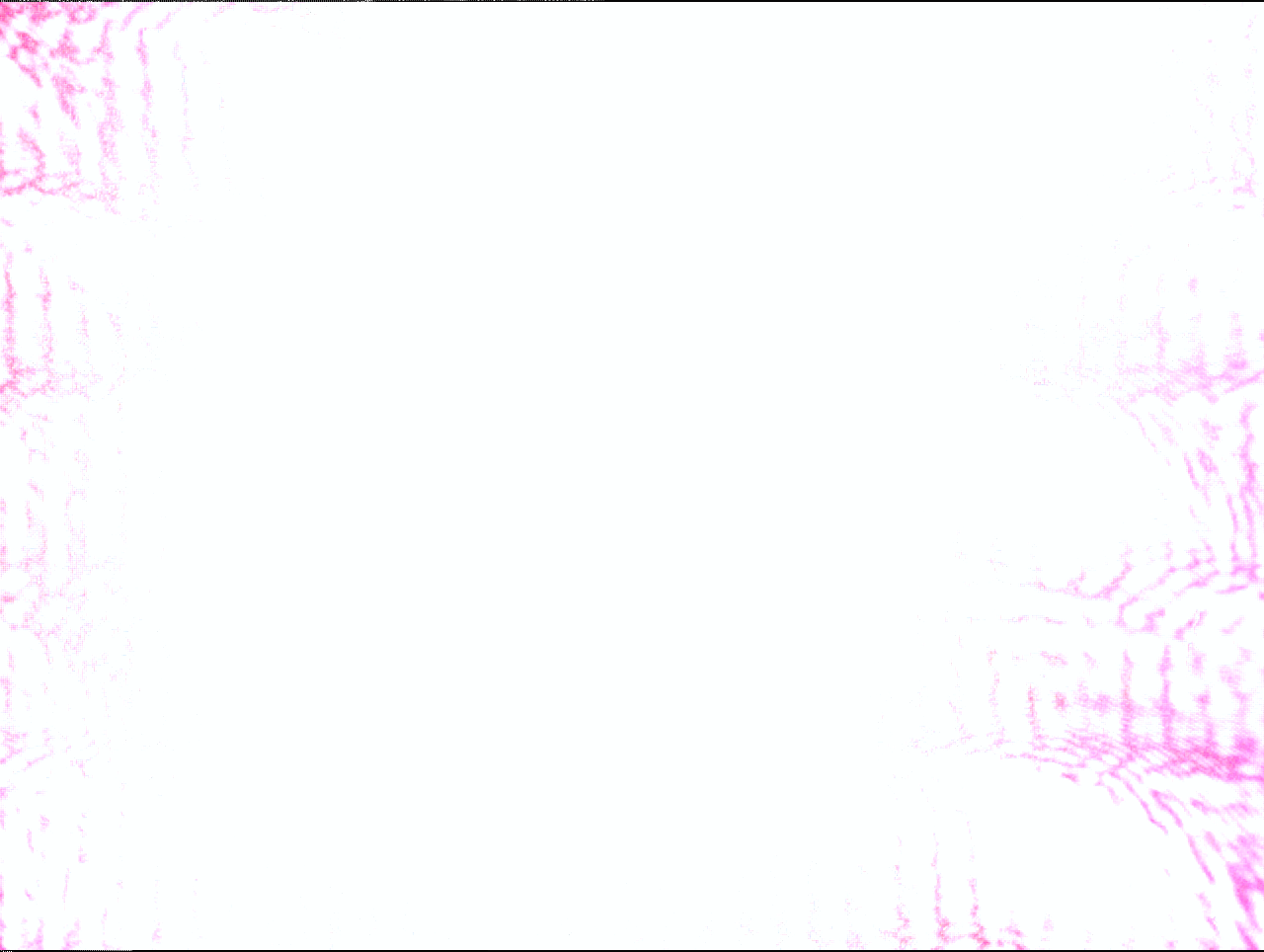}
            \label{fig:demo_overexpose}
    }
    \subfigure[Color shift]{
            \includegraphics[width=0.152\textwidth]{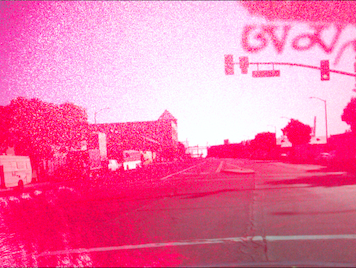}
            \label{fig:demo_colorshift}
    }
    \subfigure[Color stripe]{
            \includegraphics[width=0.152\textwidth]{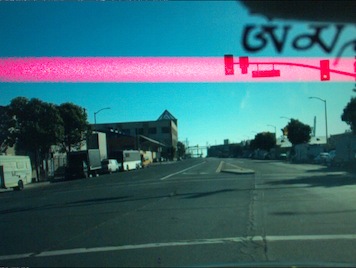}
            \label{fig:demo_colorstripe}
    }
    \subfigure[White unbalance]{
            \includegraphics[width=0.152\textwidth]{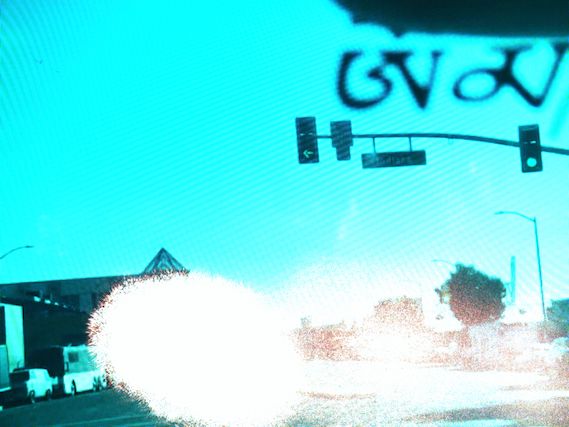}
            \label{fig:demo_whiteunbalance}
    }
    \subfigure[Out-of-focus]{
            \includegraphics[width=0.152\textwidth]{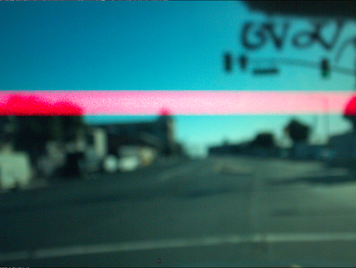}
            \label{fig:demo_outoffocus}
    }
    \vspace{-0.1in}
    \caption{Images captured by an AR0132AT evaluation board camera under various types of laser interference.}
    \vspace{-0.05in}
    \label{fig: demos}
\end{figure*}

\textit{1. Camera Imaging}: the attacker generates a laser beam and points it at the camera.  Key questions in this stage include a) what types of laser interference an attacker can generate and b) what images will a camera capture under laser interference.

\textit{2. Traffic Light Detection}: disabling traffic light detection is sufficient for a DoS attack. However, for color spoofing attacks, the attacker must ensure that the traffic light can be detected with sufficiently high confidence.

\textit{3. Traffic Light Recognition}: in this stage, the attacker intends to make a traffic light recognized as showing a different color. However, it is unknown whether this attack is feasible and what type of laser interference will be effective.

We examine the feasibility of these stages in the following.

\subsection{Laser Interference Examination}

We choose semiconductor laser diodes as the laser generator due to their low-cost, high accessibility, and high power-efficiency. There are four attack parameters to consider that determine the laser interference pattern on the image:

    \textbf{Wavelength.}
    A laser's wavelength corresponds to its color, which directly affects the color of the resultant interference on the image. A laser diode's wavelength is generally fixed as it is determined by the semiconductor's structure and material. Therefore, an attacker can inject interference of various colors using a series of laser diodes with different wavelengths. 


    \textbf{Power.} A laser's power affects the interference's intensity as well as the attack distance. A higher power enables a longer attack distance, but it may also saturate the CMOS sensor and blind the camera. In a color spoofing attack, the attacker has to carefully adjust the laser's power to ensure that the interference on the image is neither too weak nor too strong.
    Previous work~\cite{sugawara2020light} has shown that a laser's power increases linearly with the driving current once it is above a diode-specific threshold.

    \textbf{Pulse Width and Period.} 
    The laser can be continuous or in the form of a series of short pulses.
    A continuous laser affects all image frames in a video, while laser pulses can affect selected frames in a video and even a part of an image. An attacker can change the width and period of laser pulses by applying pulse width modulation (PWM) to the laser diode's driving current.



    \textbf{Incidence angle.} The incidence angle is the angle between the laser beam and the optical axis of the camera lens. The incidence angle mainly affects the brightness distribution of the interference. For example, a laser emitted from the right side of the camera will cause a brighter interference on the right part of the image.


\subsection{Imaging under Laser Interference}
Existing studies~\cite{petit2015remote,yan2016can,truong2005preventing,lan2016live} have shown that light and laser interference can significantly change the images captured by cameras. However, their work focused on DoS attacks, i.e., overexposing the camera, and cannot be used for color spoofing.
In this paper, we thoroughly examine and report all potential impacts of laser interference by experimenting with the laser parameters mentioned above.
Besides overexposure, we discover that laser interference can cause the color shift, color stripe, white unbalance, and out-of-focus of an image, as shown in Fig.~\ref{fig: demos}. The images are taken by an AR0132AT evaluation board~\cite{ar0132at} camera.





\textbf{Overexposure.}
A laser of sufficient power can easily saturate all three color channels of a CMOS sensor and cause overexposure of the image, as shown in Fig.~\ref{fig:demo_overexpose}. 
Overexposure is the most effective way to cause the denial-of-service of any camera-based system.



\textbf{Color Shift.}
By lowering the laser's power below the saturation point, we can reduce the interference's brightness to a range that only affects a single color channel, causing a ``color shift'' of the image, as shown in Fig.~\ref{fig:demo_colorshift}. The shifted color is determined by the laser's wavelength. 
In this example, a laser of 650~nm wavelength increases the red channel's value much more than the green and blue channels in the CMOS, making the image more reddish.



\begin{figure}[t]
    \centering
    \subfigure[Generation of a color stripe]{
        \includegraphics[width=0.225\textwidth]{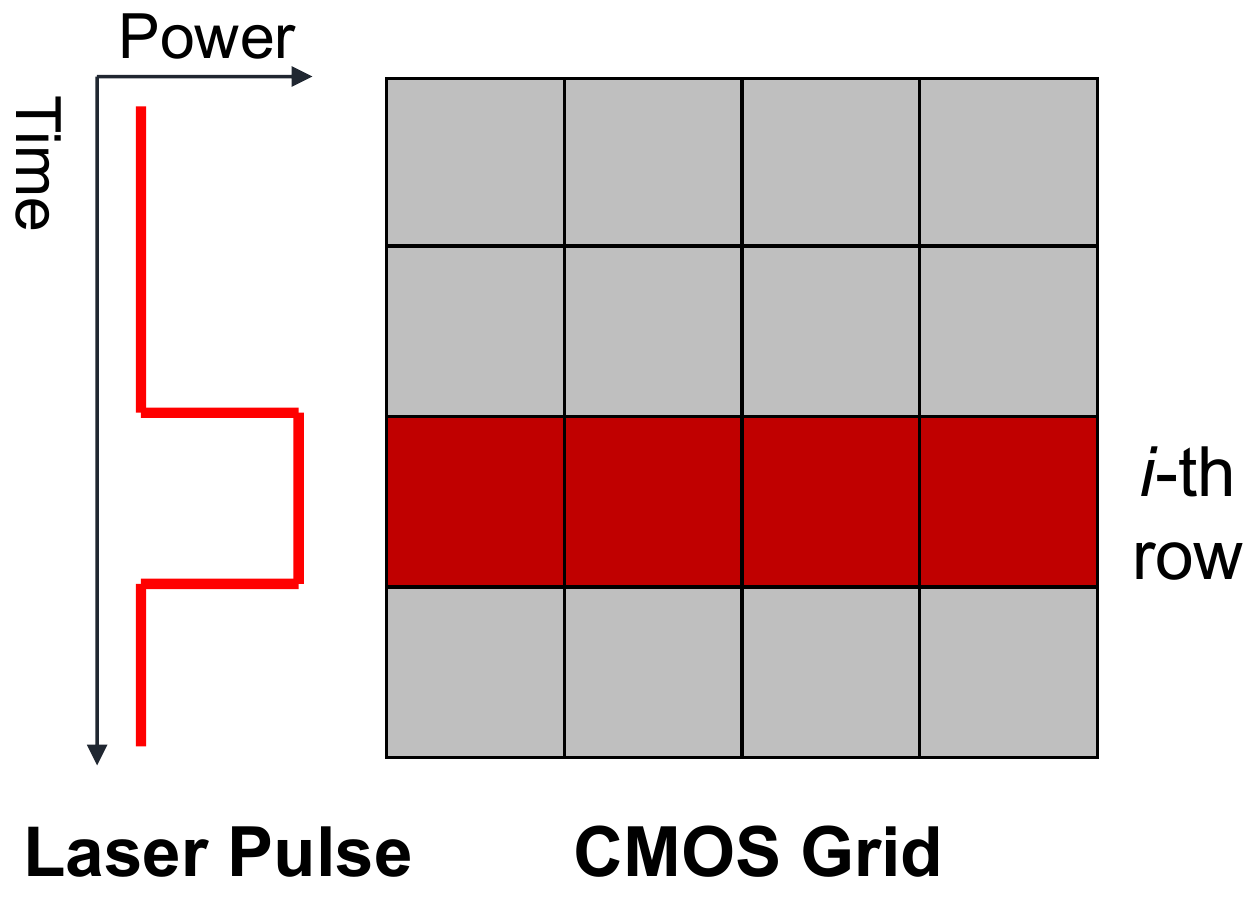}
        \label{fig:colorstripe_principle}
    }
    \subfigure[Stripe width and position]{
        \includegraphics[width=0.225\textwidth]{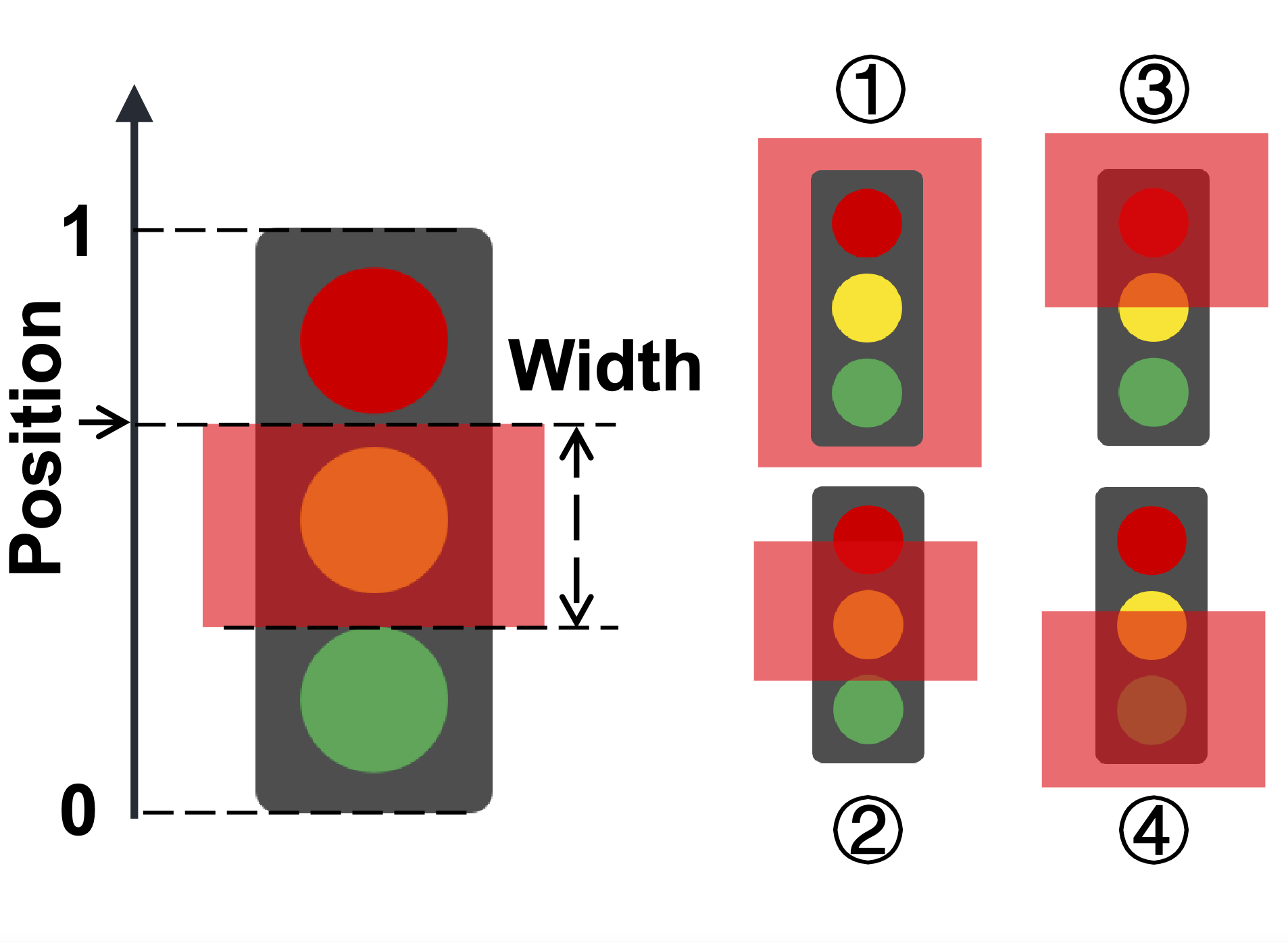}
        \label{fig:colorstripe_position}
    }
    \vspace{-0.1in}
    \caption{Illustrations of (a) the generation of a color stripe, and (b) the stripe's width and position compared with the traffic light and four typical cases.}
    \vspace{-0.1in}
    \label{fig: colorshift_CL}
\end{figure}

\textbf{Color Stripe.}
The color shift appears on the entire image and may unintentionally impede the recognition of other objects.
We wonder whether the color shift can occur only in a part of the image, especially where the traffic lights are.
We manage to create a ``color stripe'' on the image by exploiting the rolling shutter of CMOS sensors with modulated laser pulses, as shown in Fig.~\ref{fig:demo_colorstripe}. The principle is illustrated in Fig.~\ref{fig:colorstripe_principle}. As the shutter ``rolls'' from the first row of the image to the last one, a laser pulse that appears only when the $i$-th row of CMOS is activated will interfere with only the $i$-th row on the image, thus leaving a color stripe.
An attacker can change the width, position, brightness, and number of the stripes by adjusting the modulation parameters of the diode's driving signal. 
She could even stabilize the color stripe in a sequence of video frames by synchronizing the pulse period with the rolling shutter speed (video frame rate), e.g., 30~Hz.


\textbf{White Unbalance.}
White balance is an in-built function in most digital cameras that automatically balances the image's color temperature. A laser interference on a part of an image can make the white balance algorithm wrongly adjust the color gains of the rest of the image, as shown in Fig.~\ref{fig:demo_whiteunbalance}. An attacker may exploit this phenomenon to cause ``white unbalance'' of the traffic light, i.e., changing its color without overlaying interference on it.
However, the color change may not be intense due to the limited white balance ability.


\textbf{Out-of-Focus.}
We discover that moving color stripes can make the camera out-of-focus, potentially leading to denial-of-service.
Rapidly moving color stripes can periodically change the contrast of the sampling area of the focus algorithm, and hence force the camera to refocus.
During the refocusing process, the background will be out-of-focus and become blurry, as shown in Fig.~\ref{fig:demo_outoffocus}.

In summary, an attacker may achieve denial-of-service with overexposed or out-of-focus images and achieve color spoofing with color-shifted or color-striped images.
A color-shifted image is essentially a color-striped image when the stripe is as wide as the image.
In a preliminary experiment reported in Fig.~\ref{fig:stripe-width}, we find that compared with a narrow stripe, a wider stripe will greatly reduce the confidence score of traffic light detection and unnecessarily causing denial-of-service.
Thus, we will focus on attacks using a narrow stripe in this paper.


\begin{figure}[t]
    \centering
    \subfigure[Red laser (650~nm)]{
        \hspace{-0.2in}
        \begin{minipage}{0.46\linewidth}
            \centering
            \includegraphics[width=1.1\textwidth]{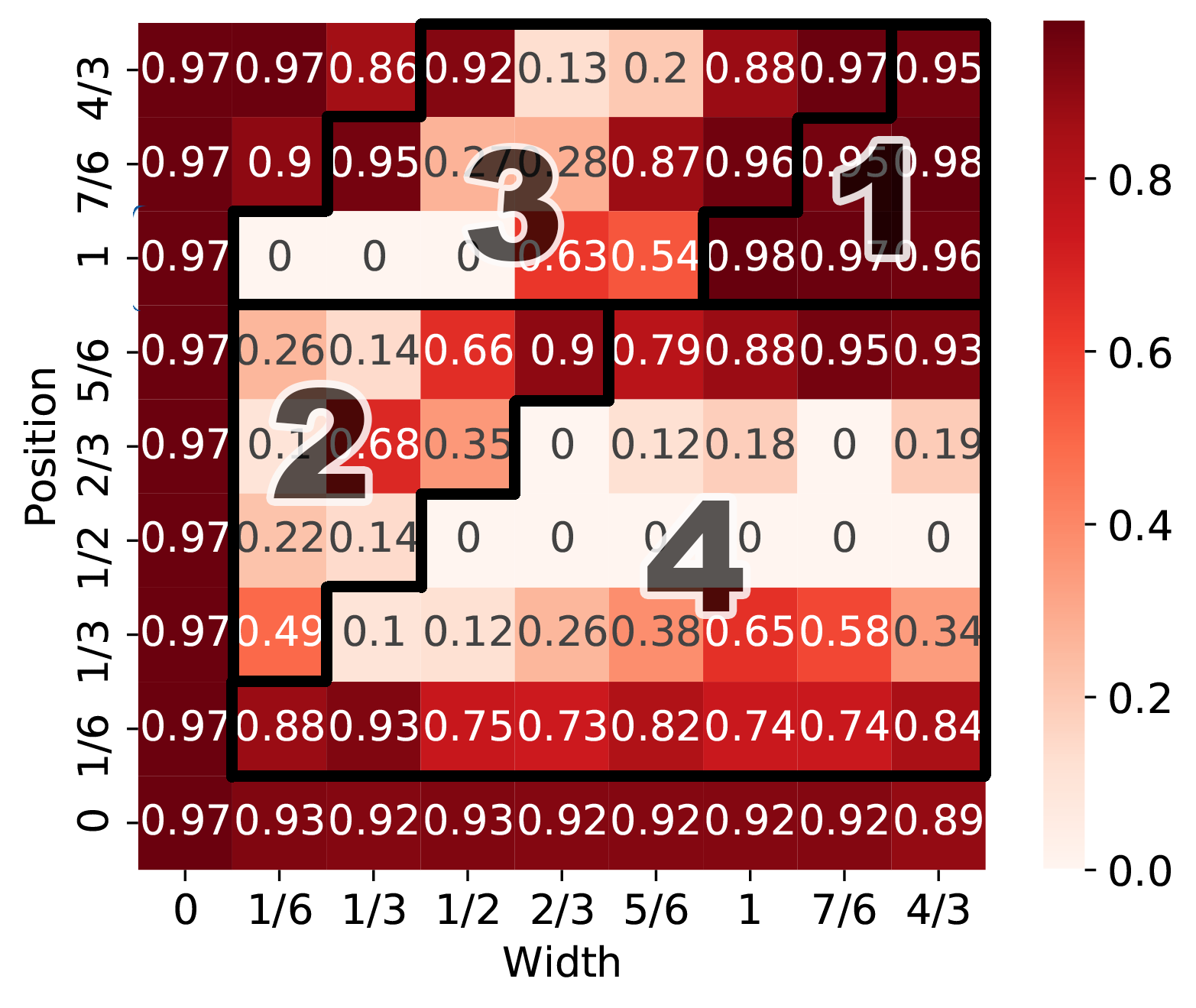}
        \end{minipage}
    }
    \subfigure[Green laser (520~nm)]{
        \hspace{0.05in}
        \begin{minipage}{0.46\linewidth}
            \centering
            \includegraphics[width=1.1\textwidth]{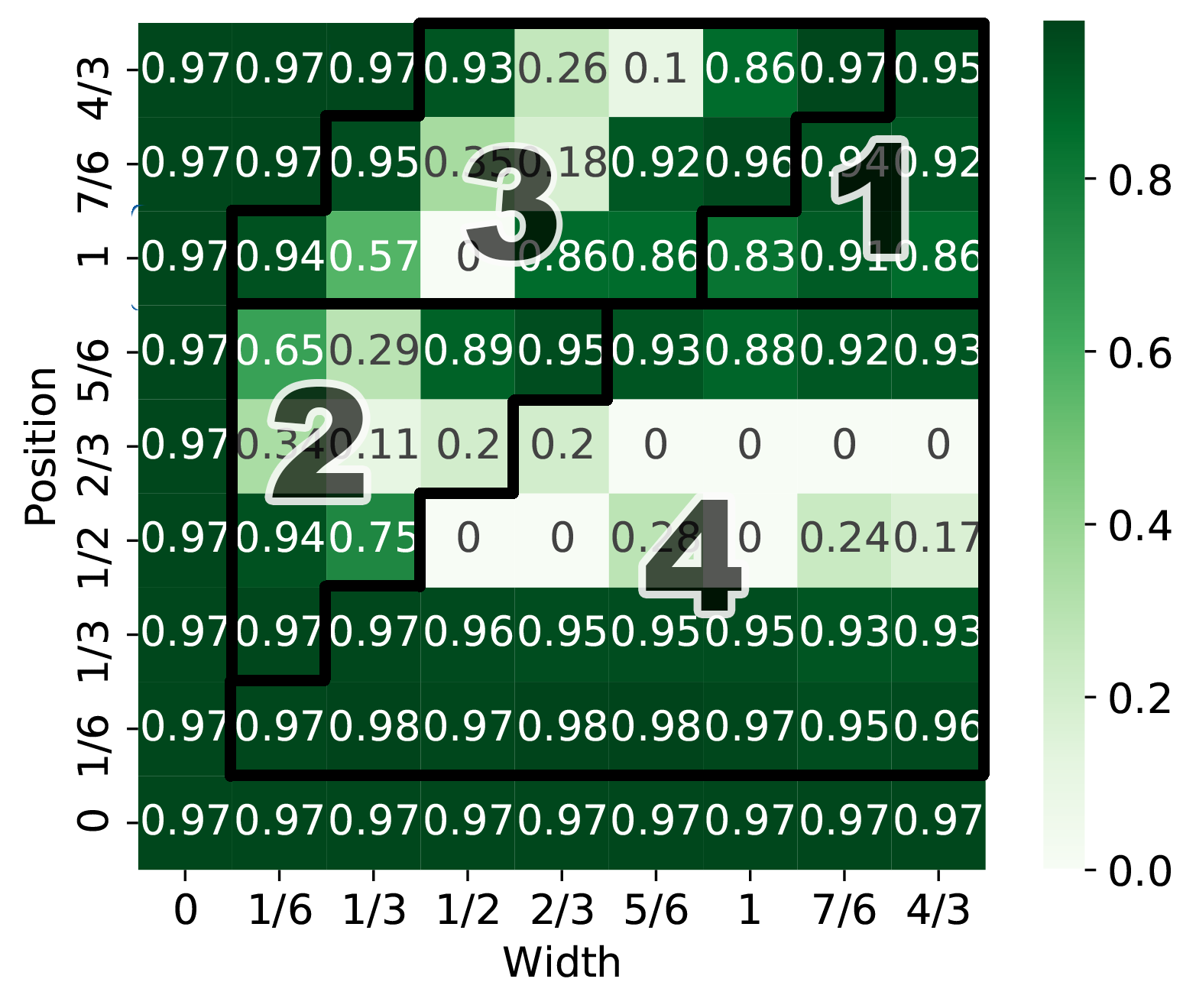}
        \end{minipage}
    }
    \vspace{-0.1in}
    \caption{The confidence scores of detecting a traffic light covered by color stripes of various widths and positions. The 4 cropped regions correspond to the 4 cases in Fig.~\ref{fig:colorstripe_position}. It shows that case \ding{172} affects traffic light detection the least.}
    \vspace{-0.1in}
    \label{fig: heatmap}
\end{figure}

\subsection{TL Detection under Laser Interference}
\label{sec:detection}
To successfully spoof the results of traffic light recognition, the attacker has to first ensure that the color stripes do not affect traffic light detection. Thus, a question of interest is what type of color stripe less impedes traffic light detection. 

We examine the impact of color stripes by comparing the confidence scores of detecting a traffic light under various combinations of stripe parameters, where we are mainly concerned with the stripe's width and position.
We use a trained YOLOv4 classifier as the traffic light detection system and use red and green lasers to inject color stripes of various widths and positions into a traffic light image.

Two heatmaps in Fig.~\ref{fig: heatmap} show the confidence scores of detecting the traffic light under red and green laser interference. 
The x-axis is the stripe's width, which varies from 0 to $4/3$ of the traffic light's height.
The y-axis indicates the position of the stripe's upper edge compared with the vertical size of the traffic light, also varying from 0 to $4/3$. Fig.~\ref{fig:colorstripe_position} provides a visual illustration of the position and width.
Position ``0'' means that the stripe's upper edge is at the bottom edge of the light, while position ``1'' suggests that the stripe's upper edge coincides with the light's upper edge.
To help understand the results in Fig.~\ref{fig: heatmap}, we divide the heatmaps into four regions that correspond to the four typical cases in Fig.~\ref{fig:colorstripe_position}.
In case~\ding{172}, the stripe covers the entire traffic light; in case~\ding{173}, the stripe covers only a part of the light between its upper and lower edges; while in case \ding{174} and \ding{175}, the stripe covers the upper or lower parts of the light including its edges.

The results in Fig.~\ref{fig: heatmap} show that, regardless of the laser's color, the accuracy of traffic light detection is barely affected by the color stripe if it covers the entire light, i.e., case~\ding{172}. We observe a great decrease in the confidence scores in other cases, which we assume is because the color stripe breaks the traffic light's continuity, especially on its original edges.
This experiment motivates us to create color stripes similar to case~\ding{172} for color spoofing attacks.


\begin{figure}[t!]
    \centering
    \subfigure[Turn a green light to red light]{
            \includegraphics[width=0.225\textwidth]{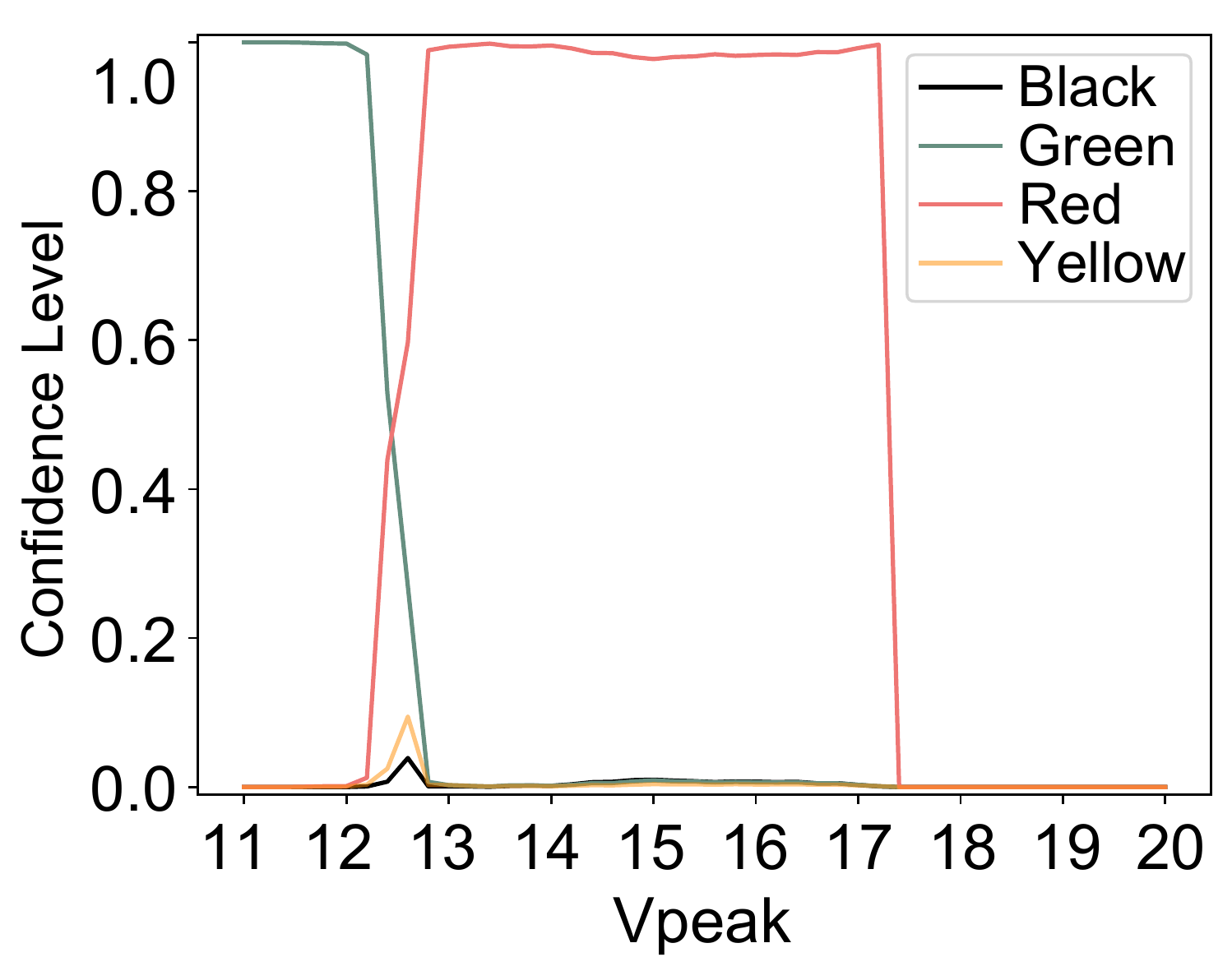}
    }
    \subfigure[Turn a red light to green light]{
            \includegraphics[width=0.225\textwidth]{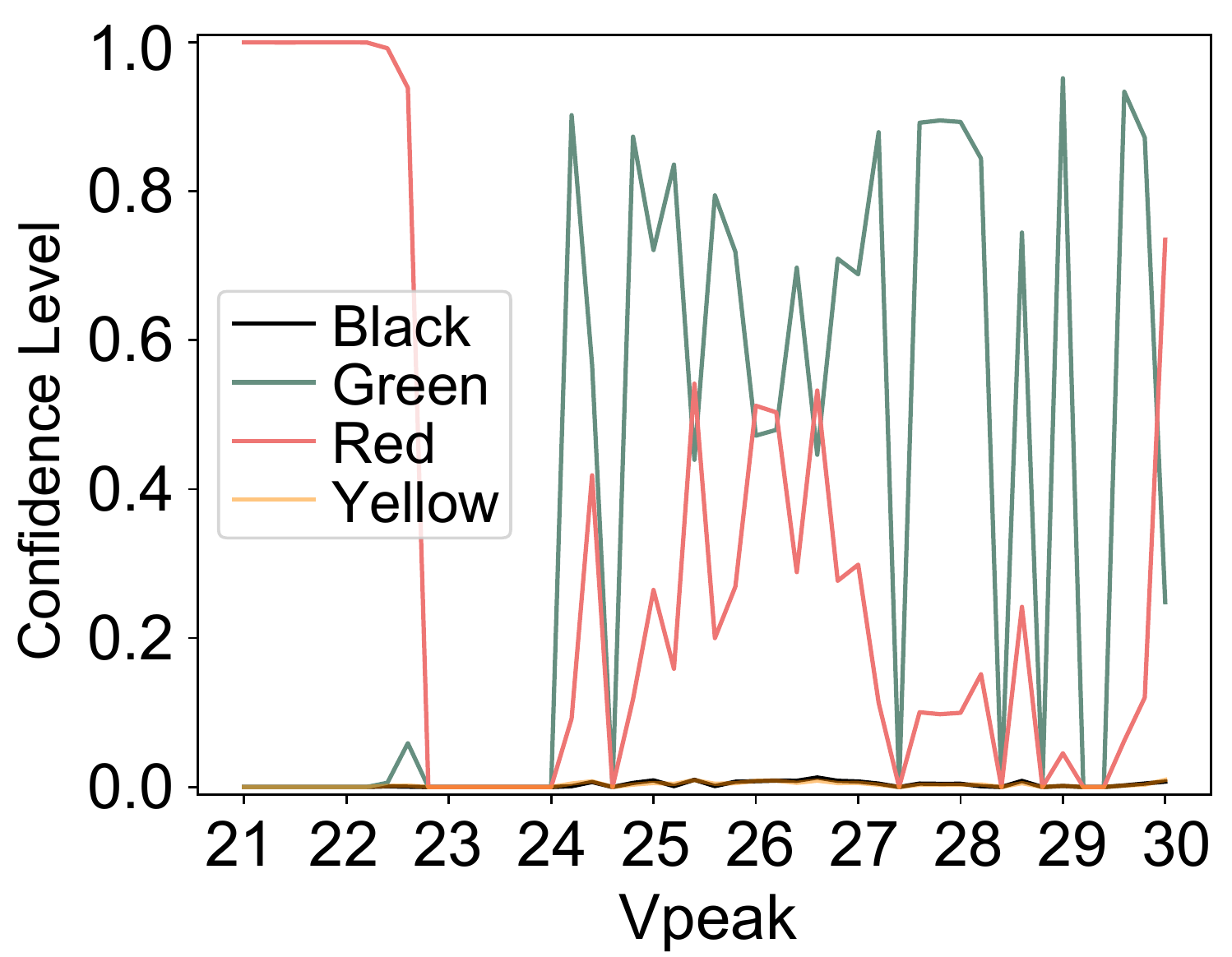}
    }
    \vspace{-0.1in}
    \caption{Confidence scores of traffic light recognition as the laser diode's driving voltage (i.e., power) varies. It shows that though the effective laser parameters may exist in a wide range, they need to be adjusted in a case-by-case manner.}
    \vspace{-0.1in}
    \label{fig:feasibility_recognition}
\end{figure}



\subsection{TL Recognition under Laser Interference}


\begin{figure*}[ht]
 \centering
 \includegraphics[width=\textwidth]{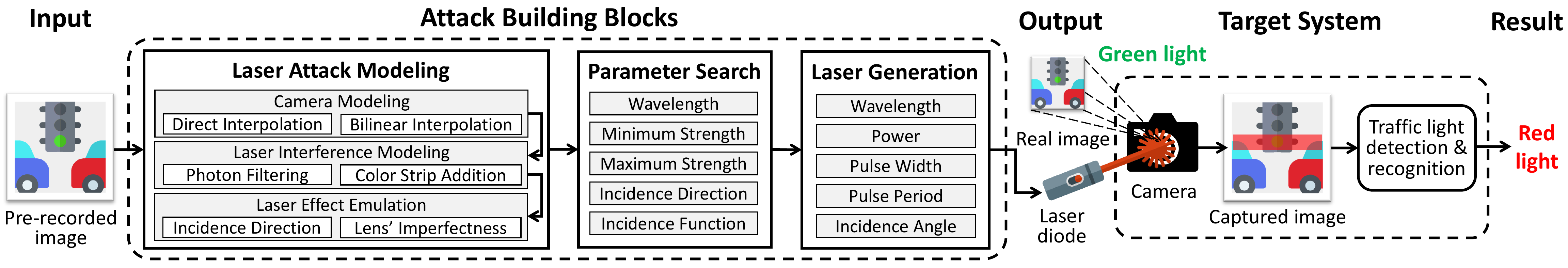}
 \vspace{-0.3in}
 \caption{The attack workflow. Based on laser attack modeling, the attacker first simulates laser interference on pre-recorded images of the target traffic and search for the most effective sets of laser attack parameters that can achieve the desired result of traffic light recognition. The attacker then uses the parameters to generate a laser attack signal and point it at the target camera.}
 \vspace{-0.1in}
 \label{fig: workflow}
\end{figure*}

To examine the feasibility of spoofing traffic light recognition, we consider two attack cases: (a) turn a green light to red with a red stripe, and (b) turn a red light to green with a green stripe. The stripes are generated using 650~nm red and 520~nm green laser diodes, respectively.
The images of the interfered traffic lights are sent to Apollo for recognition.

In Apollo, the system is programmed to first detect traffic lights within a cropped region of interest (ROI) that is previously known from the HD-Map that contains traffic lights' locations. Since the HD-Map is unavailable, we provide the ROI to the system manually to make Apollo function normally. If a traffic light is detected, Apollo outputs the confidence scores of four possibilities: red, yellow, green, and black.

The results show that color spoofing attacks are feasible for both cases. We are able to make a red light recognized as a green one and vice versa. However, we also observe that the attack does not succeed all the time---the confidence scores can be affected greatly by the laser parameters, especially the power. For example, Fig.~\ref{fig:feasibility_recognition} shows the scores of the recognition result when we adjust the laser's power by the diode's driving voltage.
It shows a red laser can spoof a green light with high confidence scores when the voltage is between 13 to 17 volts. However, there is no ``safe zone'' of power when we try to make a red light recognized as green. The effective voltage region is discrete and appears random.
In our experiment, we find that the attack result can also be affected greatly by the laser's incidence angle, the traffic light itself, and even the image background. 
Therefore, it would be inadvisable to use the same set of laser parameters to attack different traffic lights.
In practice, it is critical to fine-tune the laser parameters according to specific traffic lights and attack scenarios.






\section{Attack Design}


To obtain the most effective laser parameters case-by-case, we model the laser attack process and emulate the attack based on pre-recorded images of the target traffic light.
The attack workflow is summarized in Fig.~\ref{fig: workflow}.
                              
First, to simulate laser interference on cameras, we develop an empirical laser attack model that takes a raw image as input and outputs an image with simulated laser interference. The model consists of three parts: camera modeling, laser interference modeling, and laser effect emulation. Camera modeling describes the transformation from the RAW images captured by the CMOS to common image files such as JPG. Laser modeling simulates the laser interference by mathematically modeling the addition of color stripes to an image. Laser effect emulation further involves a laser's incidence direction and noise to the model to make the simulated color stripe look more realistic. 
Then based on the laser attack model, we use a grid search method to seek the most effective combinations of attack parameters. 
Finally, we map the attack parameters derived from our model to adjustable laser parameters and generate laser attack signals in the real world.


\subsection{Camera Modeling}
\par The first step is to transform the CMOS sensor's raw images to jpg images. 
As introduced in Section~\ref{sec:background}, CMOS, as well as the raw file, will store the color information of the image in the form of R, Gr, Gb, B in the four adjacent pixels respectively that form a square. Therefore, interpolation is required to transform the matrix from a raw file to a tensor that represents RGB values. There are mainly two types of interpolation: direct and bilinear. We consider both of them in our model, and the specific model to use depends on the target camera.
We leave the mathematical details to Appendix~\ref{sec:appendix_cameramodeling}.

\textbf{Direct Interpolation.}
For each pixel in the jpg image, its RGB values are determined by the closest pixel that stores Red, the mixture of Gr and Gb, or Blue in the matrix from the raw file. As for each two-by-two block in the raw file, there will be at least one pixel representing ``red,'' one representing ``blue,'' and two representing ``Gr'' and ``Gb,'' respectively, the RGB value of the jpg-image-pixels in this block will be directly determined by those pixels values in the raw file.


\textbf{Bilinear Interpolation.}
For each pixel in the jpg image, its RGB value is determined by at least two closest pixels in the raw file. In general, for each interior pixel in the jpg file, it will take a three-by-three block centered at the pixel. All the pixel values representing the corresponding color (R, G, or B) in the raw file are averaged to calculate the RGB value in the jpg file. For each pixel on the edge of the image where the three-by-three block is not available, we will still find the two closest pixels in the raw file representing the same color and calculate their average.


\subsection{Laser Interference Modeling}
We consider two steps in the laser interference modeling: photon filtering and color stripe addition.

\begin{figure}[t]
    \centering
    \includegraphics[scale=0.25]{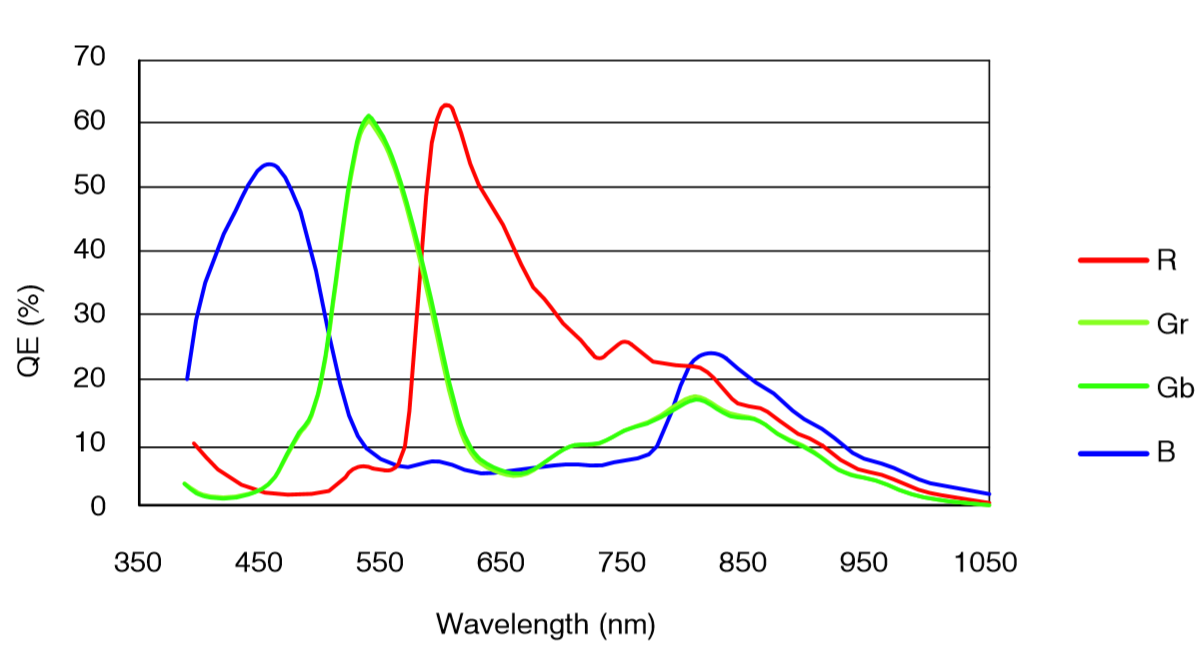}
    \vspace{-0.1in}
    \caption{Quantum Efficiency curves for ON Semiconductor CMOS Digital Image Sensor MT9P006~\cite{qe}.}
    \vspace{-0.05in}
    \label{fig: QE}
\end{figure}

\textbf{Photon Filtering.}
When a laser is shed into a CMOS, the photons that each photodiode can detect are determined by its quantum efficiency, which describes the ratio of the number of electrons measured by a photodiode to the number of photons fed to it. Each CMOS has its unique quantum efficiency. Also, the color filters (R, Gr, Gb, B) on each photodiode make the quantum efficiency unique for the light of different wavelengths. For example, Fig.~\ref{fig: QE} shows the quantum efficiency curves for an ON Semiconductor CMOS Digital Image Sensor MT9P006. Green light of 550~nm wavelength can be measured by the Gr and Gb photodiodes with a quantum efficiency of 60\%, while it is less received by the R and B photodiodes because the quantum efficiency is less than 10\%. However, when the light strength is sufficiently high, its effect on the red and blue pixels becomes obvious, which explains why in Fig.~\ref{fig:demo_overexpose}, a single-colored laser can saturate all color channels and make the entire image white.

\par Suppose $\textit{I}$ stands for the laser's strength, $\lambda$ stands for its wavelength, and $R^*,G^*,B^*$ stands for the effect of the laser on the Red, Green, Blue channels of the image, respectively, the filtering effect of the CMOS sensor due to quantum efficiency could be represented by
    $X^* = I f_X(\lambda), X \in \{R,G,B\}$.
Here, $f_R(\cdot), f_G(\cdot), f_B(\cdot)$ stand for the quantum efficiency curves of the CMOS sensor on the target camera. Since the curves could not be expressed parametrically, we interpolate the functions by tracing the quantum efficiency curves of the target camera.

\textbf{Color Stripe Addition.}
Suppose the effect a laser interference has on an RGB image is the addition on the three color channels, we can express the RGB values of the post-image (image taken under the laser attack), $R_p,G_p,B_p$, by the RGB values of the original image, $R_o,G_o,B_o$, and the effect of the laser, $R^*,G^*,B^*$ as $X_p = H(X_o,X^*), X\in\{R,G,B\}$.
Here, there is no specific choice for function $H(\cdot,\cdot)$. Considering the color channel overflow, here we use a relu-like function for $H(\cdot,\cdot)$:
    $H(x,y) = \min\left(x+y,c_{max}\right)$.
Here, $c_{max}$ stands for the overflow value for a specific channel, and we choose 255 for simplicity.

\par Combining the photon filtering and color stripe addition processes, the effect of the laser interference is determined by the RGB values of the original image $X_o$, the wavelength $\lambda$ and strength $I$ of the laser, and the quantum efficiency $f_X(\cdot)$ of the CMOS sensor as:
\begin{equation*}
    X_p = H(X_o,I f_X(\lambda)) \qquad X \in \{R,G,B\}
\end{equation*}

\begin{figure}[t]
    \centering
    \subfigure[Real laser interference]{
        \includegraphics[width=0.225\textwidth]{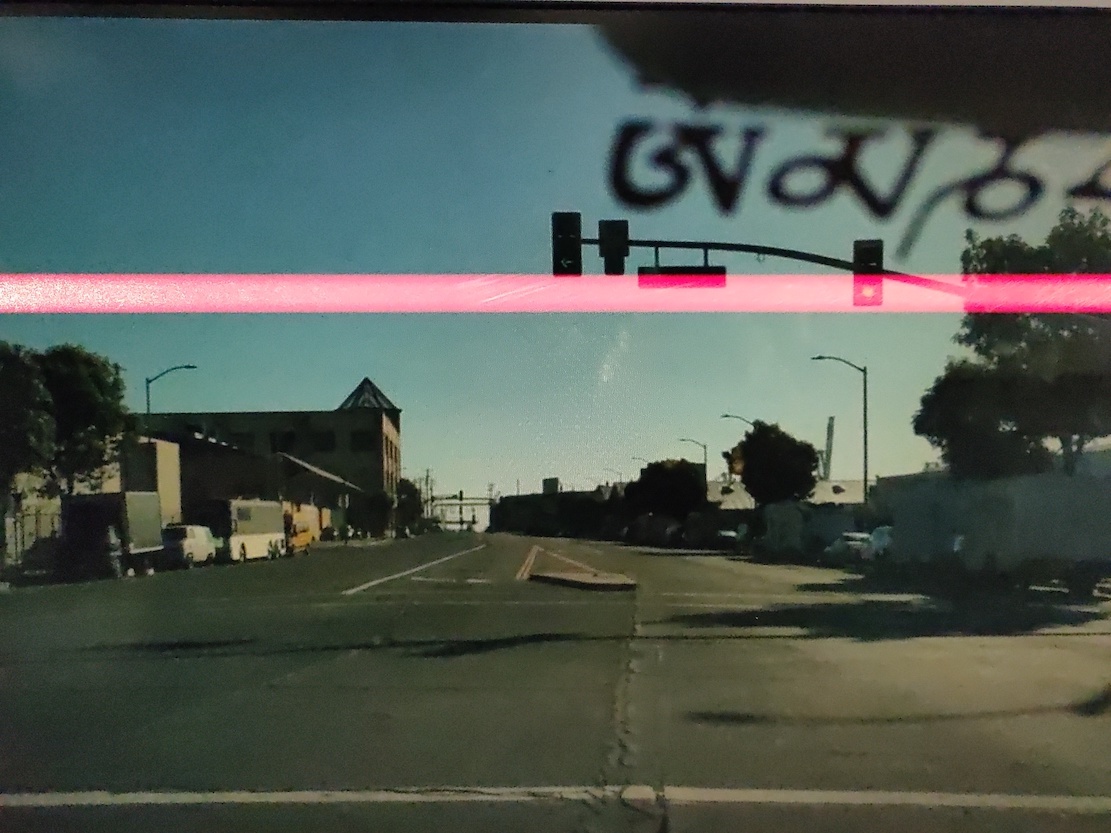}
        \label{fig:reallaser}
    }
    \subfigure[Simulated laser interference]{
        \includegraphics[width=0.225\textwidth]{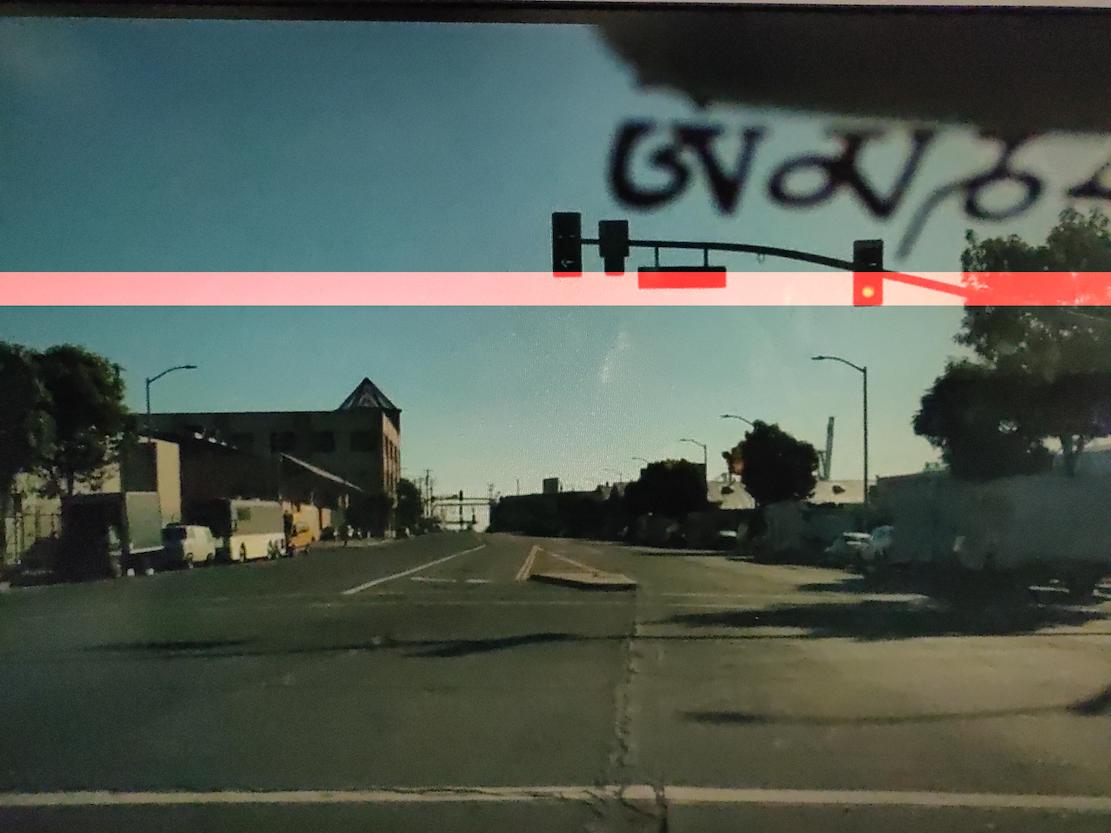}
    }
    \vspace{-0.1in}
    \caption{Comparison between the real laser interference and our simulation, showing a high similarity.}
    \vspace{-0.1in}
    \label{fig: Real vs Simulated}
\end{figure}

Fig.~\ref{fig: Real vs Simulated} shows a comparison between the real laser interference and our simulation.
Notice that our simulated laser interference is highly similar to the real one, indicating our model's efficacy. However, there are slight differences in the details, which we assume are caused by the laser's incidence direction and noise.

\subsection{Laser Effect Emulation}
To make the simulation more realistic, we include the laser's incidence direction and lens imperfectness into our model. The former determines the color stripe's brightness distribution on the image, and the latter produces noise.


\textbf{Laser's Incidence Direction.}
The laser's incidence direction has a significant effect on the color stripe's distribution in the image. For example, in Fig.~\ref{fig:reallaser}, the laser is irradiated from the left side of the camera. Therefore, the laser's intensity is higher on the left than on the right, which results in a channel overflow (white pixels) on the left and an increased level of noise. For simplicity, in our model, we consider the laser from left, right, or front of the camera. Instead of using a single intensity $I$ for the laser, we define the minimum and maximum intensities $I_{min},I_{max}$ measured by the CMOS, and assume the width and height of the color stripe are $w,h$, respectively. The intensity of the laser at a specific point $(x,y)$ can be represented as $I = D\left(I_{min},I_{max},x,y,h\right)$. Empirically, we consider three potential cases for the function $D(\cdot)$: linear, sigmoid, and Gaussian. Linear or sigmoid functions are used when the incidence direction is from the sides; the Gaussian function is used when the light is irradiated from the camera's front. We leave the details to Appendix~\ref{sec:appendix_lasereffect}.

\textbf{Lens Imperfectness.}
Empirically, we observe snow-flake-like noise in the color stripe, which is due to the abrasion of the lens. Therefore, if the lens has abrasion, when the laser is shed into the camera, the diffuse reflection will occur instead of direct reflection, which will result in some areas of the CMOS receiving a stronger laser beam that may cause overexpose at the corresponding pixels in the image.
As it is hard to measure or accurately model the abrasion inside the lens, we assume the horizontal distribution of the noise follows an exponential distribution, and the vertical distribution of the noise follows a normal distribution.
For laser irradiated from the left, we model the horizontal distribution of the noise as:
\begin{equation*}
    h(y) = \frac{e^{-\beta_1 y/w}}{\int_0^\infty e^{-\beta_1 t/w} dt}
\end{equation*}
\par Here, $\beta_1$ is the hyper-parameter that controls the decaying rate of the distribution, and the denominator is a standardized function to ensure the sum of the probability density function is 1. If the incidence direction is right, then we only need to replace $y$ by $w-y$.

The vertical distribution of the noise follows:
\begin{equation*}
    v(x) = Norm\left(\mu=x_0+\frac{h}{2},\sigma = \frac{h}{\beta_2}\right)
\end{equation*}
\par Here, $\beta_2$ is the hyper-parameter that controls the distribution's sparsity. The vertical distribution is identical for both incidence directions.
\par Furthermore, as the noise does not always result in overflow of the target color channel, we randomly choose a value between 240 and 255 to replace the channel's value. 
The procedure of adding noise is detailed in Appendix~\ref{sec:appendix_lasereffect}.
The effect of incidence direction and noise emulation is shown in Fig.~\ref{fig: Noise}, which highly resembles the real scenario in Fig.~\ref{fig:reallaser}.

\begin{figure}[t]
    \centering
    \subfigure[Red laser (wide stripe)]{
        \centering
        \includegraphics[width=0.22\textwidth]{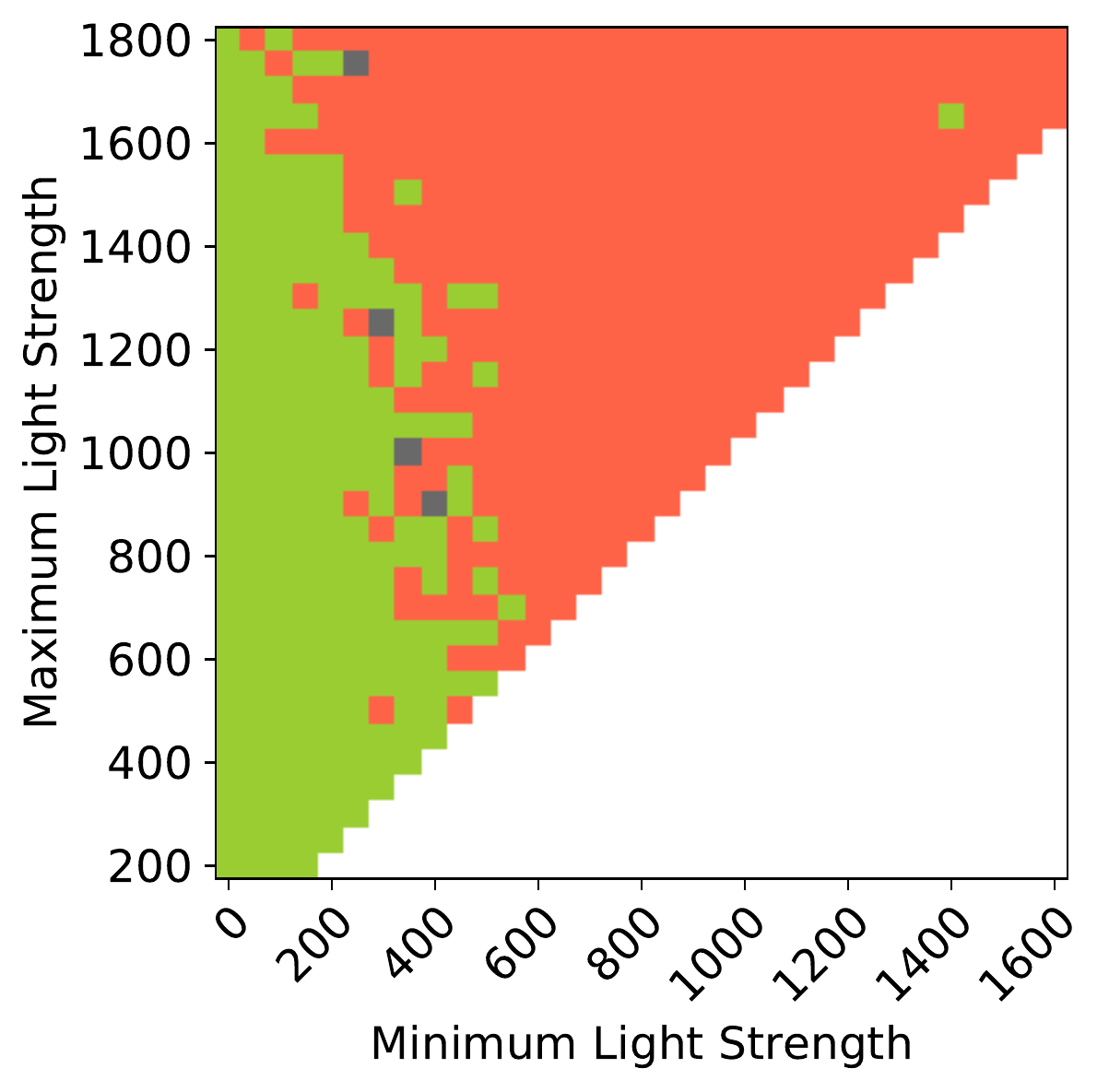}
    }
    \subfigure[Green laser (wide stripe)]{
        \centering
        \includegraphics[width=0.22\textwidth]{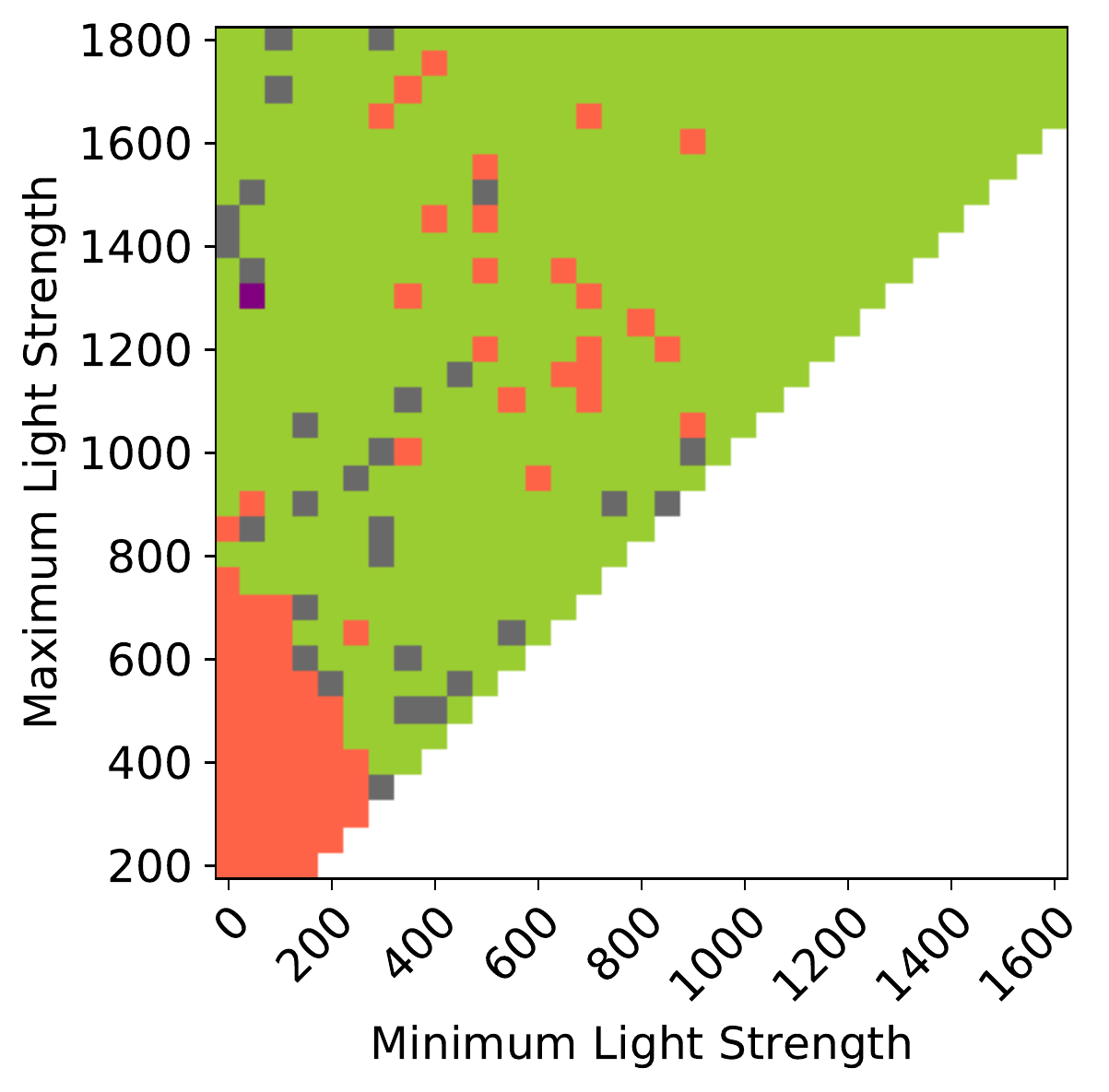}
    }
    \vspace{-0.1in}
    \caption{Refined grid search of light strength for red and green lasers that create wide stripes. The feasible solutions are red and green pixels, respectively.}
    \vspace{-0.05in}
    \label{fig: Refined wide}
\end{figure}

\subsection{Searching Laser Attack Parameters}
\label{sec:search}

Based on the laser attack modeling, we can simulate a large number of color stripes added on a target traffic light using various combinations of laser parameters. 
To acquire the most effective range of parameter sets, we apply the grid-search method. 
We do not use the optimization methods that are common in generating adversarial examples because our attack can succeed with a range of laser parameters instead of a particular set. This property may mitigate the influence of model inaccuracy and real-world uncertainties, therefore increasing the attack's robustness.
The parameters we consider in the grid-search include the minimum light strength $I_{min}$, maximum light strength $I_{max}$, wavelength of the laser $\lambda$, and incidence direction. The noise function follows the incidence direction.
The search grids are reported in Appendix~\ref{sec:appendix_searchresults}.

We tested the parameter search on red and green traffic light images.
The search results show a range of effective laser attack parameters (\textbf{feasible solutions}).
9.63\% and 22.59\% of all parameter combinations are effective for red laser attack and green laser attack, respectively, which proves that the feasible solutions of our attack are not unique.
In addition, we conduct a more thorough search on the minimum and maximum light strength using a more fine-grained search grid. Fig.~\ref{fig: Refined wide} shows the results for red and green stripes with various combinations of minimum and maximum light strengths. 
In the figures, the recognition results of red, yellow, green, black, and DoS (failed to detect the light) are represented by red, yellow, green, dark gray, and purple pixels, respectively. The results show that feasible solutions for light strength exist in a wide range and can be regarded as nearly continuous.

\subsection{Generating Laser Attack Signals}

After acquiring the feasible attack parameters, we map them to adjustable laser parameters to recurrent the model-generated images with real laser.
We mainly consider the wavelength, pulse width and period, power, and incidence angle.


\textbf{Wavelength.}
The relationship between wavelength in the model, $\lambda$, and reality is explicit. The wavelength of the laser diode is the same as the wavelength in the model.

\textbf{Pulse Width and Period.}
The pulse width determines the width of the color stripe in the image. Suppose the image has $N$ rows of pixels and the traffic light takes $n$ rows. Our previous observations in the real-world experiment and emulation both suggest that the color stripe affects traffic light detection the least if its width is larger than that of the light. Therefore, we consider a color stripe with a width of $1.5n$ rows. Suppose the rolling shutter speed is $f$, then the pulse width is $1.5n/Nf$.
The pulse period is set the same as the rolling shutter time $1/f$ to stabilize the stripe's position.


\textbf{Power.}
Power is related to the minimum and maximum light strength, $I_{min},I_{max}$ in the model. The minimum/maximum light strength is a constructed parameter in the model which does not have direct physical meaning. Though there is no deterministic transformation between light strength and power, as they both change monotonically, the attacker could measure the light strength with various laser power beforehand in a similar setup and fit the curve.


\textbf{Incidence Angle.}
The incidence angle is determined by the incidence direction and the relationship between the minimum and maximum light strength. The incidence direction provides a definitive guide on where the laser shall be radiated from. However, it is hard to acquire the incidence angle from the min./max. light strength mathematically because their relationship varies case-by-case for various cameras. Similarly, the attacker can derive their relationship empirically.

Despite the challenges in obtaining accurate mappings between model parameters and laser parameters, the wide range of feasible solutions for our attack can alleviate this problem, as we will show in the next section.
The necessity of the above design in comparison with a naive attacker that subjectively adjusts the attack parameters is discussed in Appendix~\ref{sec:appendix_necessity}.


\begin{figure*}[t]
    \centering
    \subfigure[Impact of the traffic light's size]{
        \includegraphics[width=0.32\textwidth]{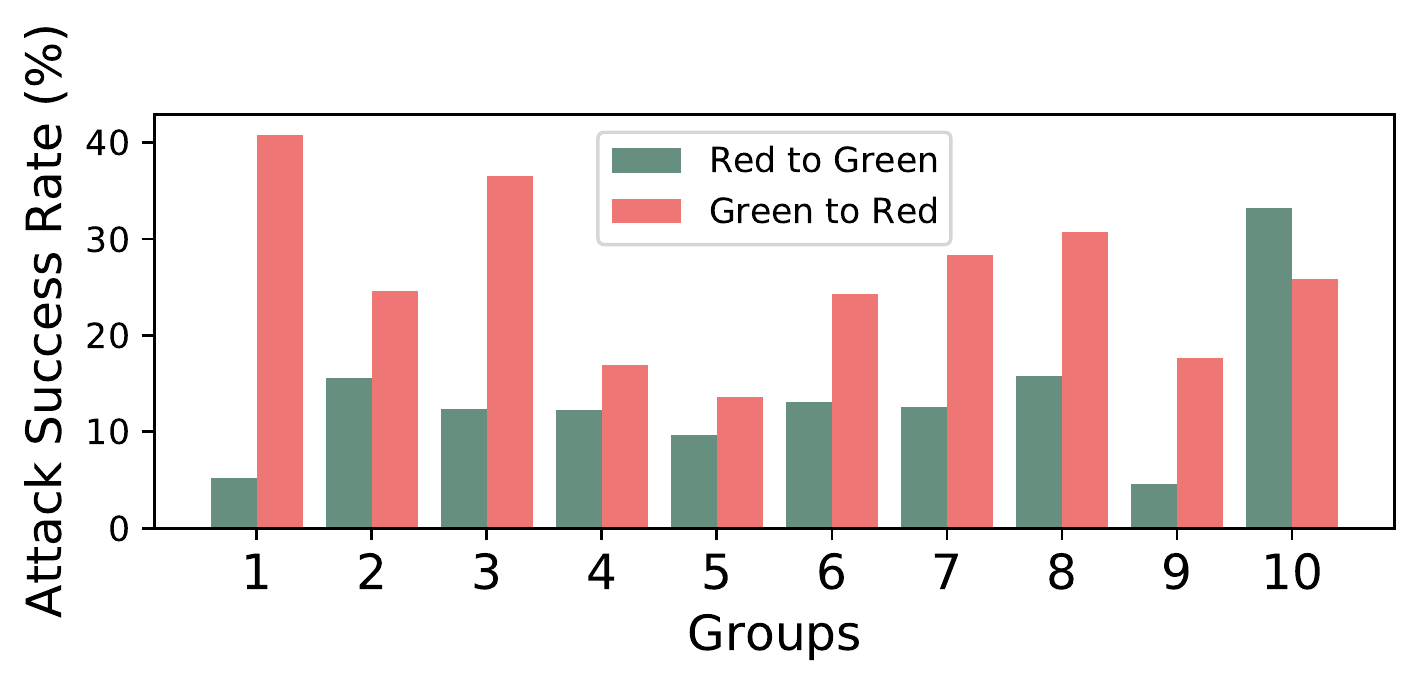}
        \label{fig:lightsize}
    }
    \subfigure[Impact of the traffic light's horizontal position]{
        \includegraphics[width=0.32\textwidth]{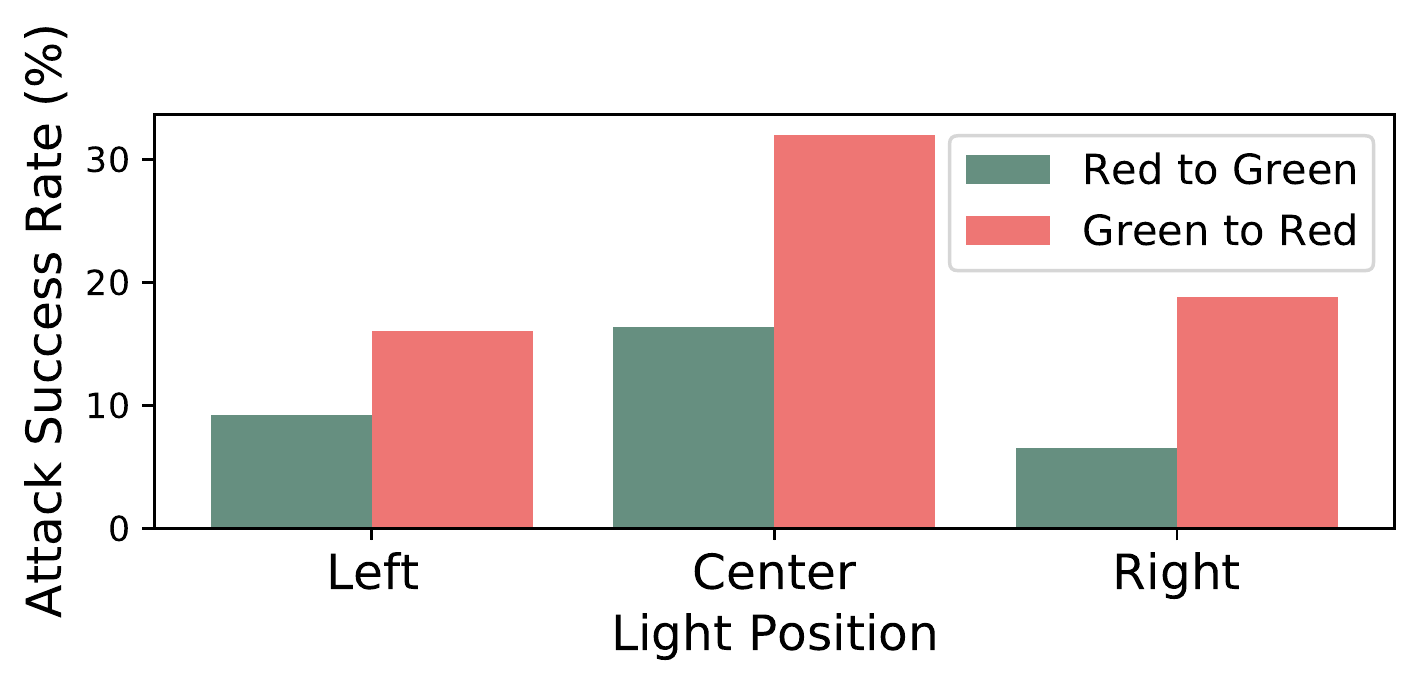}
        \label{fig:lightposition_horizontal}
    }
    \subfigure[Impact of the traffic light's vertical position]{
        \includegraphics[width=0.32\textwidth]{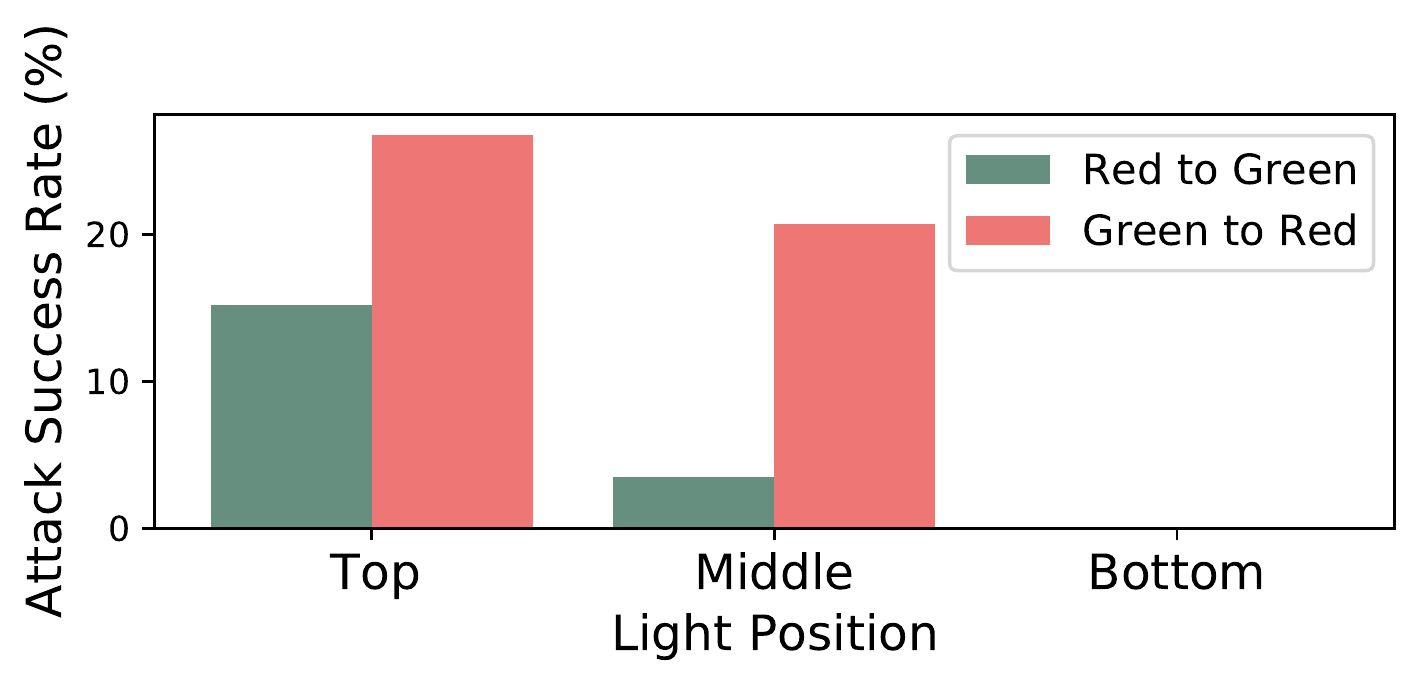}
        \label{fig:lightposition_vertical}
    }
    \vspace{-0.2in}
    \caption{Impact of the traffic light's size and position on the attack's success rates.}
    \vspace{-0.1in}
    \label{fig:trafficlight}
\end{figure*}

\section{Evaluation}

We evaluate the attack's performance in 3 scenarios: (a) emulated attacks based on public traffic light images and simulated laser interference, (b) real-world attacks in stationary setups using real laser and cameras, and (c) real-world attacks against a moving vehicle in practical settings.






\subsection{Emulated Attacks}
\subsubsection{Dataset and Experimental Setup}
We select 240 images that contain traffic lights from the BDD100K dataset~\cite{yu2020bdd100k}. We target all traffic lights in each image that can be correctly detected and recognized by Apollo. In total, there are 250 green lights and 248 red lights. 
The hyper-parameters of the model are determined as follows. For all green lights, we use the red laser with a wavelength of 650~nm; for all red lights, we use the green laser with a wavelength of 520~nm. The laser wavelengths correspond to the laser diodes in real-world experiments. 
Based on our previous observations, we adopt an incidence direction opposite to the light's position in the image to avoid over-exposure in the light's area and use the linear function to model the intensity distribution.
For the minimum and maximum light strength, we applied a grid search for each image. For each variable, we set 7 levels in total: the minimum light strength varies from 100 to 700 with a step length of 100, and the maximum light strength varies from 900 to 1500 with a step length of 100.

\subsubsection{Overall Performance}
\begin{figure}[t]
    \centering
    \subfigure[R$\rightarrow$G attacks]{
    \hspace{-0.3in}
        \begin{minipage}[t]{0.45\linewidth}
        \centering
        \includegraphics[scale=0.32]{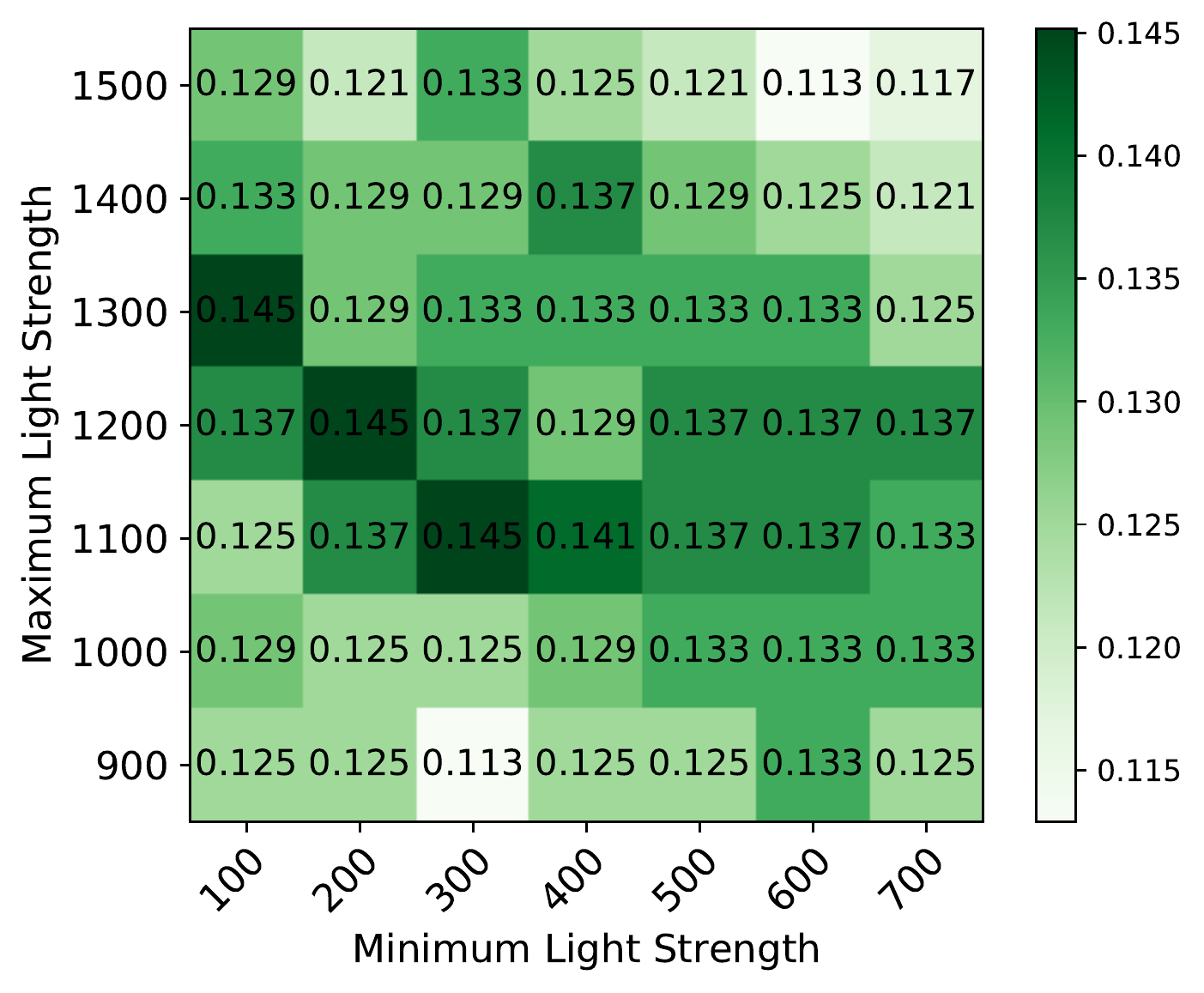}
        \end{minipage}
    }
    \subfigure[G$\rightarrow$R attacks]{
        \begin{minipage}[t]{0.45\linewidth}
        \centering
        \includegraphics[scale=0.32]{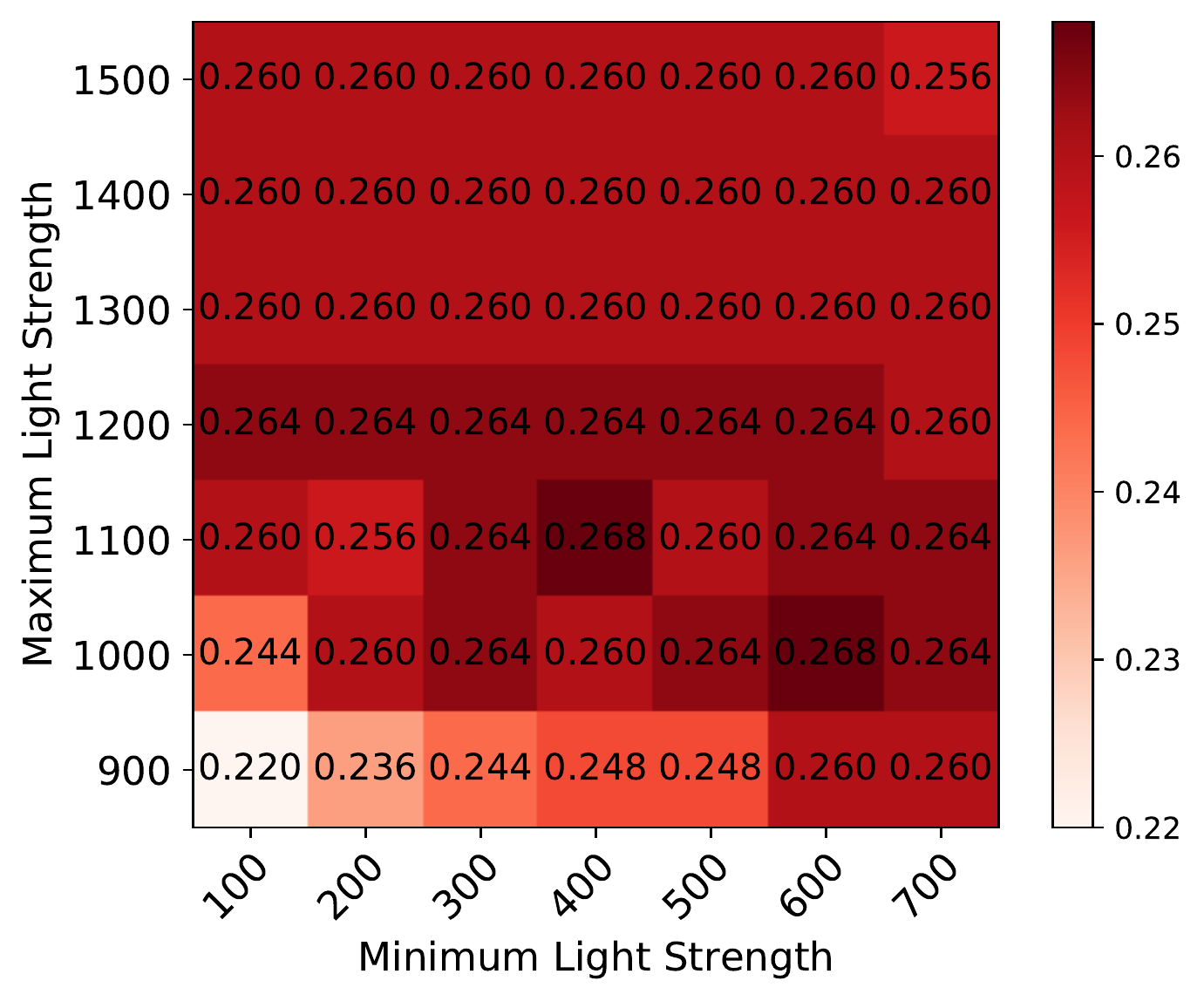}
        \end{minipage}
    }
    \vspace{-0.1in}
    \caption{The attack's average success rates on all images using various combinations of min. and max. light strengths.}
    \vspace{-0.1in}
    \label{fig: Grid Search from Test}
\end{figure}

By using the optimal set of attack parameters for each traffic light, we are able to make a red light recognized as a green light (R$\rightarrow$G) in 45 cases, i.e., with an attack success rate of 18.2\%, and make a green light recognized as a red light (G$\rightarrow$R) in 86 cases, with an attack success rate of 34.4\%.
In addition, we find that the attack can succeed on different images even using the same set of parameters. For example, Fig.~\ref{fig: Grid Search from Test} shows that R$\rightarrow$G is successful on 14.5\% of all images when the min. and max. light strengths are (100,1300), (200,1200), or (300,1100), and G$\rightarrow$R is successful on 26.8\% of all images when the min. and max. light strengths are (400,1100) or (600,1000).
These results show that our attack is feasible on various traffic lights and does not require unique parameters as feasible solutions.

\subsubsection{Impact of the Traffic Lights}
Our dataset contains 489 traffic lights of various sizes and positions in the image. In the following, we investigate whether the traffic light can affect the attack's success rate.

\textbf{Impact of Size.}
To explore the effect of traffic light's size, we sort all traffic lights into 10 groups by their height in the image (in pixels), and calculate the attack success rate for each group.
Group 1 contains the smallest (0\%--10\% of all heights) lights, while group 10 represents the largest (90\%--100\%).
The result is shown in Fig.~\ref{fig:lightsize}.
We observe that for the G$\rightarrow$R cases, the attack success rates are comparatively higher for relatively smaller or larger traffic lights. However, for R$\rightarrow$G cases, there is no clear relationship.
Overall, the correlation between the traffic light's size and the attack's success rate is not significant. 

\textbf{Impact of Position.}
We consider both the horizontal and vertical positions of the traffic light in the image.
We divide an image into three-by-three blocks: left, center, and right horizontally; top, middle, and bottom vertically. All blocks have the same size. We plot the attack's success rates for the horizontal and vertical divisions, respectively, in Fig.~\ref{fig:lightposition_horizontal} and \ref{fig:lightposition_vertical}.
Horizontally, the attack's success rates for both G$\rightarrow$R and R$\rightarrow$G cases are higher when the traffic light is in the center of the image than at the left or right. 
Vertically, the attack's success rate appears to be higher if the traffic light is at a higher position in the image.  Since no image contains a traffic light in the bottom blocks, the attack's success rate is not available. 
In summary, we assume that such impacts of traffic light's position are mainly due to the background.
In most cases, the background will be clearer at the top and center positions of the image, which typically is the sky. In contrast, in other positions of the image, the background is usually buildings or traffic that are more complex. 



\begin{figure}[t]
    \centering
    \subfigure[Illustration of setup]{
        \includegraphics[width=0.225\textwidth]{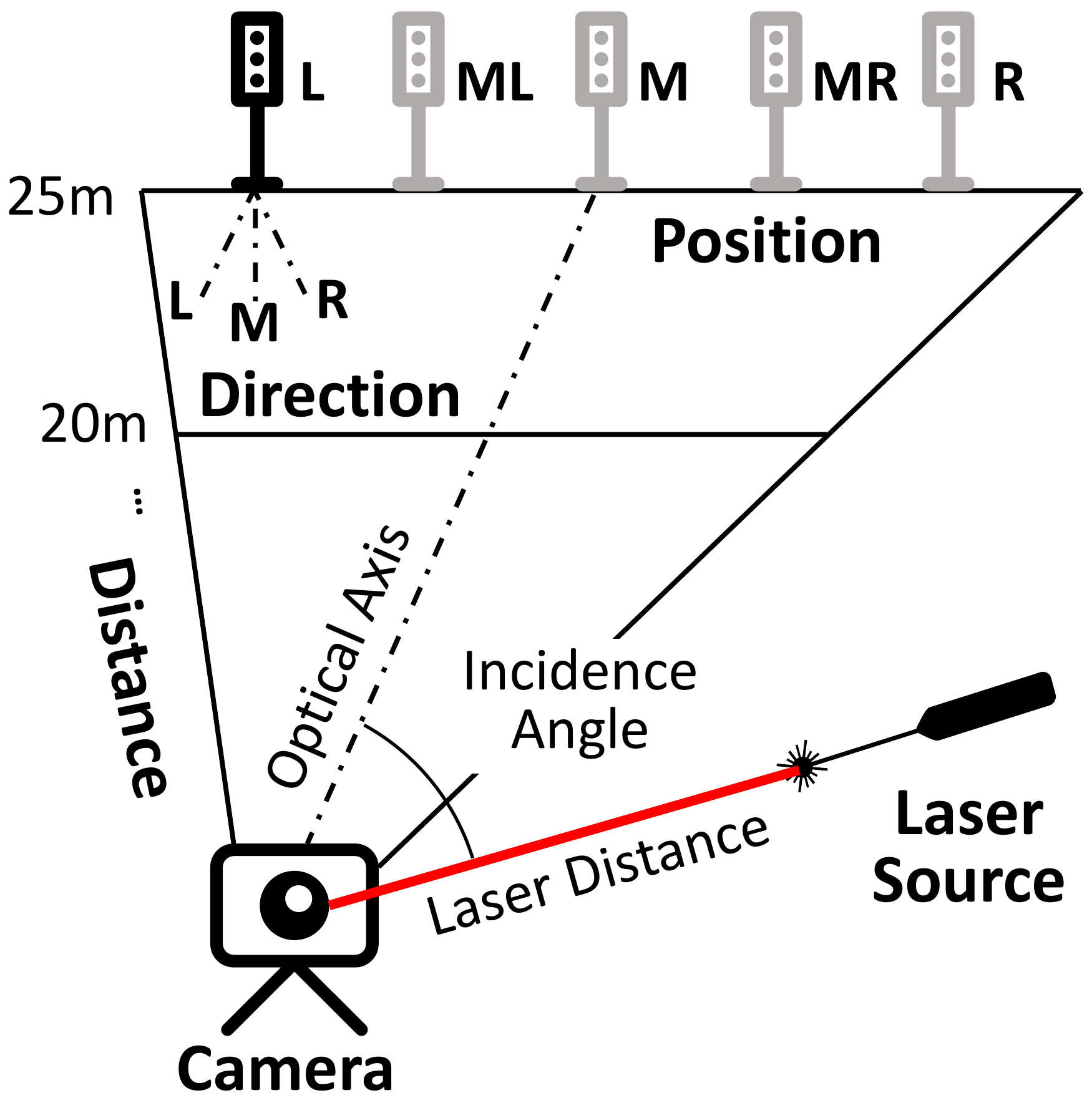}
        \label{fig:setup_illustration}
    }
    \subfigure[Real setup]{
        \includegraphics[width=0.225\textwidth]{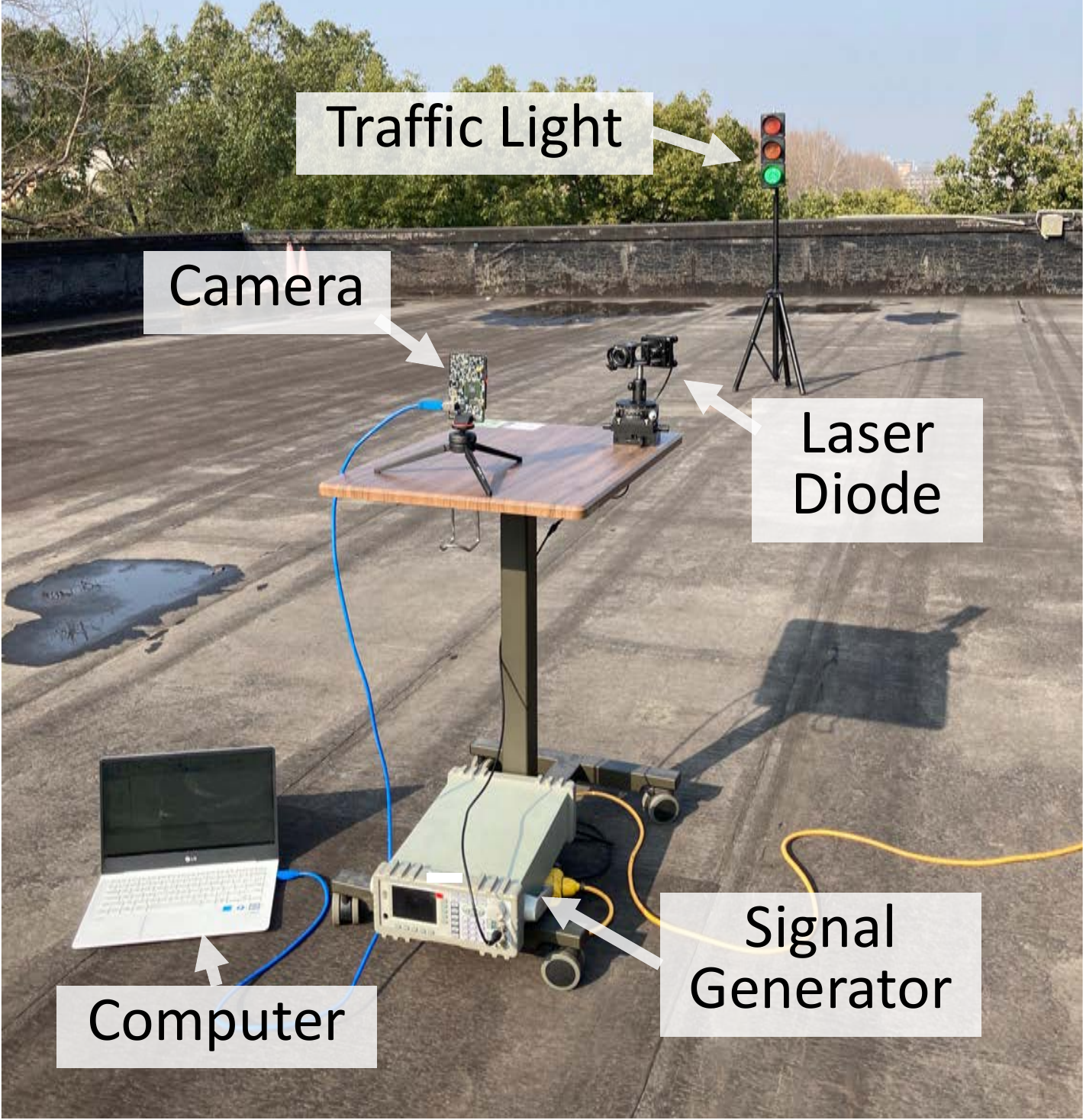}
    }
    \vspace{-0.1in}
    \caption{Illustration and image of the experiment setup.}
    \vspace{-0.1in}
    \label{fig: attack setup}
\end{figure}

\subsection{Real-World Attacks in Stationary Setups}
We validate the attack's effectiveness in stationary setups using real cameras, laser diodes, and traffic lights at various distances, positions, and directions. 

\subsubsection{Experimental Setup}
The experimental setup is shown in Fig.~\ref{fig: attack setup}. 
The attack target is an experimental traffic light that can shine in green, red, and yellow. The diameter of the lamps is 10~cm\footnote{The diameter of lamps on regular traffic lights is 20~cm. Therefore, the traffic light's distance to the camera that we report in this paper shall be doubled so that it is equivalent to the attacks against regular traffic lights.}. We experiment on 5 cameras: an AR0132AT evaluation board camera~\cite{ar0132at}, two dashcams from Xiaomi~\cite{xiaomi} and Hikvision~\cite{hikvision}, an OPPO K3 smartphone~\cite{oppo}, and an OpenMV H7 Devboard camera~\cite{openmv}. The AR0132AT sensor is considered to be the camera used on Tesla vehicles~\cite{tesla1,tesla2}. We attack the cameras using two laser diodes (505~nm for green laser, 650~nm for red laser) driven and controlled by a signal generator. The images of the traffic light under laser interference is sent to two traffic light detection and recognition systems: Apollo~\cite{apollo} and a deep-learning-based model developed for the Nexar traffic light recognition challenge~\cite{nexar}, which are the most popular and state-of-the-art open-source models we can find for traffic light recognition. 
To increase the variety of traffic light instances, we conduct the attack in various combinations of the traffic light's distance, position, and facing direction, as illustrated in Fig.~\ref{fig:setup_illustration}.
The traffic light's distance to the camera in the optical axis varies from 5~m to 15~m. At each distance, the traffic light is placed in 5 positions, left (L), middle left (ML), middle (M), middle right (MR), and right (R). On each position, we set the light to face three directions: left (L), middle (M), and right (R). In total, there are 120 instances of traffic lights for a single color.
We repeat the 120 instances for different light colors and cameras.
For safety reasons, all the experiments are conducted on a rooftop in an enclosed environment with proper laser protection.



\begin{table}[t]
    \centering
    \setlength\tabcolsep{1.4pt} 
    \small
    \vspace{-0.05in}
    \caption{Success rates of attacking 2 systems and 5 cameras.}
    \label{tab:overall_performance}
    \vspace{0.05in}
    \begin{tabular}{|c|c|c|c|c|c|c|c|}
    \hline
    \multirow{2}{*}{Sys.} &
      \multirowcell{2}{Attack\\Scenario} &
      \multicolumn{5}{c|}{Target Camera} &
      \multirow{2}{*}{\begin{tabular}[c]{@{}c@{}}Avg.\end{tabular}} \\ \cline{3-7}
     &
       
       &
      \begin{tabular}[c]{@{}c@{}}Tesla\end{tabular} &
      \begin{tabular}[c]{@{}c@{}}Xiaomi\end{tabular} &
      \begin{tabular}[c]{@{}c@{}}Hikv\end{tabular} &
      \begin{tabular}[c]{@{}c@{}}OPPO\end{tabular} &
      \begin{tabular}[c]{@{}c@{}}OpMV\end{tabular} &
       \\ \hline 
    \multirow{4}{*}{\rotatebox{90}{Apollo}} & R$\rightarrow$G   & 31.67\% & 32.47\% & 15.38\% & 4.44\%  & 17.39\% & 20.27\% \\ \cline{2-8} 
                            &                    R$\rightarrow$DoS & 21.67\% & 40.26\% & 15.38\% & 73.33\% & 13.04\% & 32.74\% \\ \cline{2-8} 
                            & G$\rightarrow$R   & 73.33\% & 17.24\% & 39.13\% & 32.22\% & 47.37\% & 41.86\% \\ \cline{2-8} 
                            &                    G$\rightarrow$DoS & 21.67\% & 32.18\% & 39.13\% & 36.67\% & 52.63\% & 36.46\% \\ \hline
    \multirow{4}{*}{\rotatebox{90}{Nexar}}  & R$\rightarrow$G   & 28.33\% & 14.77\% & 1.23\%  & 59.55\% & 0\%     & 20.78\% \\ \cline{2-8} 
                            &                    R$\rightarrow$DoS & 12.50\% & 57.95\% & 38.27\% & 31.46\% & 100\%   & 48.04\% \\ \cline{2-8} 
                            & G$\rightarrow$R   & 99.17\% & 64.00\% & 21.31\% & 45.21\% & 100\%   & 65.94\% \\ \cline{2-8} 
                            &                    G$\rightarrow$DoS & 0.83\%  & 30.67\% & 78.69\% & 53.42\% & 0\%     & 32.72\% \\ \hline
    \end{tabular}
    \end{table}

\begin{figure}[t]
    \centering
        \includegraphics[width=0.4\textwidth]{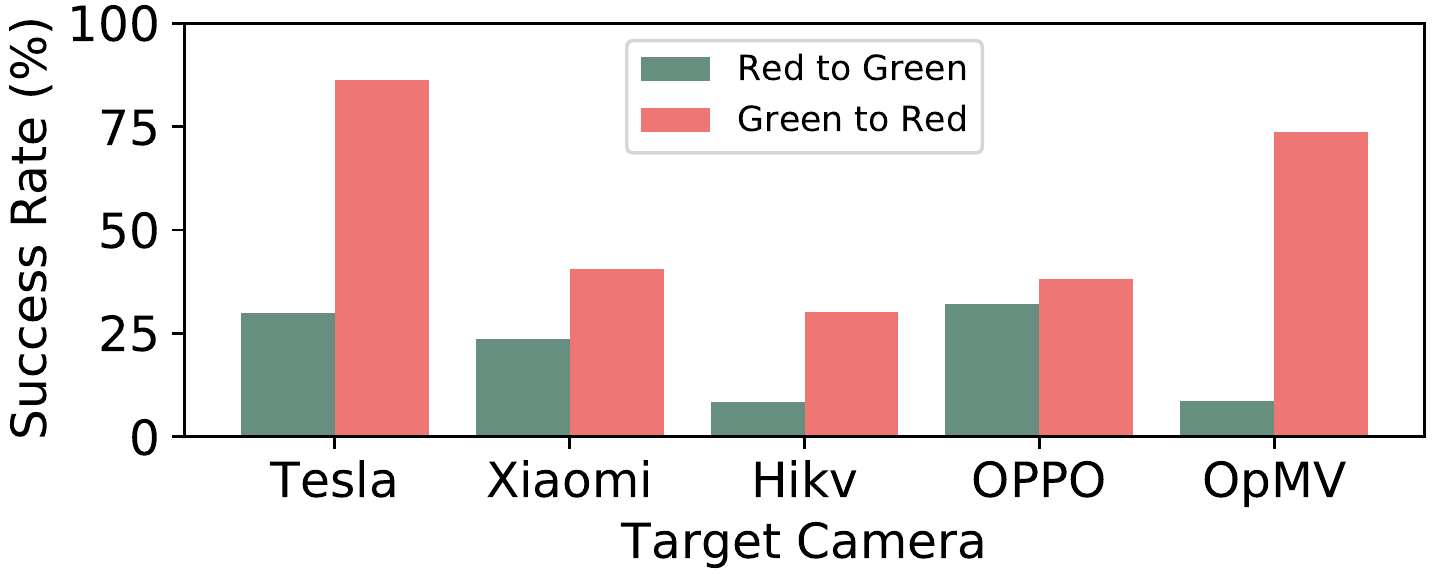}
    \vspace{-0.1in}
    \caption{Average success rates of attacking 5 cameras. }
    \vspace{-0.1in}
    \label{fig:result_5cameras}
\end{figure}

\subsubsection{Overall Performance}
We report the attack's overall success rates in Table~\ref{tab:overall_performance}. The experiments are conducted in two attack scenarios: R$\rightarrow$G and G$\rightarrow$R. We also report the percentage of DoS cases.
The overall success rate is 20.53\% for R$\rightarrow$G attacks and 44.95\% for G$\rightarrow$R attacks, suggesting that it is relatively easier to cause a green light to be recognized as a red light using a red laser. DoS happens 34.59\% and 26.76\% of the time for R$\rightarrow$G and G$\rightarrow$R attacks, respectively.
Note that the above results involve all the instances. The average success rate is higher than 90\% when the distance is less than 10~m (equivalent to 20~m for regular traffic lights, which is the width of a typical four-lane urban road).

\begin{figure*}[t]
    \centering
    \subfigure[No. of successful G$\rightarrow$R at various locations]{
            \includegraphics[width=0.32\textwidth]{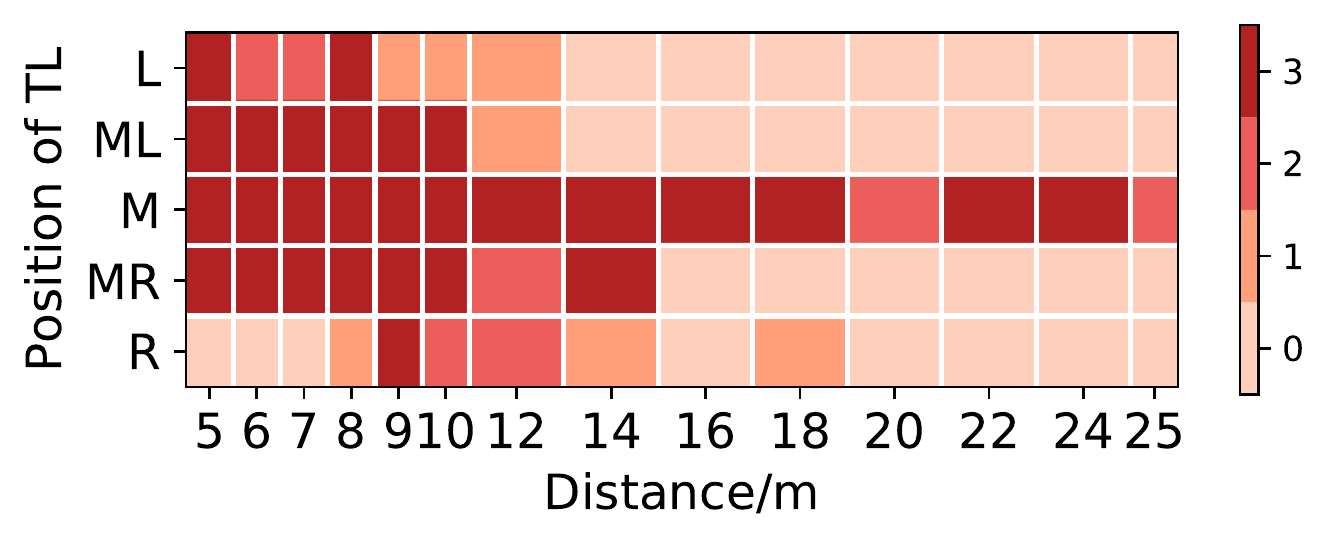}
            \label{fig:result_g2r}
    }
    \subfigure[Results of G$\rightarrow$R attack at various distances]{
            \label{fig:result_g2r_distance}
            \includegraphics[width=0.32\textwidth]{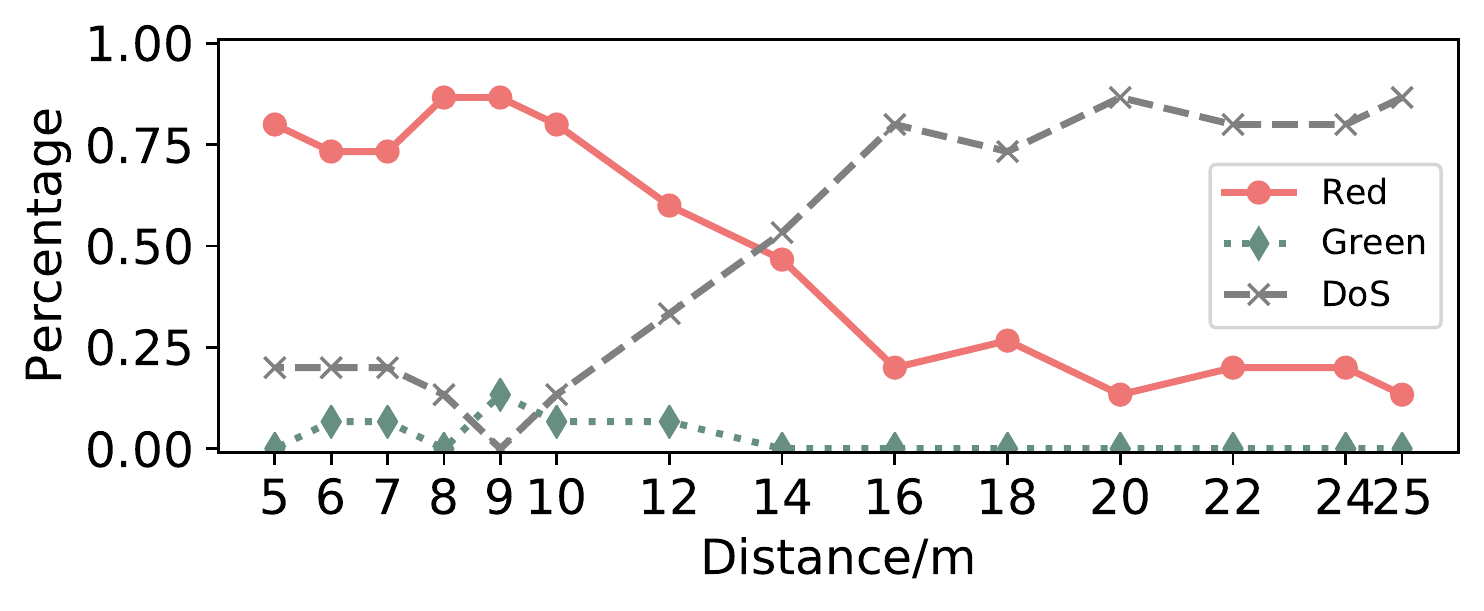}}
    \subfigure[Results of G$\rightarrow$R attack at various positions]{
            \label{fig:result_g2r_position}
            \includegraphics[width=0.32\textwidth]{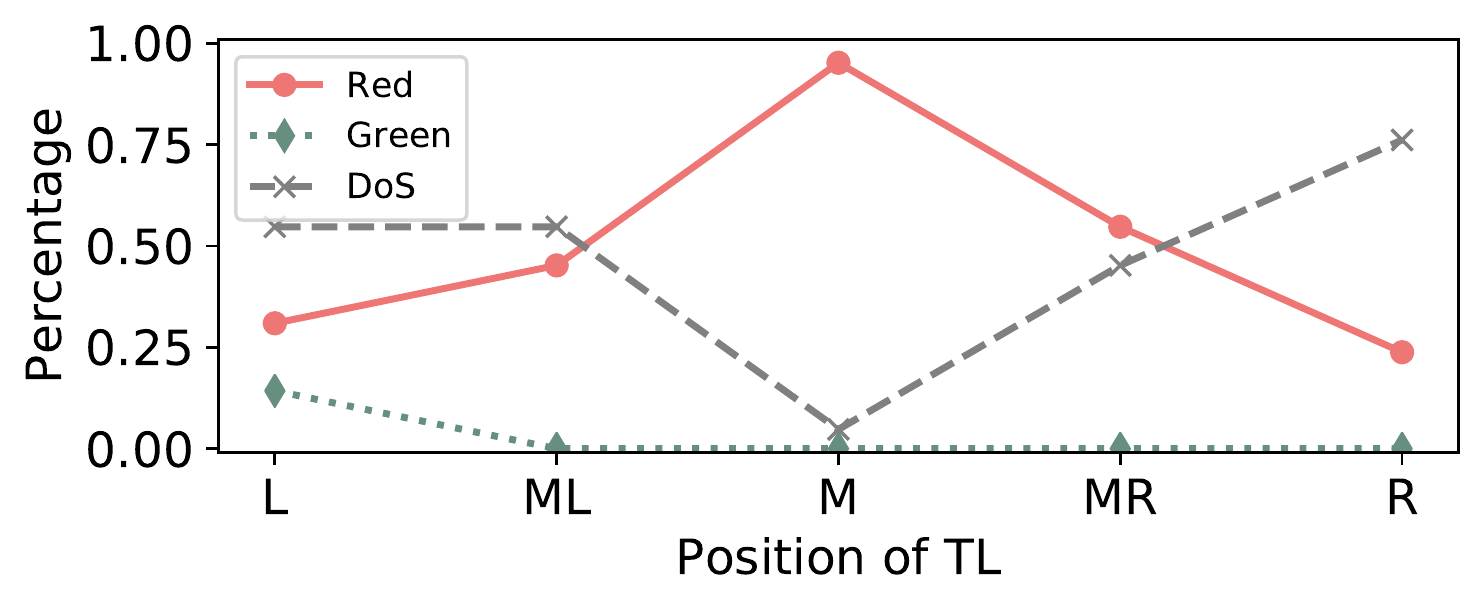}}

    \vspace{-0.05in}
    \subfigure[No. of successful R$\rightarrow$G at various locations]{
            \includegraphics[width=0.32\textwidth]{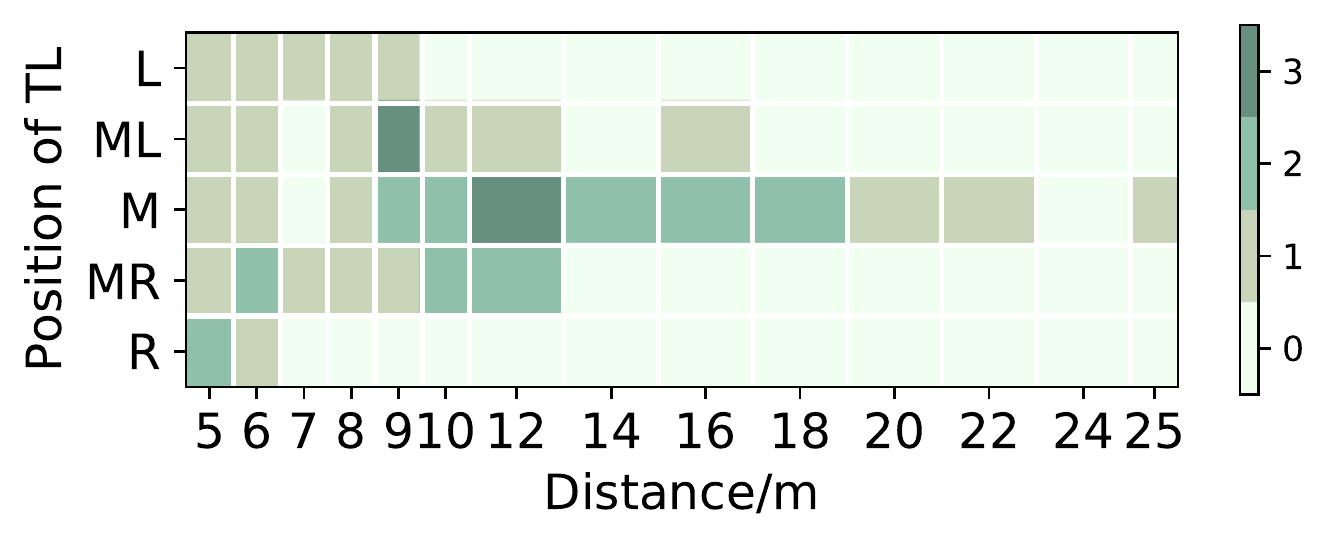}
            \label{fig:result_r2g}
    }
    \subfigure[Results of R$\rightarrow$G attack at various distances]{
            \label{fig:result_r2g_distance}
            \includegraphics[width=0.32\textwidth]{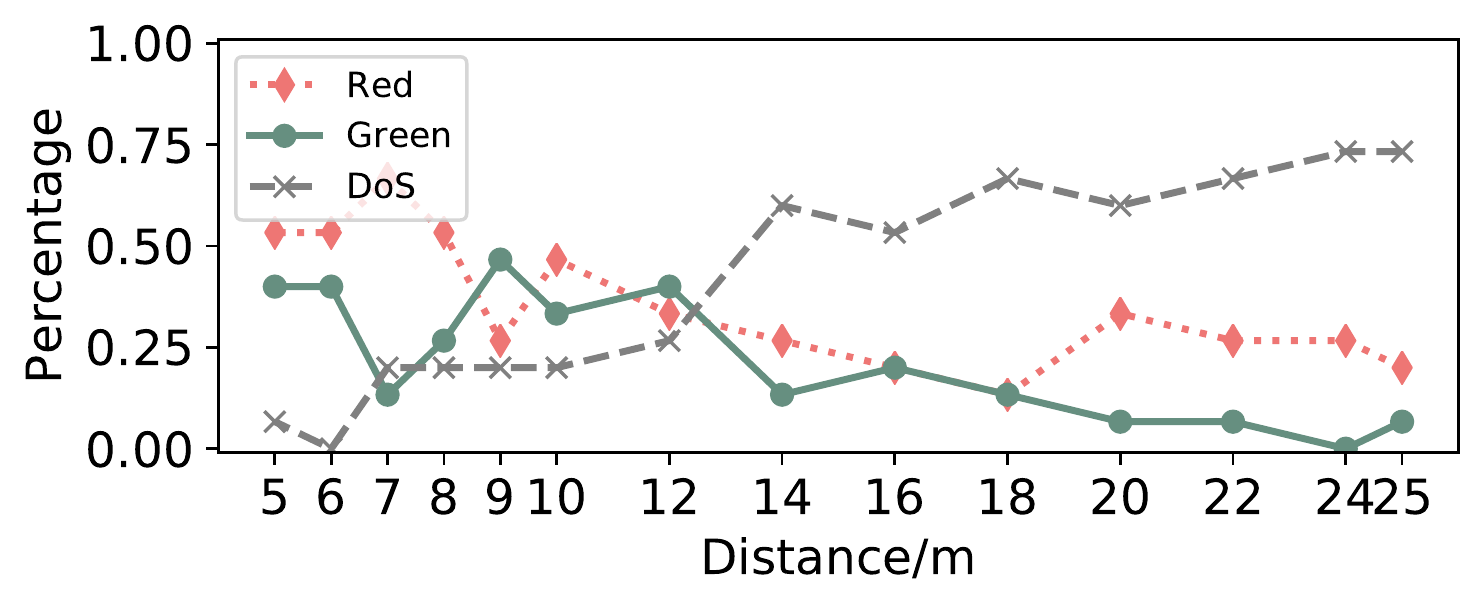}}
    \subfigure[Results of R$\rightarrow$G attack at various positions]{
            \label{fig:result_r2g_position}
            \includegraphics[width=0.32\textwidth]{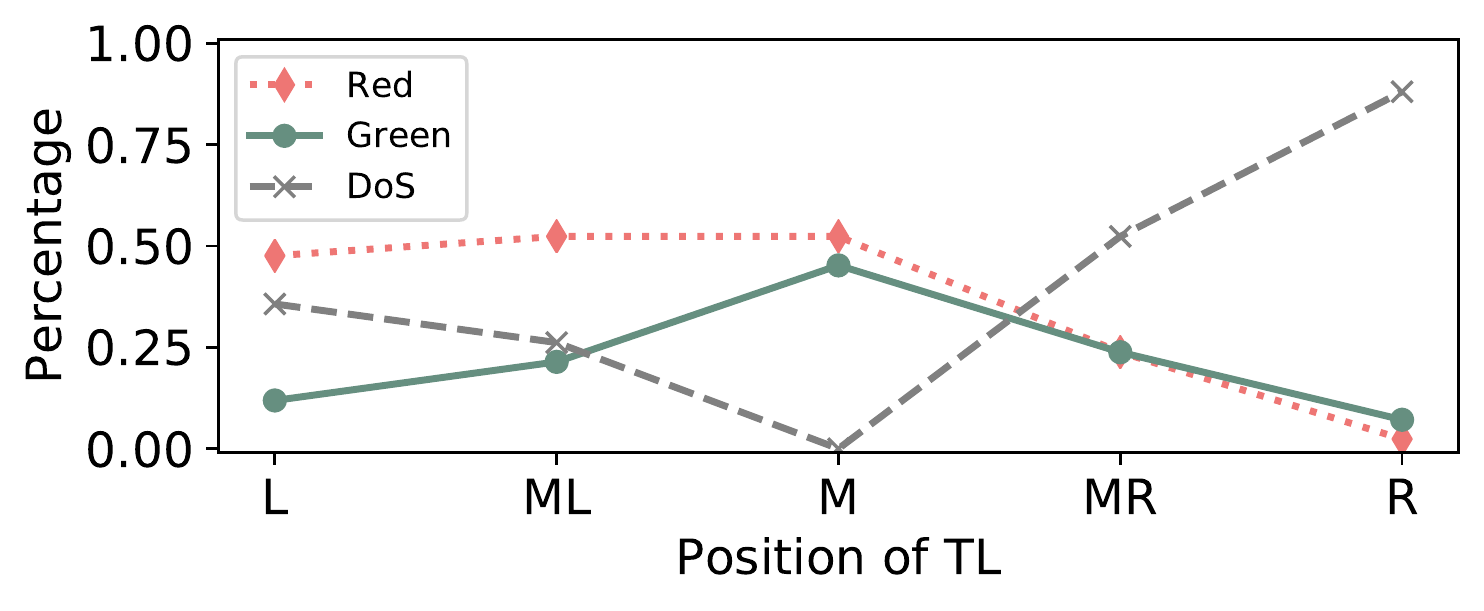}}
    \vspace{-0.1in}
    \caption{Attack results showing impacts of the traffic light's distance and position based on the AR0132AT camera.}
    \vspace{-0.1in}
    \label{fig: g2r TL position and distance}
\end{figure*}


In addition, we find that the attack result varies according to the target recognition system and camera.
The average attack success rate on Apollo is 31.07\%, while it is 43.36\% on Nexar. The comparison of attacking 5 cameras in Fig.~\ref{fig:result_5cameras} shows that the AR0132AT camera used on Tesla is the most vulnerable---the average success rates for R$\rightarrow$G and G$\rightarrow$R are 30\% and 86.25\%.
The lower success rates on other cameras may be caused by the low image qualities or large field-of-view as listed in Table~\ref{tab: spec sheet}.
For example, Xiaomi and Hikvision are both dashcams that have a large field-of-view ($130^{\circ}$) in comparison with the Tesla camera ($70^{\circ}$). The wide angle makes the traffic light smaller in the image and thus more difficult to be recognized. As the traffic light becomes closer and larger, it moves toward the edge of the image and suffers from more wide-angle distortions than others. 
Nonetheless, G$\rightarrow$R attacks show higher success rates than R$\rightarrow$G regardless of the recognition system and camera.

\begin{figure}[t]
    \centering
    \subfigure[Attack angles and distances]{
            \includegraphics[width=0.22\textwidth]{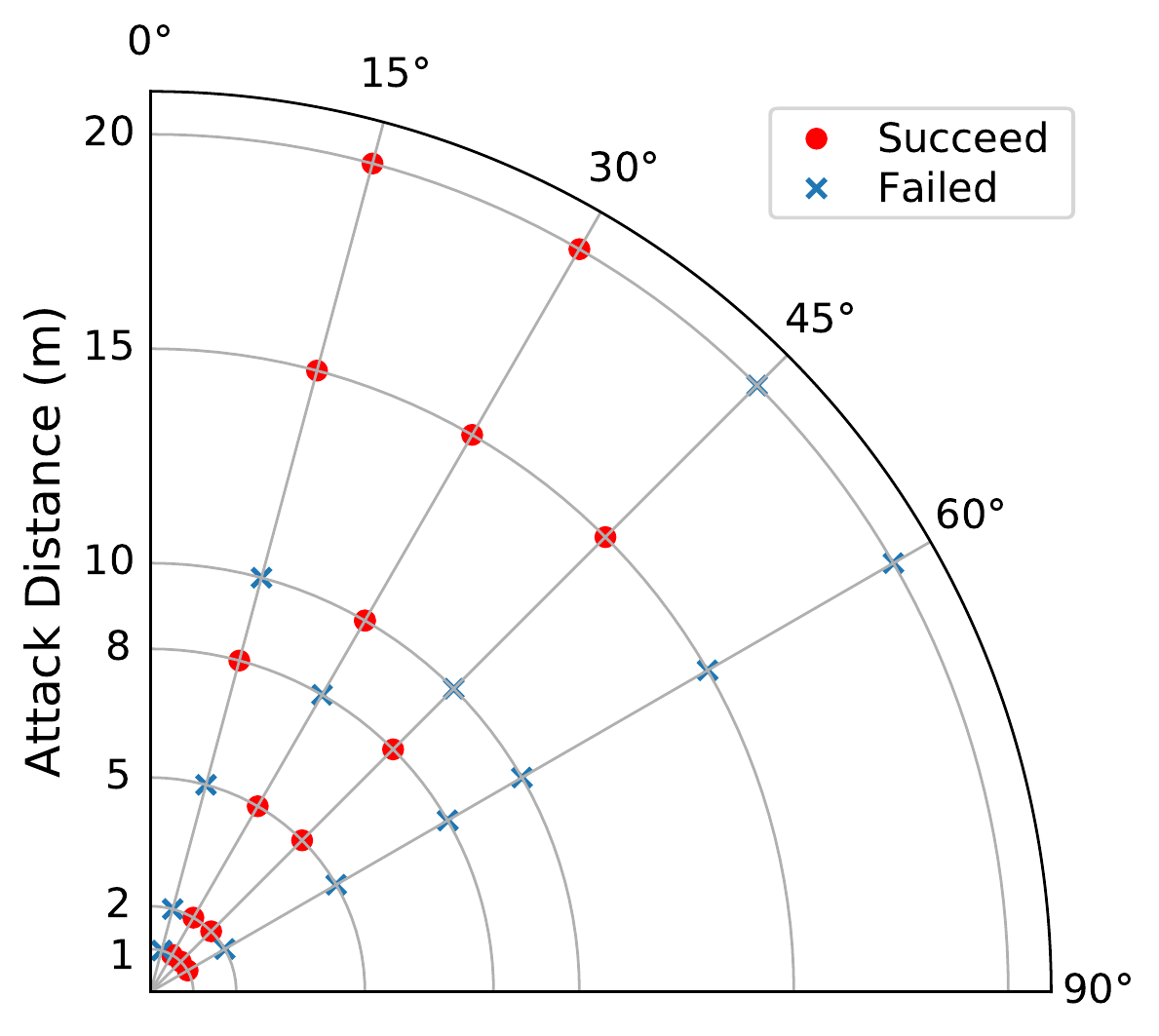}
            \label{fig:distance_angle}
    }
    \subfigure[Attack from 35~m away]{
            \includegraphics[width=0.21\textwidth]{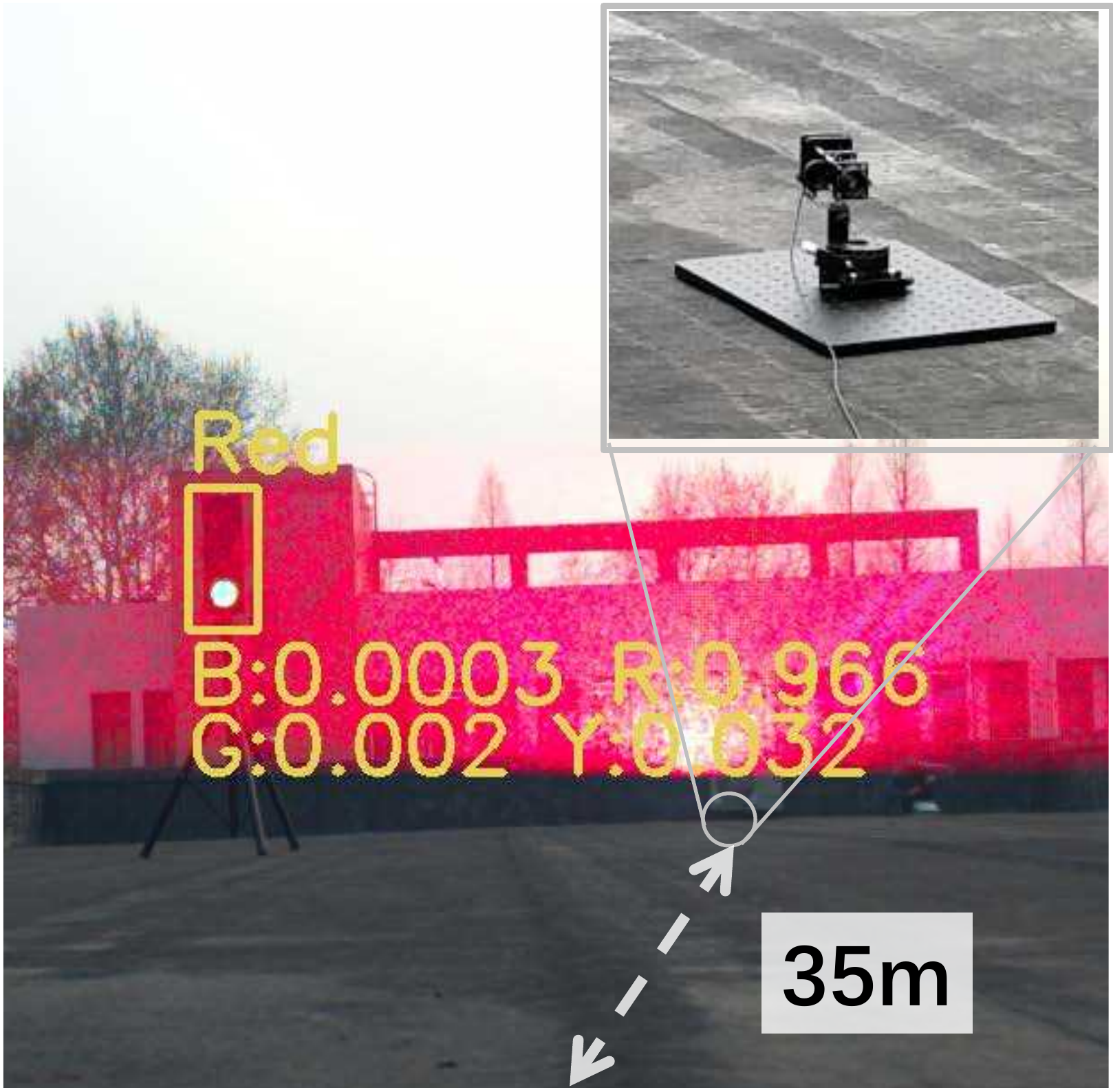}
            \label{fig:longrange}
    }
    \vspace{-0.1in}
    \caption{Results at various attack distances and angles.}
    \vspace{-0.15in}
    \label{fig: laser angle and distance}
\end{figure}

\subsubsection{Impact of the Traffic Light}

We investigate the impacts of traffic light's distance, position, and direction based on the AR0132AT camera. In this experiment, we extend the maximum traffic light distance to 25~m. The number of successful attacks when the traffic light is at various locations is visualized in Fig.~\ref{fig:result_g2r} and \ref{fig:result_r2g}.

\textbf{Impact of Distance.}
The results in Fig.~\ref{fig:result_g2r_distance} and \ref{fig:result_r2g_distance} show that the attack success rate decreases as the traffic light's distance increases, while the percentage of DoS increases. 
We speculate that the traffic light becomes smaller as the distance increases, making it harder to be detected after being overlapped with a color stripe.
However, we can still attack a traffic light that is 25~m away (equivalent to 50~m for regular traffic lights) if it is in the middle position.

\textbf{Impact of Position.}
Fig.~\ref{fig:result_g2r_position} and \ref{fig:result_r2g_position} show that the attack success rate is higher when the traffic light is in the middle of the image. DoS dominates when the traffic light is on the right part of the image because the laser is emitted from the right direction. As the light strength is not strong enough on the left part of the image, the traffic light is recognized as its original color in the left positions. The above observations are consistent with our emulation results.


\textbf{Impact of Direction.}
A successful attack requires the camera to have a clear view of the traffic light's lamp.
Under this condition, we did not find a clear correlation between the attack result and the traffic light's facing direction.


\subsubsection{Attack Distance and Angle}

\begin{figure*}[t]
    \centering
    \subfigure[Setup for continuous video frame experiment]{
    \includegraphics[width=0.345\textwidth]{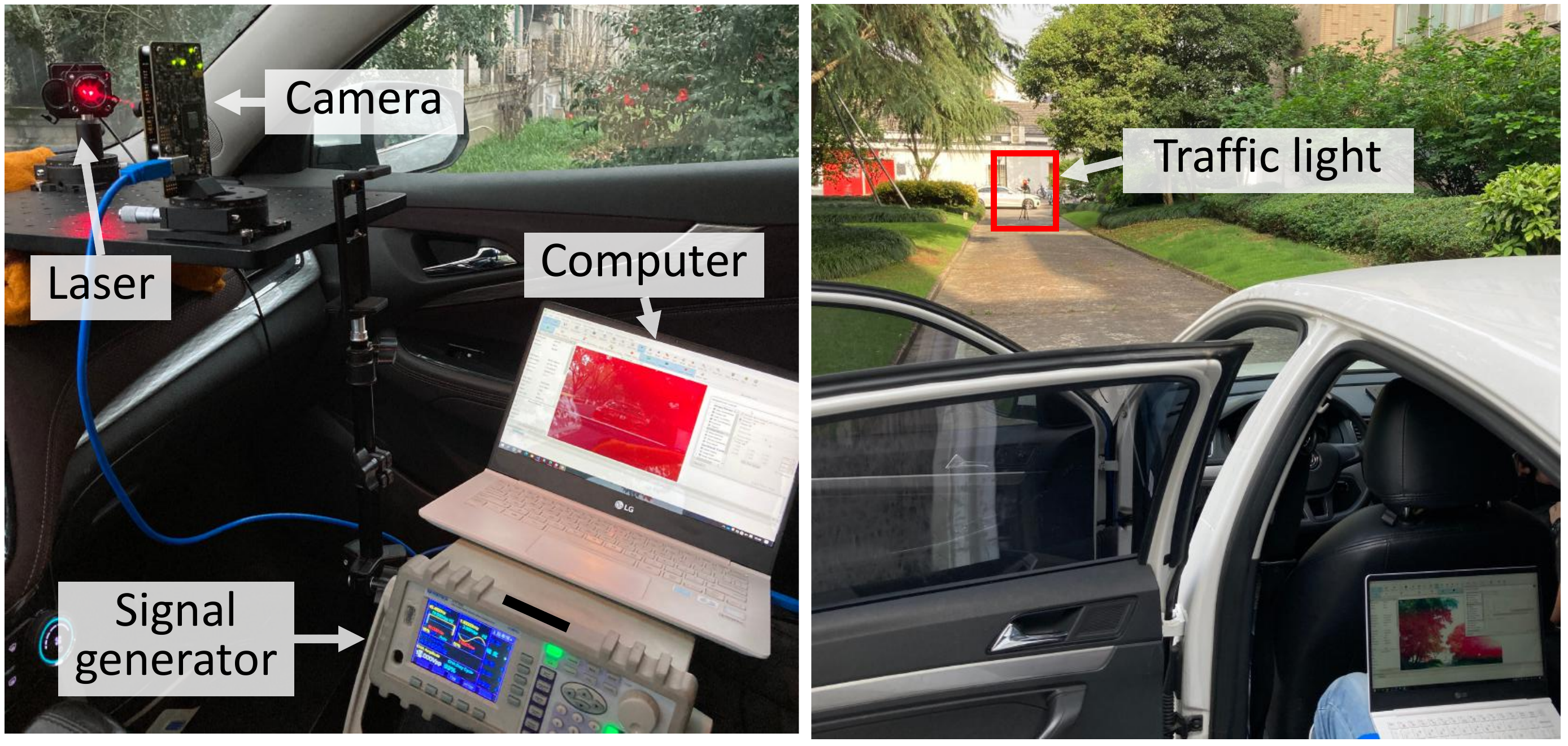}
    \label{fig:setup2}
    }
    \subfigure[Attack equipment for tracking and laser aiming]{
    \includegraphics[width=0.305\textwidth]{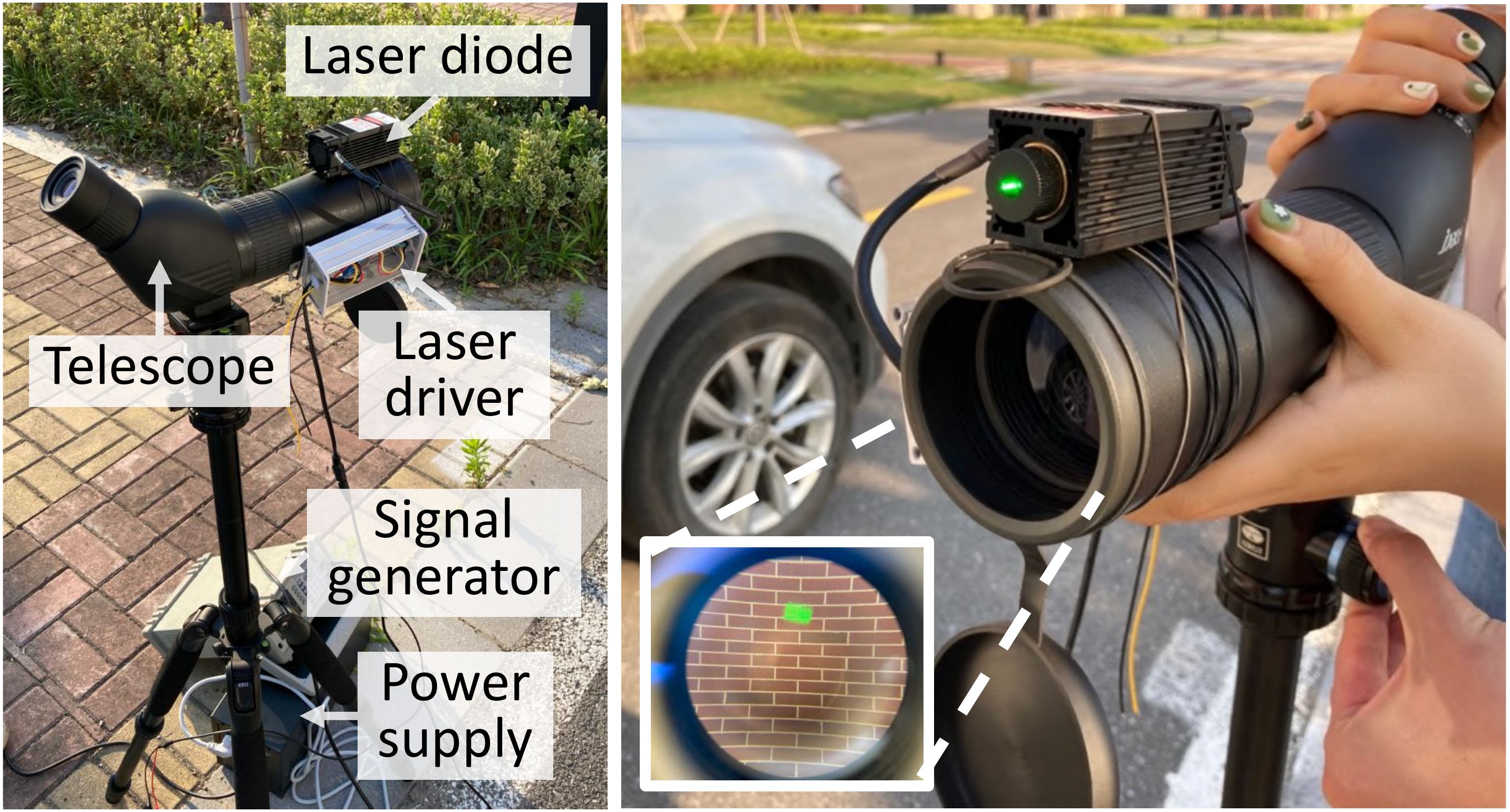}
    \label{fig:equipment-longrange}
    }
    \subfigure[Setup for long-range laser aiming experiment]{
    \includegraphics[width=0.31\textwidth]{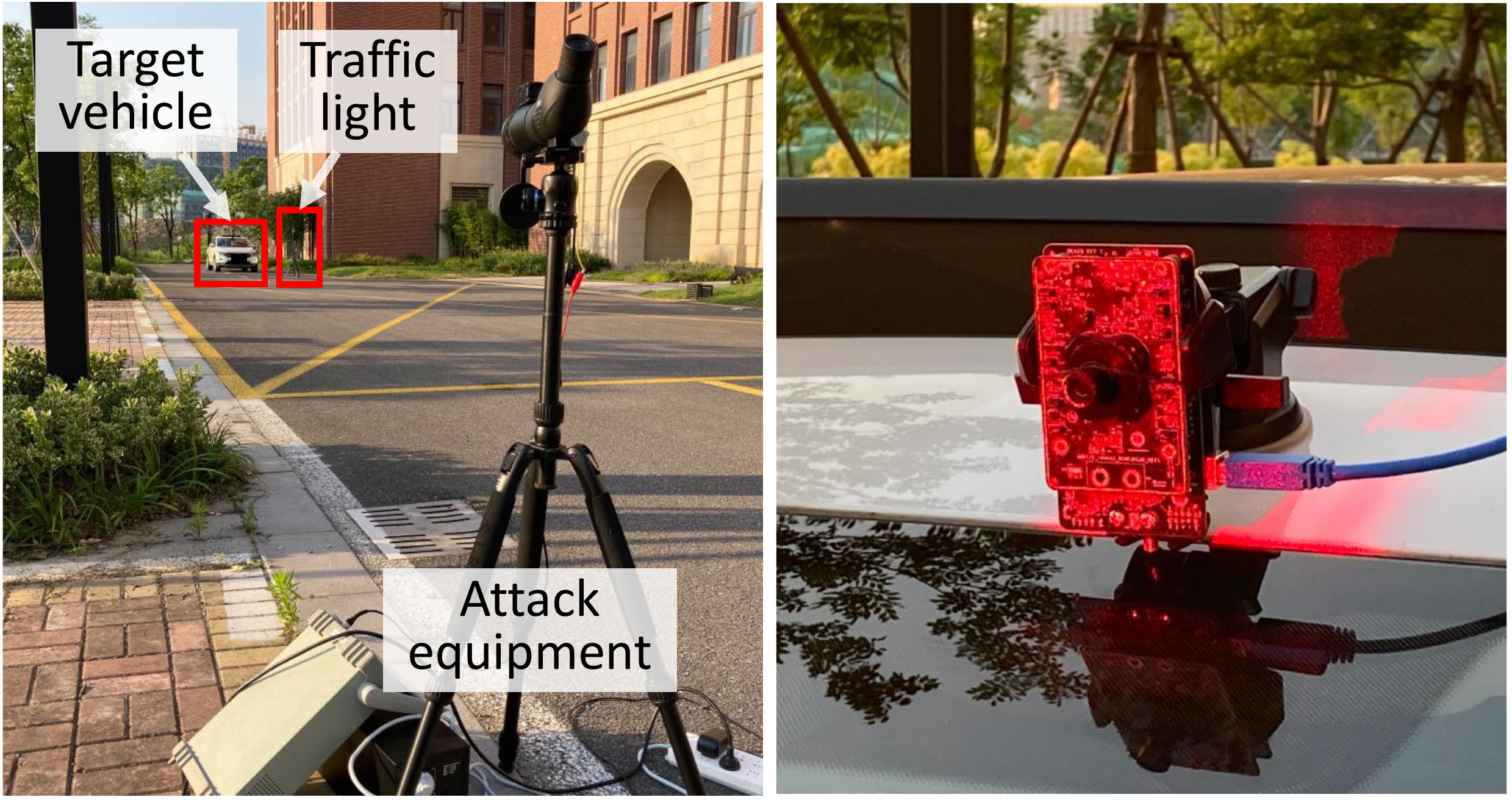}
    \label{fig:setup4}
    }
    \vspace{-0.2in}
    \caption{Experimental setups for real-world attacks:
    (a) the setups inside and outside a vehicle for continuous video frame experiments,
    (b) the self-made equipment for manual target tracking and laser aiming,
    and (c) the setup for long-range laser aiming experiment with the attacker on the roadside and an AR0132AT camera mounted on the target vehicle's windshield.}
    \vspace{-0.1in}
\end{figure*}

We investigate the attack's effectiveness when the laser source is at various distances and off-axis angles. 
We attack the camera from [1, 2, 5, 8, 10, 15, 20] meters away at [15, 30, 45, 60] degrees off the optical axis. The results are shown in Fig.~\ref{fig:distance_angle}. 
It shows that the attack is affected more by the laser's angle than the distance. Successful attacks appear mostly within 45 degrees off-axis from 1~m to 20~m. 
We assume the reason is that the color stripe becomes fainter as the incidence angle increases.
To find the longest attack distance, we move the laser source to the outermost location allowed by our test site (a rooftop), which is approximately 35~m away from the camera. As shown in Fig.~\ref{fig:longrange}, the attack is still successful at such a distance---making a green light recognized as red with a confidence score of 0.966. We envision that the longest attack distance will be larger than 35~m, especially if we use a high-power laser diode.




\subsection{Real-World Attacks in Motion}
Attacks in practical settings require to target a moving vehicle and achieve the following: 
(1) be effective across continuous video frames when the captured scene is changing, 
(2) track and aim laser at the target camera from a distance away from the vehicle, and 
(3) cause end-to-end impact on driving.
We study the attack's capability from the three perspectives separately on a real vehicle and a driving simulator.
Videos of attacks can be found at \url{https://sites.google.com/view/rollingcolors}.

\subsubsection{Effectiveness across Continuous Video Frames}\label{sec:continuous}
First of all, we study the attack’s effectiveness across continuous video frames when the target camera is on a moving vehicle. 
As a preliminary experiment, we rule out the influence of aiming and assume an ideal setting, where the attacker can \textit{continuously inject laser into the camera} when the vehicle is moving.
The feasibility of continuous laser injection will be studied separately.

\textbf{Setup.}
To achieve this setting, we placed the laser diode inside the vehicle as shown in Fig.~\ref{fig:setup2}. 
We performed 27 attack trials in total.
In each trial, we drove the vehicle backward or forward (with speeds between 5-20 km/h) towards an experimental traffic light 30~m ahead and recorded videos with the camera while launching the attack. 
For safety considerations, all experiments were performed in enclosed fields. 

\begin{figure}[t]
    \centering
        \includegraphics[width=0.37\textwidth]{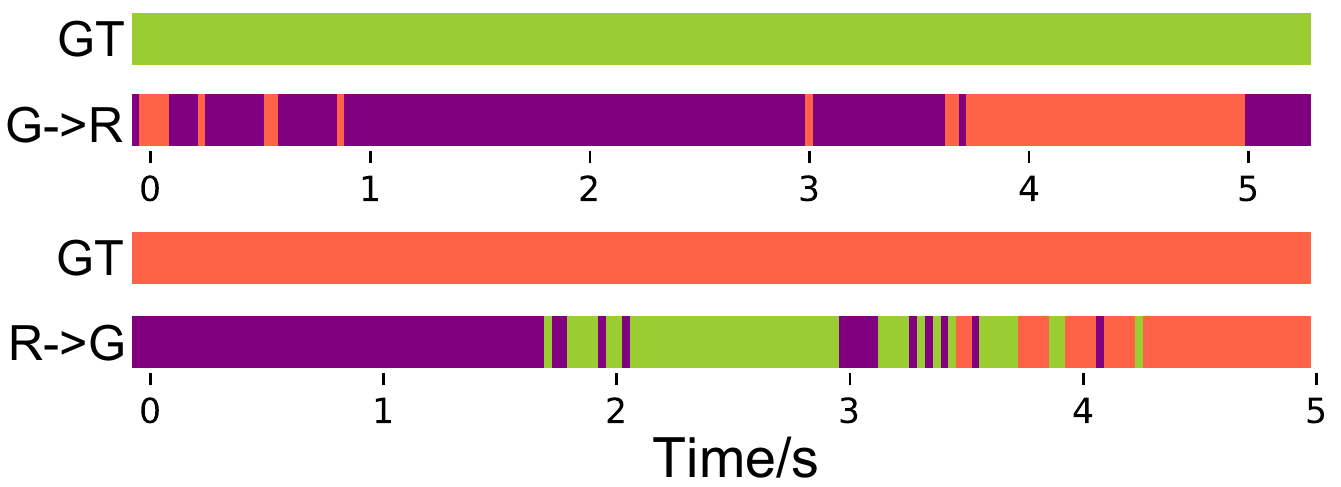}
    \vspace{-0.1in}
    \caption{Attack results across continuous frames, showing continuous success for more than 1 second for both cases. }
    \vspace{-0.15in}
    \label{fig:continuous}
\end{figure}

\textbf{Results.} We analyzed the attack results based on Apollo and found that the attack could successfully spoof traffic light recognition across continuous frames in 23 trials, rendering an attack success rate of 85.2\%.
Fig.~\ref{fig:continuous} shows two best cases for G$\rightarrow$R and R$\rightarrow$G attacks, which achieve the longest duration of successful spoofing.
The recognition results are represented in the corresponding colors (except for purple, which means Apollo fails to detect any traffic light), and ``GT'' represents the ground truth of the light color. The results show that the attack can continuously spoof traffic light recognition for more than one second.
This preliminary experiment demonstrates the attack's effectiveness across continuous video frames when the captured scene is changing.

\subsubsection{Feasibility of Tracking and Laser Aiming}\label{sec:aiming}
In the above experiment, we assume the attacker can continuously inject laser into the camera.
However, in practice, the attacker and laser diode should be outside the target vehicle, which makes it challenging to continuously aim laser at the camera’s tiny lens especially from far away.

\textbf{Manual Tracking and Aiming Equipment.} 
To overcome this challenge, we built a manual tracking and aiming equipment that combined a 500-1000 mW laser diode with a monocular telescope mounted on a tripod, as shown in Fig.~\ref{fig:equipment-longrange}. 
We used the high-power laser diode to create a larger speckle that eases aiming while ensuring the light injected into the camera has sufficient strength. 
We fixed the laser diode on top of the telescope in a position where the speckle is visible from the eye lens of the telescope. 
With this equipment, an attacker can manually track the target and aim the laser at the same time even from far away.




\textbf{Setup.}
Based on the equipment, we conducted real-world experiments in a practical setup to examine the feasibility of continuous aiming.
As shown in Fig.~\ref{fig:setup4}, the attacker was on the roadside and 40-80~m away from the vehicle. 
The target camera was mounted near the top of the vehicle's windshield. 
We drove the vehicle forward or backward towards a traffic light, while another researcher acted as the attacker and controlled the attack equipment on the roadside.
All laser operations were performed in an enclosed field following safety regulations and coordinated with the safety committee\footnote{According to the FDA, laser pointers with power more than 500~mW are Class-4 laser products that must avoid direct or scattered exposure to for eye and skin protection. Though the large speckle we used has greatly reduced the laser's power density at the target vehicle, eye damage may happen to the passengers when viewed directly for long periods of time.
For example, given that the laser power is 1~W, speckle diameters at the diode and at 40~m are 0.8~cm and 15~cm, then based on a rough calculation, the laser's power density is equivalent to 2.9~mW, 11.4~mW, and 213.3~mW at 40~m, 20~m, and 10~m away respectively. Safety measures such as laser goggles must be ensured in all experiments, especially when in proximity to the diode.
}.

\textbf{Results.}
Out of the 88 trials conducted, we found the attacker can continuously inject laser into the camera in 77 trials, rendering an aiming success rate of 89.8\%.
The results show that manual aiming is feasible with the equipment we built even when the vehicle is moving at 20 km/h.


\subsubsection{End-to-End Impact on Driving}
We believe the experimental setup in Section~\ref{sec:aiming} represents a practical setting for attacks in the real world. Thus, we further analyze the experiment's result on traffic light recognition
and investigate the attack's potential end-to-end impact on Apollo's driving behavior with the LGSVL simulator.

\textbf{Result of Attacking TL Recognition.}
Under the practical setup in Section~\ref{sec:aiming}, the attack can successfully spoof a red light to green in 16 trials (out of 38) and spoof a green light to red in 9 trials (out of 50), rendering an attack success rate of 42.1\% for R$\rightarrow$G and 18\% for G$\rightarrow$R.
The average attack success rate is 28.4\%, which is lower than the result in Section~\ref{sec:continuous} due to the unstable laser injection caused by manual aiming from a distance.
We envision that the attack success rate can be increased with better equipment, e.g., an automatic aiming and laser stabilization system.

\begin{figure}[t]
    \centering
    \subfigure[Running a red light]{
        \includegraphics[width=0.22\textwidth]{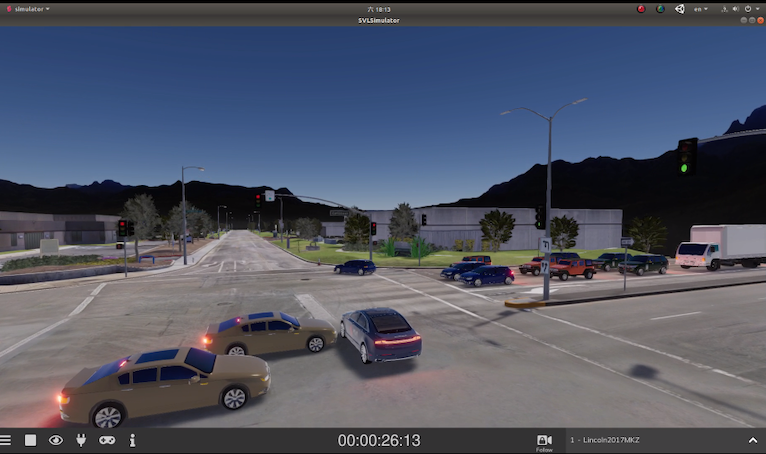}
    }
    \subfigure[Emergency stop]{
        \includegraphics[width=0.225\textwidth]{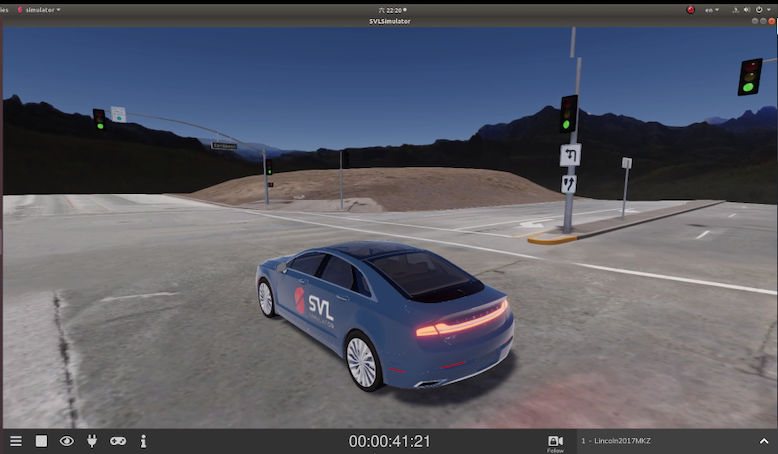}
    }
    \vspace{-0.1in}
    \caption{Simulated end-to-end impacts of the attack.}
    \label{fig:lgsvl}
    \vspace{-0.15in}
\end{figure}

\textbf{Impact on Driving Behavior.} As we could not find production self-driving vehicles, we study the attack’s potential end-to-end impact on Apollo running in the LGSVL driving simulator. 
For simplicity, we replay the best attack results in the previous experiments as the color of the virtual traffic light in the simulator. 
We find the vehicle will start to brake or move in less than 0.5 seconds after recognizing a sudden change of the traffic light.
In some cases such as Fig.~\ref{fig:lgsvl}, such behaviors can make the vehicle run a red light and collide with the traffic, or suddenly decelerate and stop in the middle of the intersection, potentially causing rear-end collisions.




\section{Discussion}

\textbf{Universal Black-box Attack.}
An universal black-box attack is desirable as it does not require any knowledge of the target camera. Though our attack does not require a full white-box knowledge of the camera, knowing the refresh rate is necessary for attacking all cameras in order to stabilize the injected interference. As the refresh rate varies between devices, it is difficult to achieve an universal black-box attack that uses the same attack parameters for different target cameras.

\textbf{Timing Challenge.}
The timing of laser injection determines the stripe’s position in the image and directly affects the attack’s effectiveness. Unfortunately, it is impossible to predict the exact timing without direct feedback from the camera, which can lower the attack's effectiveness in practice. We consider the influence of timing as a probability problem.
A wider stripe will increase the probability of overlapping with the traffic light, however, it may also unnecessarily disrupt traffic light detection and the recognition of other objects in the image, such as cars and lanes, which may trigger alarms and disable self-driving. 
The precise effect of a wide stripe on object detection is unknown and shall be investigated in future work.
The attacker may also inject a stripe that slowly moves in the image across consecutive frames. Therefore, the attack may succeed during the time when the stripe overlaps with the traffic light, e.g., for a few seconds depending on the stripe’s moving speed, which is sufficient to affect self-driving according to our simulation.

\textbf{Countermeasures.}
To mitigate the threat, 
we suggest two methods to redesign the rolling shutter mechanism.  The  electronic shutter may (1) expose the CMOS rows in a random sequence, or (2) start the exposure from a random row for every frame, as illustrated in Fig.~\ref{fig:defense2}.
We simulate the first method, and Fig.~\ref{fig:defense-image} shows the images before and after applying random rolling shutter sequence. It shows that this method will scatter the injected stripe over the whole image, thus making the attack ineffective. 
The second method will make the injected stripe appear at different locations in the image across continuous frames. With this design, the attack may only succeed occasionally in discontinuous frames, and thus is difficult to affect the driving decision.




\section{Related Work}

A few studies have shown the insecurity of camera-based object detection and recognition systems.
Earlier studies~\cite{petit2015remote,yan2016can,truong2005preventing,lan2016live} showed that strong light could saturate the camera and disable the system.
To spoof the system, Man et al.~\cite{man2020ghostimage} managed to inject ghost images of a road sign into the camera using a regular projector, and Li et al.~\cite{li2019adversarial} injected adversarial image patterns by physically placing stickers on the camera's lens.
Other studies change the scene being filmed by physically modifying the object of interest~\cite{eykholt2018robust,zhao2019seeing,chernikova2019self} or projecting light to the scene\cite{nassi2020phantom,zhou2018invisible,nguyen2020adversarial,lovisotto2020slap} as adversarial patterns.
Our work differs from these papers in that we use laser to inject adjustable color stripes into the image as adversarial patterns, which could be conducted from a longer distance.
More recently, several works proposed to fool camera-based systems by attacking relevant non-camera sensors. For example, Ji et al.~\cite{ji2021poltergeist} used sound to attack a camera's image stabilization sensor, causing blurry images that deceive object detectors.
Tang et al.~\cite{tang2021fooling} proposed a region-of-interest attack that fools traffic light detection by fabricating localization errors.

Some studies have used laser to attack non-camera-based recognition systems, including microphones~\cite{sugawara2020light} and LIDAR~\cite{shin2017illusion,cao2019adversarial,petit2015remote}. 
These attacks exploit sensors' vulnerabilities and are known as transduction attacks~\cite{yan2020sok}. Other transduction attacks against vehicles have investigated ultrasonic sensors~\cite{xu2018analyzing}, rotational speed sensors~\cite{shoukry2013non}, and radars~\cite{yan2016can}.

Recently, two concurrent works~\cite{sayles2020invisible,kohler2021they} also exploited the rolling shutter to attack camera-based recognition systems. Like ours, they injected stripe-shaped interference on images. However, the goal of \cite{sayles2020invisible} is to cause targeted \textit{image misclassification} by flashing the LED illumination on the object in the scene, and \cite{kohler2021they} aimed to maximally \textit{disrupt object detection}, i.e., to hide objects in the image, by irradiating blue laser at the camera.
Both of them generated adversarial patterns for the entire image.
While in this paper, we focus on a completely different scenario, i.e., \textit{fooling traffic light recognition}, which requires different attack methods that inject adversarial patterns onto a specific part of the image where the traffic light is and considering aiming and synchronization in dynamic attack scenarios.


\section{Conclusion}

In this paper, we show the feasibility of fooling traffic light recognition with a laser.
We manage to create adjustable color stripes in the image that can alter the recognition results by exploiting the vulnerability of rolling shutters in CMOS cameras. By modeling the laser attack process, we are able to simulate the attack and search for effective laser parameters.
Our evaluation with real cameras shows the attack's effectiveness and demonstrates its potential real-world impact.
To alleviate the threat, we propose defenses for existing and future vehicles.

\section*{Acknowledgments}
We thank our shepherd Prof. Earlence Fernandes and the reviewers for their constructive feedback.
This work is supported by China NSFC Grant 61941120, 62071428, 61925109 and 2019 Zhejiang University Academic Award for Outstanding Doctoral Candidates.



\bibliographystyle{plain}
\bibliography{references/reference}

\appendix

\section*{Appendix}

\section{Laser Attack Modeling}

\subsection{Camera Modeling}
\label{sec:appendix_cameramodeling}
\textbf{Direct Interpolation.}
Suppose \textit{T} represents the tensor in RGB form, and \textit{M} represents the matrix in raw file. Then for pixel in the $i\; th$ row and $j\; th$ column in \textit{T}, mathematically,
\setlength{\abovedisplayskip}{3pt}
\setlength{\belowdisplayskip}{3pt}
\begin{align*}
    T[i,j,0] = \begin{cases}
    M[i,j] \quad \mbox{if}\; i \; mod \; 2 = 0\; \mbox{and}\; j \; mod \; 2 = 0 \\
    M[i-1,j] \quad \mbox{if}\; i \; mod \; 2 \neq 0\; \mbox{and}\; j \; mod \; 2 = 0 \\
    M[i,j-1] \quad \mbox{if}\; i \; mod \; 2 = 0 \; \mbox{and}\; j \; mod \; 2 \neq 0 \\
    M[i-1, j-1] \quad \mbox{else}
    \end{cases}
\end{align*}

\begin{align*}
    T[i,j,1] = \begin{cases}
    M[i,j+1] \quad \mbox{if}\; i \; mod \; 2 = 0\; \mbox{and}\; j \; mod \; 2 = 0 \\
    M[i,j] \quad \mbox{if}\; i \; mod \; 2 \neq 0\; \mbox{and}\; j \; mod \; 2 = 0 \\
    M[i,j] \quad \mbox{if}\; i \; mod \; 2 = 0 \; \mbox{and}\; j \; mod \; 2 \neq 0 \\
    M[i, j-1] \quad \mbox{else}
    \end{cases}
\end{align*}

\begin{align*}
    T[i,j,2] = \begin{cases}
    M[i+1,j+1] \quad \mbox{if}\; i \; mod \; 2 = 0\; \mbox{and}\; j \; mod \; 2 = 0 \\
    M[i,j+1] \quad \mbox{if}\; i \; mod \; 2 \neq 0\; \mbox{and}\; j \; mod \; 2 = 0 \\
    M[i+1,j] \quad \mbox{if}\; i \; mod \; 2 = 0 \; \mbox{and}\; j \; mod \; 2 \neq 0 \\
    M[i, j] \quad \mbox{else}
    \end{cases}
\end{align*}

\textbf{Bilinear Interpolation.}
For the edges of the image, bilinear interpolation takes the same procedure as direct interpolation. For the interior pixels, we use the following rules to implement interpolation for pixel in $i\; th$ row and $j\; th$ column:
\begin{equation*}
    T[i,j,0] =  \frac{\sum\limits_{k=-1}^1 \sum\limits_{l=-1}^1 M[i+k,j+l]\mathbf{1}\{\delta_{i+k}+\delta_{j+l}=0\}}{\sum\limits_{k=-1}^1 \sum\limits_{l=-1}^1 \mathbf{1}\{\delta_{i+k}+\delta_{j+l}=0\}}
\end{equation*}
Here, for simplicity, we define
$$ \delta_{i} = i \mod 2 $$
\begin{align*}
    \mathbf{1}\{A\} = \begin{cases}
    1 \quad \mbox{if A is true} \\
    0 \quad \mbox{else}
    \end{cases}
\end{align*}
Similarly, we can define for green and blue, respectively:
\begin{equation*}
    T[i,j,1] =  \frac{\sum\limits_{k=-1}^1 \sum\limits_{l=-1}^1 M[i+k,j+l]\mathbf{1}\{\delta_{i+k}+\delta_{j+l}=1\}}{\sum\limits_{k=-1}^1 \sum\limits_{l=-1}^1 \mathbf{1}\{\delta_{i+k}+\delta_{j+l}=1\}}
\end{equation*}

\begin{equation*}
    T[i,j,2] =  \frac{\sum\limits_{k=-1}^1 \sum\limits_{l=-1}^1 M[i+k,j+l]\mathbf{1}\{\delta_{i+k}+\delta_{j+l}=2\}}{\sum\limits_{k=-1}^1 \sum\limits_{l=-1}^1 \mathbf{1}\{\delta_{i+k}+\delta_{j+l}=2\}}
\end{equation*}
\par After interpolation, the raw file matrix is transformed into an tensor that could be used for image processing in RGB.

\subsection{Laser Effect Emulation}
\label{sec:appendix_lasereffect}

\begin{figure}[t]
    \centering
    \subfigure[Incidence from the left]{
        \centering
        \includegraphics[width=0.225\textwidth]{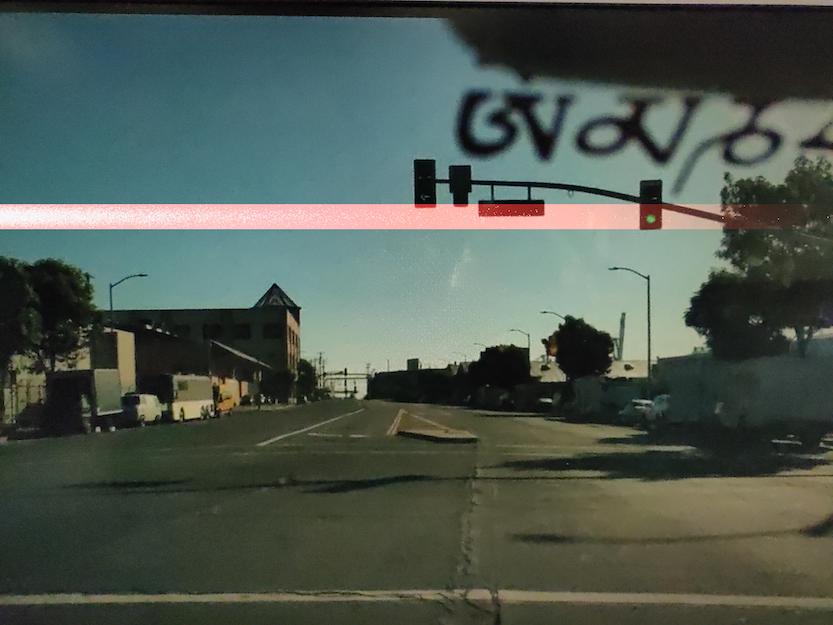}
    }
    \subfigure[Incidence from the front]{
        \centering
        \includegraphics[width=0.225\textwidth]{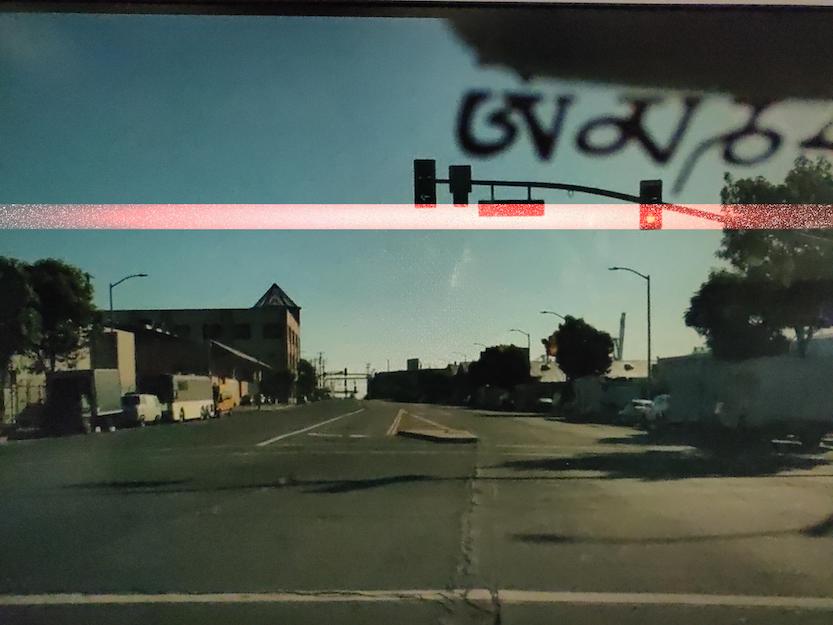}
    }
    \vspace{-0.2in}
    \caption{Illustration of the emulated color stripe with noise and irradiated from the left and front of the camera. (Sigmoid model, $I_{max}=1200, I_{min}=0, \beta_1 = 8, \beta_2 = 5$).}
    \vspace{-0.1in}
    \label{fig: Noise}
\end{figure}

\par \textbf{Linear Function: }Linear function is used when the incidence direction is from the left or the right. We made the following assumptions:
\begin{itemize}
    \item The effect of laser attack is uniform vertically: the light intensity difference only occurs among the x-axis;
    \item The light intensities at the two ends are $I_{max},I_{min}$: For example, if the incidence direction is from the left, the $I_{max}$ will occur at the left side of the image, while $I_{min}$ at the right;
    \item The light intensity changes linearly horizontally;
\end{itemize}
\par Therefore, given $I_{max},I_{min}$, we can express the light intensity function for any point $(x,y)$ on the color strip as (if the incidence direction is from the left):
\begin{equation*}
    I = I_{min} + \frac{y}{w}\left(I_{max}-I_{min}\right)
\end{equation*}
\par If the incidence direction is from the right, we only need to replace $y$ by $w-y$. The illustrations of the linear model are shown as following:
\begin{figure}[H]
    \centering
    \vspace{-0.2in}
    \subfigure[Incidence from the left]{
        \includegraphics[width=0.225\textwidth]{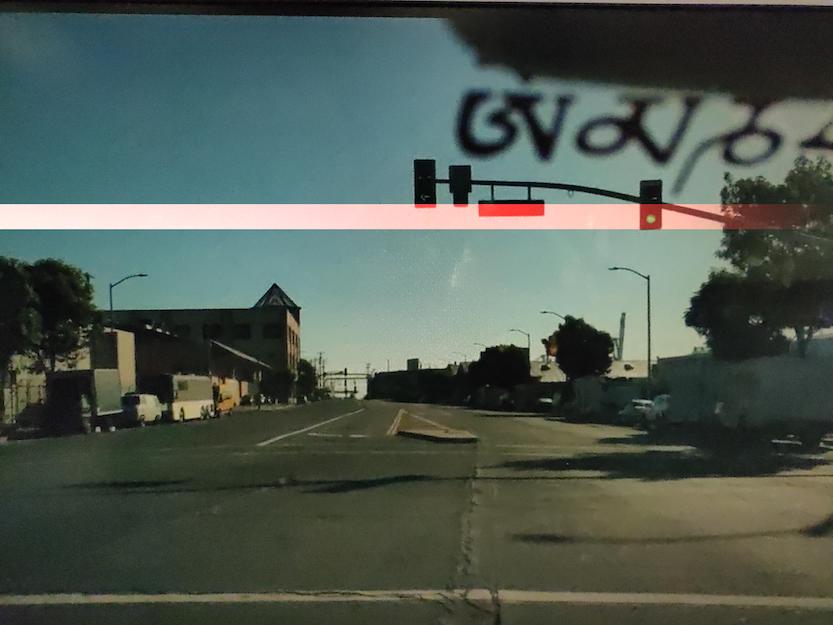}
    }
    \subfigure[Incidence from the right]{
        \includegraphics[width=0.225\textwidth]{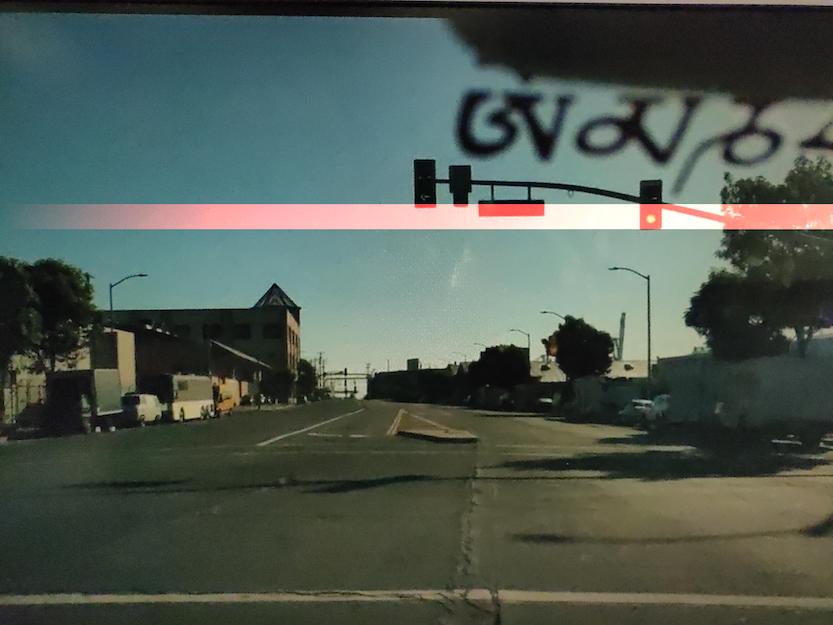}
    }
    \vspace{-0.2in}
    \caption{Illustration of the Linear Model for both directions with $I_{max}=2000,I_{min}=0$.}
    \vspace{-0.1in}
    \label{fig: Linear Model}
\end{figure}

\par \textbf{Sigmoid Function:} Another reasonable assumption is that the color intensity changes exponentially instead of linearly. Sigmoid Function is also used for the case when the incidence direction is from left or right. The assumption of the sigmoid function is the same as the linear case except for the last one: the light intensity changes according to a sigmoid-like function horizontally.
Therefore, given $I_{max},I_{min}$, we can express the light intensity function for any point $(x,y)$ on the color strip as (if the incidence direction is from the left):
\begin{equation*}
    I = I_{min} + \frac{1}{1+e^{-\alpha_1/(y-w/\alpha_2)}}\left(I_{max} - I_{min}\right)
\end{equation*}
\par Here, $\alpha_1,\alpha_2$ are hyper-parameters. If the incidence direction is from the right, we need to replace $y-\frac{w}{\alpha_2}$ by $\frac{w}{\alpha_2} - y$. The illustrations of the sigmoid model are shown as following:
\begin{figure}[H]
    \centering
    \vspace{-0.2in}
    \subfigure[Incidence from the left]{
        \centering
        \includegraphics[width=0.225\textwidth]{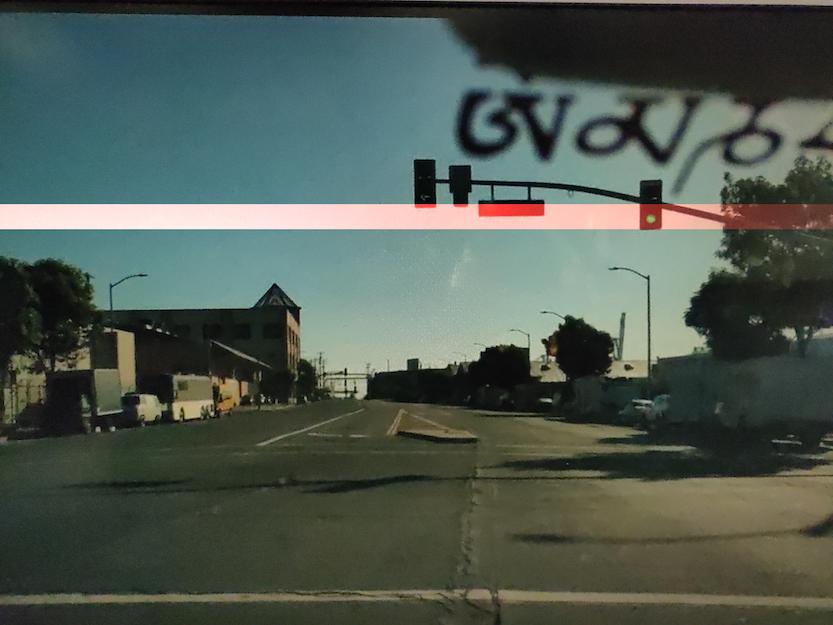}
    }
    \subfigure[Incidence from the right]{
        \centering
        \includegraphics[width=0.225\textwidth]{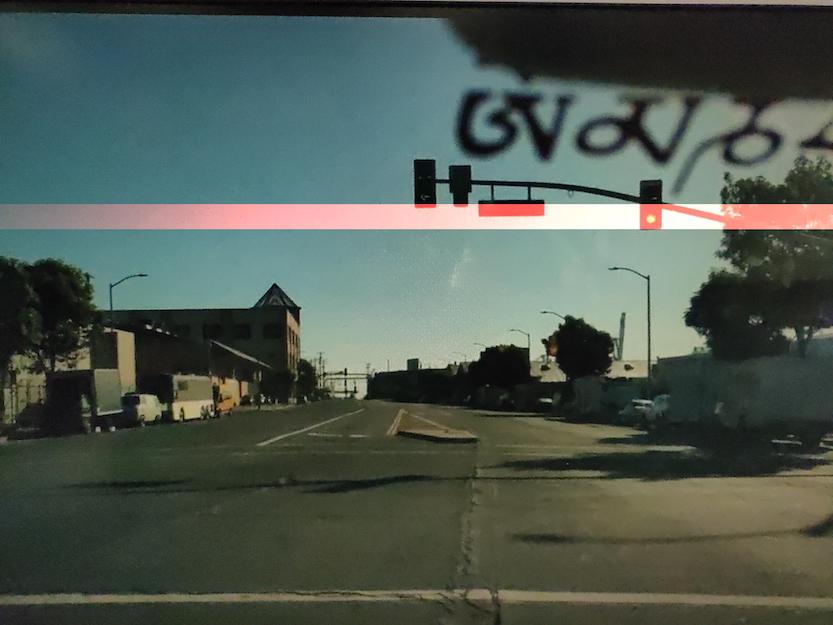}
    }
    \vspace{-0.2in}
    \caption{Illustration of the Sigmoid Model for both directions with $I_{max}=2000, I_{min}=0, \alpha_1 = 5, \alpha_2 = 2$.}
    \vspace{-0.1in}
    \label{fig: Sigmoid Model}
\end{figure}

\par \textbf{Gaussian Function:} If the incidence direction of the light is from the front, then the channel overflow is most likely to occur in the middle of the color strip. The assumptions of the Gaussian model is different:
\begin{itemize}
    \item The effect of laser attack follows two dimensional multi-normal distribution with the peak at the center of the stripe;
    \item The light intensity at the peak of the distribution is $I_{max}$ and converges to 0 at the periphery;
    \item The correlation between two directions is 0 ($\rho=0$, can be relaxed);
\end{itemize}
\par To implement the Gaussian model, we need the maximum light intensity $I_{max}$, and the top position of the color strip $x_0$. The light intensity function for any points $(x,y)$ on the color strip can be expressed as following:
\begin{equation*}
    a = \frac{\left(x-x_0-h/2\right)^2}{(h/\sigma_1)^2}
\end{equation*}
\begin{equation*}
    b = \frac{\left(y-w/2\right)^2}{(w/\sigma_2)^2}
\end{equation*}
\begin{equation*}
    c = \rho\frac{2\left(x-x_0-h/2\right)\left(y-w/2\right)}{(hw)/(\sigma_1 \sigma_2)}
\end{equation*}
\begin{equation*}
    I = \frac{1}{2\pi hw\sqrt{1-\rho^2}/(\sigma_2\sigma_2)} e^{-\frac{1}{2(1-\rho^2)}\left(a+b+c\right)} I_{max}hw
\end{equation*}
\par Here, $\sigma_1,\sigma_2$ are hyper-parameters representing the decaying rate of the light intensity from the center to the periphery. The illustration of the Gaussian model is shown as following:
\begin{figure}[H]
    \centering
    \vspace{-0.1in}
    \includegraphics[width=0.225\textwidth]{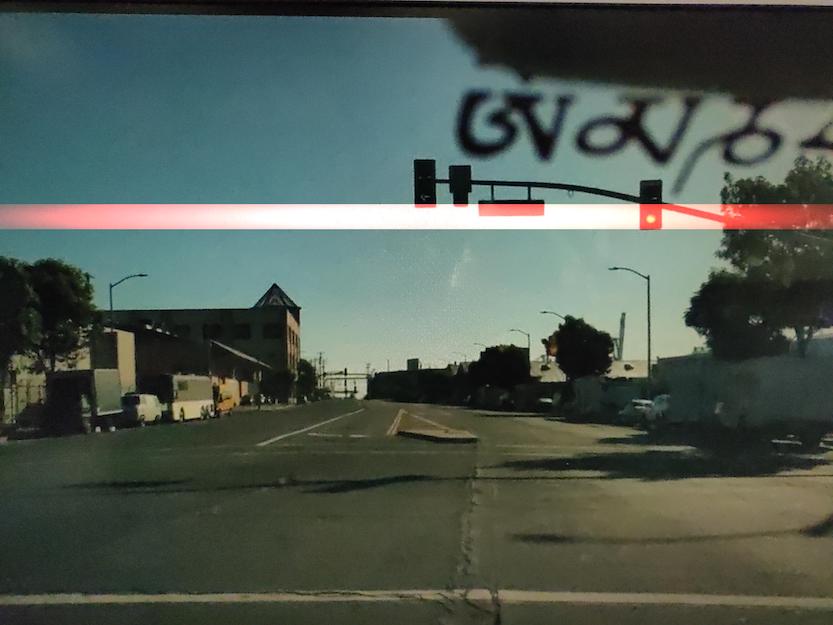}
    \vspace{-0.1in}
    \caption{Illustration of the Gaussian Model with $I_{max}=2000, \sigma_1=2, \sigma_2=4$.}
    \vspace{-0.1in}
    \label{fig:Gaussiam Model}
\end{figure}

\textbf{Adding Noise:}
The procedure is described in the following:
\begin{enumerate}
    \item Pick a random number $n$ from $n_1$ to $n_2$ to represent the number of noisy pixels;
    \item Randomly select $\{h_0,h_1,\cdots,h_{n-1}\}\times\{w_0,w_1,\cdots,w_{n-1}\}$, where $h_i,w_j \in \{1,2,3,4,5\}$ to represent the sizes of the noisy pixels;
    \item If the incidence direction is left/right, then according to distributions $v(x),h(y)$, generate tuples $\{(x_0,y_0),(x_1,y_1),\cdots,(x_{n-1},y_{n-1})\}$ to represent the upper left corner of the noisy pixels; If the incidence direction is from the front, then generate tuples $\{(x_0,y_0),(x_1,y_1),\cdots,(x_{n-1},y_{n-1})\}$ according to two-dimensional random distribution;
    \item Randomly select $\{b_0,b_1,\cdots,b_{n-1}\}$ to represent the brightness of the pixels;
    \item Then for $i=0,\cdots,n-1$, we replace the red channel value of all pixels in the square from $(x_i,y_i)$ (upper left corner)  to $(x_i+h_i,y_i+w_i)$ (lower right corner) by $b_i$.
\end{enumerate}

\subsection{Searching Laser Attack Parameters}
\label{sec:appendix_searchresults}

\begin{table}[h]
    \small
    \centering
    \vspace{-0.1in}
    \caption{Grids for parameter search.}
    \label{tab:Grids of Params}
    \begin{tabular}{ccc}
    \toprule
    Parameter & Symbol & Grids \\
    \midrule
    Min. Light Strength & $I_{min}$ & 0,400,800,1200,1600 \\
    Max. Light Strength & $I_{max}$ & 200,600,1000,1400,1800 \\
    Wavelength (Red) & $\lambda_{red}$ & 632,650,660 \\
    Wavelength (Green) & $\lambda_{green}$ & 505,520 \\
    Incidence Direction & None & Left, Right, Front \\
    Incidence Function & $D$ & Linear, Sigmoid, Gaussian \\
    \bottomrule
    \end{tabular}
    \vspace{-0.15in}
\end{table}

\subsection{Necessity of the Laser Attack Modeling}
\label{sec:appendix_necessity}

We conducted baseline experiments to evaluate the necessity of the
methodology design. We performed two groups of experiments, one with laser
parameters derived from the model-based search, and the other without (as the
naive baseline). In particular, the laser parameters in the naive baseline
experiments were picked randomly by a volunteer without knowledge of the
recommended parameters, e.g., he can use any laser strength as long as he
thinks it may fool traffic light recognition. Green-to-red attacks were conducted 20
times for each group using our previous setup (Fig.~\ref{fig: g2r TL position and distance}) at various traffic light
distances and angles. The model-guided group was successful 14/20 times,
rendering a success rate of 70\%, while the naive baseline group succeeded 6/20
times, rendering a success rate of 30\%. Thus, the comparison shows that the
methodology we designed can significantly increase the attack’s effectiveness.
\section{Evaluation and Defense}



\begin{figure}[h!]
    \centering
    \includegraphics[width=0.46\textwidth]{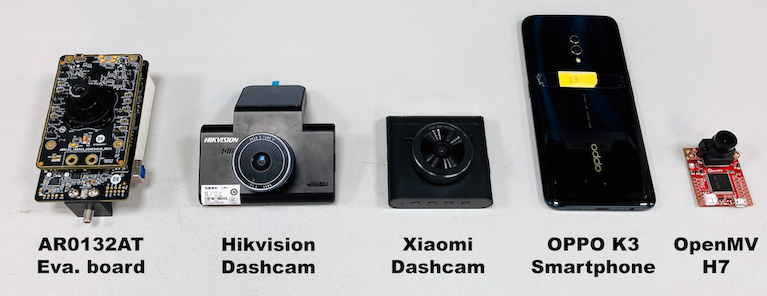}
    \vspace{-0.1in}
    \caption{The five cameras experimented in the evaluation.}
    \vspace{-0.25in}
    \label{fig: target devices}
\end{figure}

\begin{figure}[h!]
    \centering
    \vspace{-0.1in}
    \subfigure[G$\rightarrow$R]{
        \includegraphics[width=0.225\textwidth]{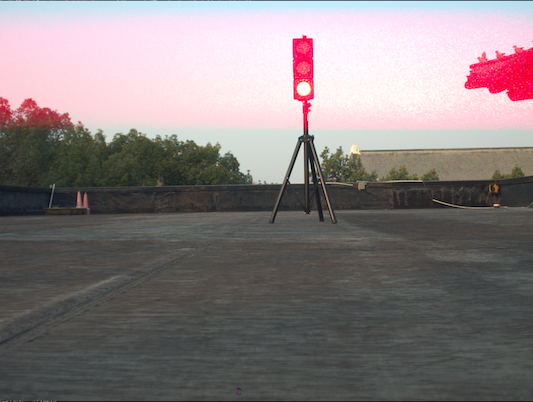}
    }
    \subfigure[R$\rightarrow$G]{
        \includegraphics[width=0.225\textwidth]{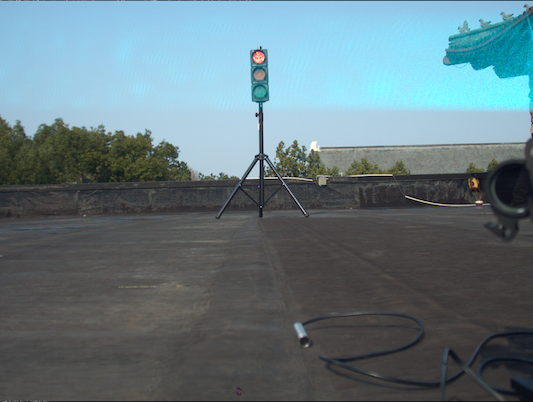}
    }
    \vspace{-0.2in}
    \caption{Images of attacking the AR0132AT eval. board.}
    \label{fig:camera1}
\end{figure}

\begin{figure}[h!]
     \centering
     \subfigure[G$\rightarrow$R]{
         \includegraphics[width=0.225\textwidth]{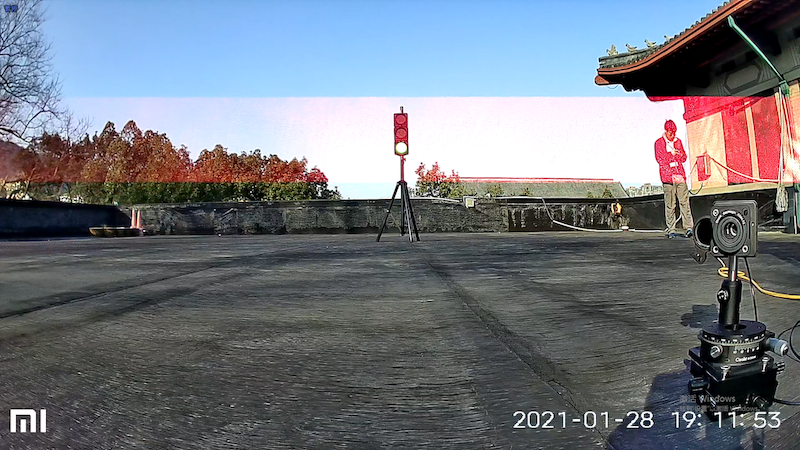}
     }
     \subfigure[R$\rightarrow$G]{
         \includegraphics[width=0.225\textwidth]{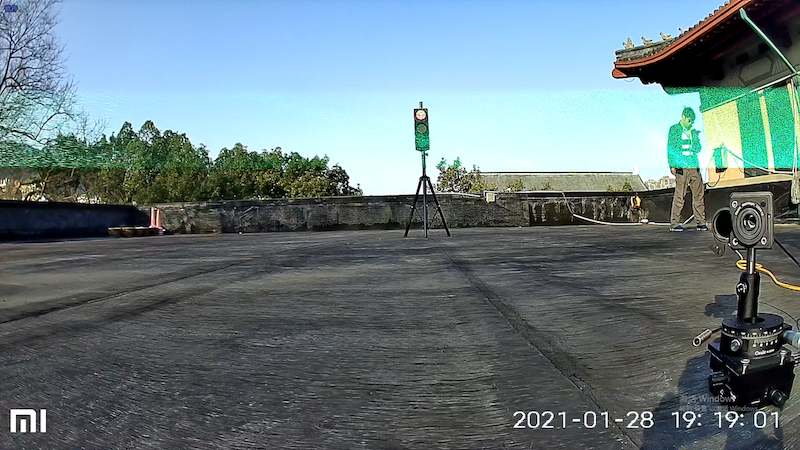}
     }
     \vspace{-0.2in}
     \caption{Images of attacking the Xiaomi Dashcam.}
     \label{fig:camera2}
     \vspace{-0.2in}
 \end{figure}

\begin{figure}[h!]
     \centering
     \subfigure[G$\rightarrow$R]{
         \includegraphics[width=0.225\textwidth]{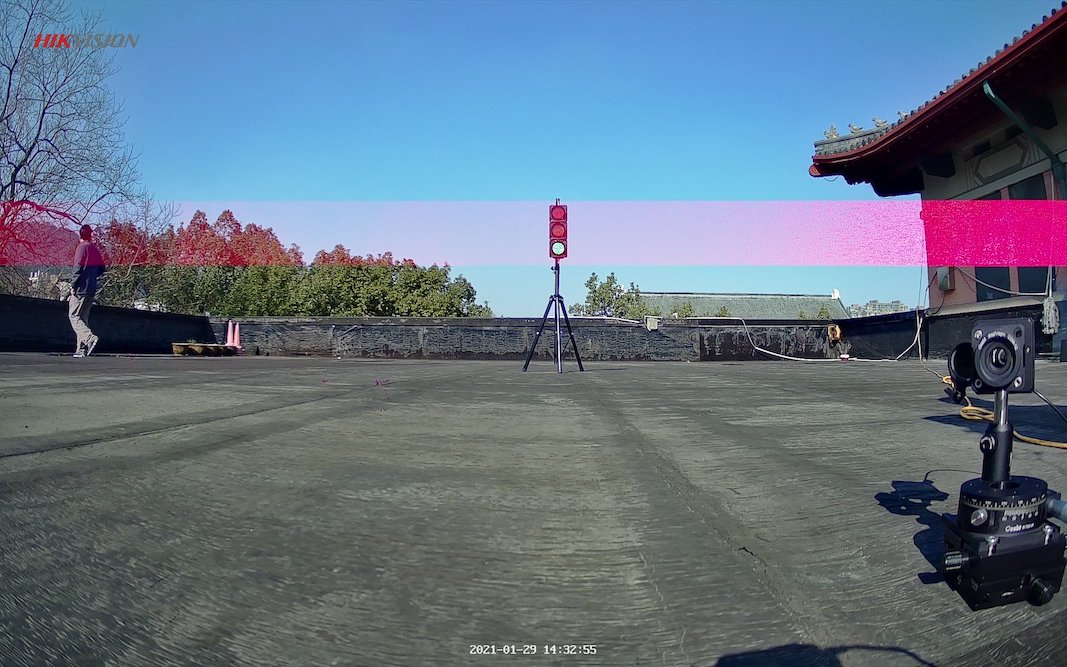}
     }
     \subfigure[R$\rightarrow$G]{
         \includegraphics[width=0.225\textwidth]{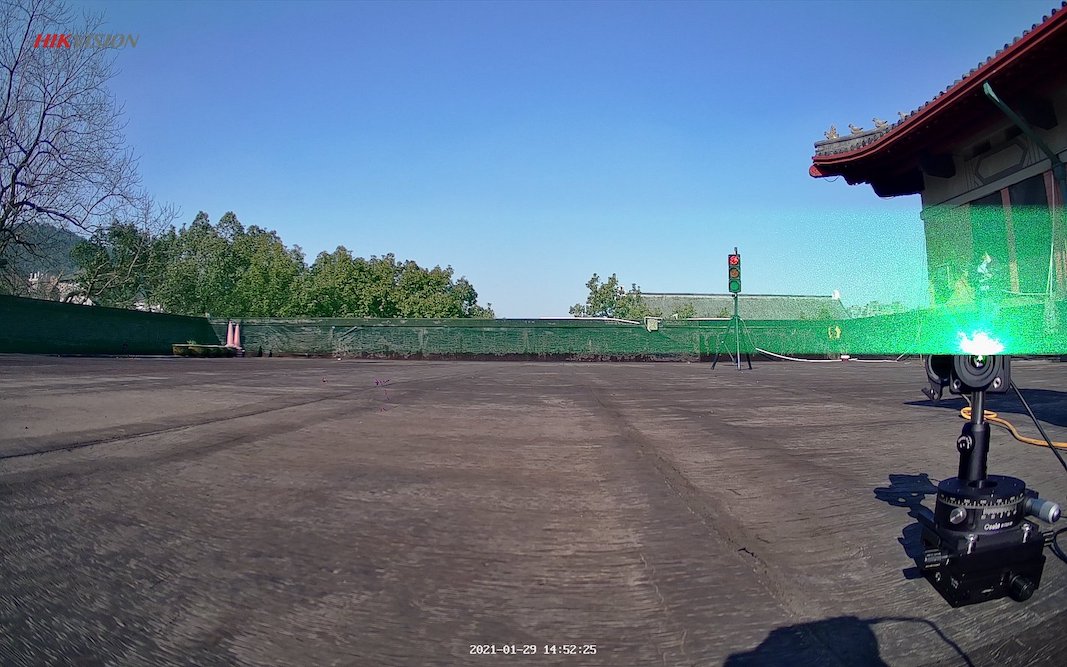}
     }
     \vspace{-0.2in}
     \caption{Images of attacking the Hikvision Dashcam.}
     \label{fig:camera3}
     \vspace{-0.2in}
 \end{figure}

\begin{figure}[h!]
     \centering
     \subfigure[G$\rightarrow$R]{
         \includegraphics[width=0.215\textwidth]{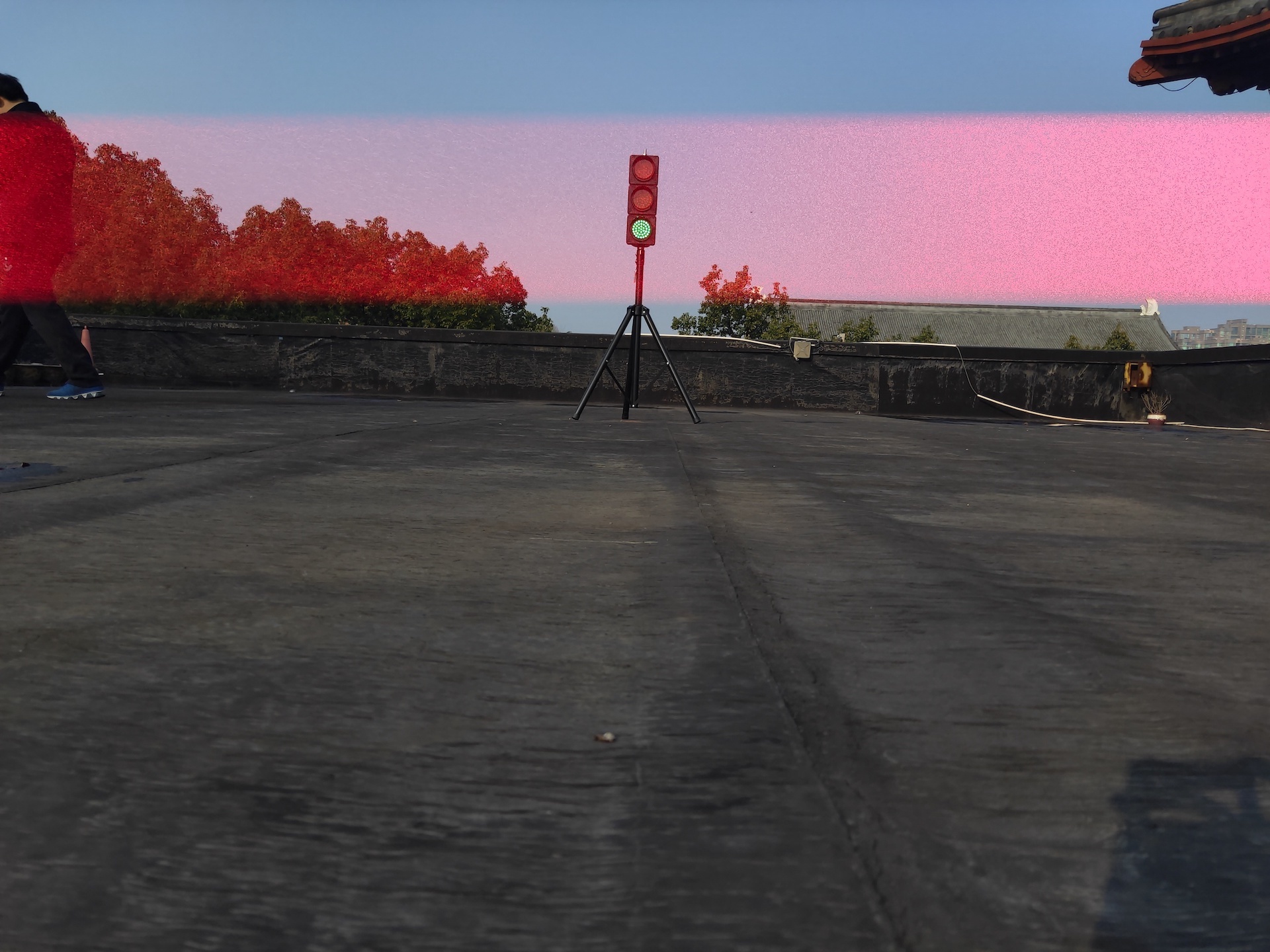}
     }
     \subfigure[R$\rightarrow$G]{
         \includegraphics[width=0.215\textwidth]{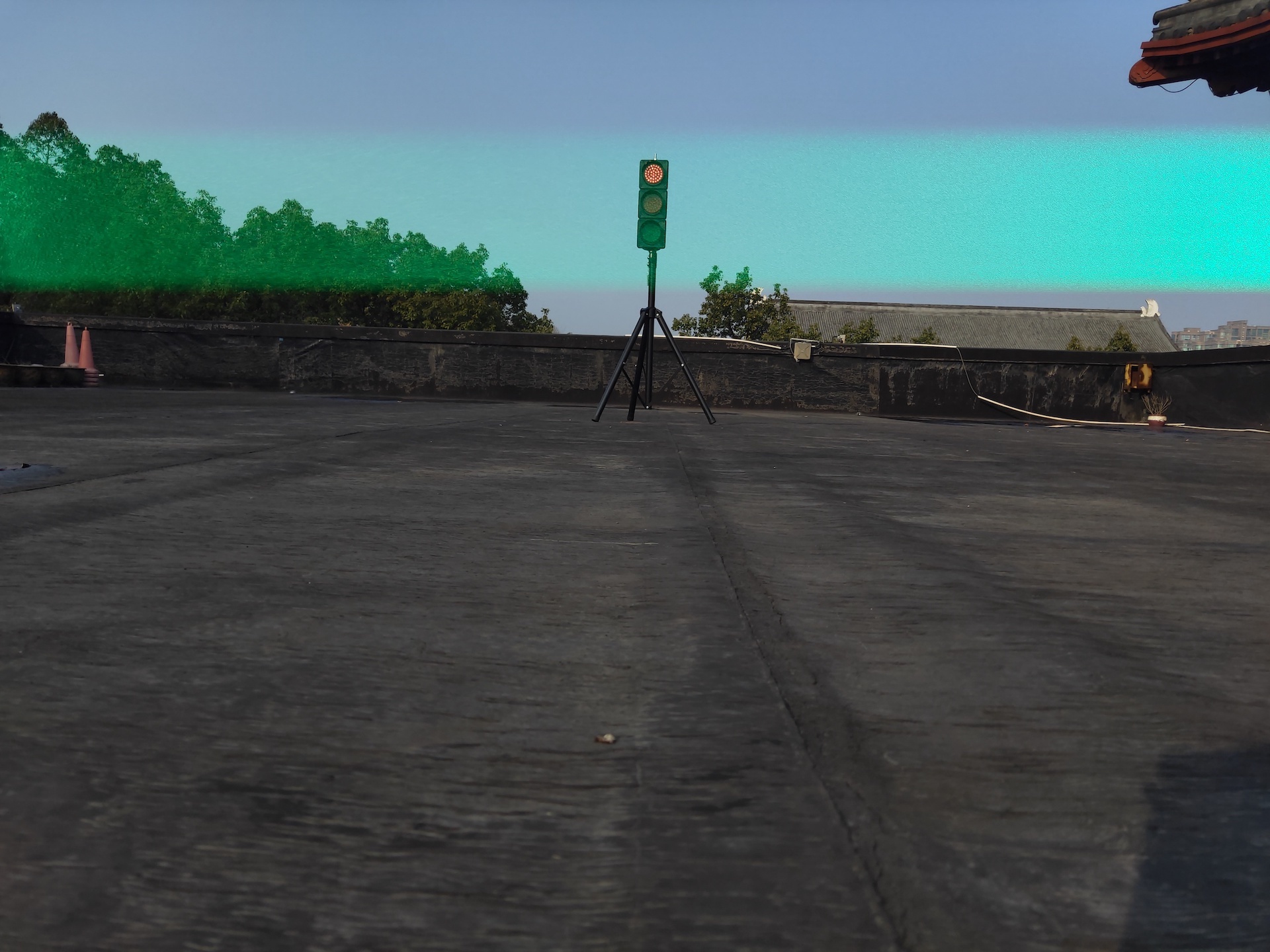}
     }
     \vspace{-0.15in}
     \caption{Images of attacking the OPPO K3 camera.}
     \label{fig:camera4}
     \vspace{-0.2in}
 \end{figure}

\begin{figure}[h!]
     \centering
     \subfigure[G$\rightarrow$R]{
         \includegraphics[width=0.215\textwidth]{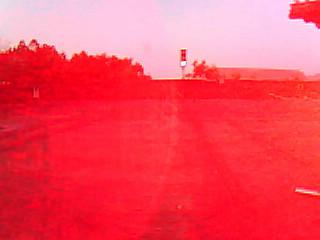}
     }
     \subfigure[R$\rightarrow$G]{
         \includegraphics[width=0.215\textwidth]{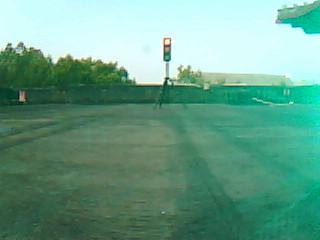}
     }
     \vspace{-0.15in}
     \caption{Images of attacking the OpenMV H7 devboard.}
     \label{fig:camera5}
 \end{figure}

The cameras we tested are shown in Fig.~\ref{fig: target devices} and their specs are listed in 
Table~\ref{tab: spec sheet}. The images they capture under attack are shown in Fig.~\ref{fig:camera1} to \ref{fig:camera5}.

\begin{table}[h!]
    \caption{Specs of the five experimented cameras.}
    \setlength\tabcolsep{2pt} 
    \small
    \begin{tabular}{@{}cccccc@{}}
    \toprule
    Parameter &
      Tesla &
      Hikv &
      Xiaomi &
      OPPO &
      OpMV \\ \midrule
    Sensor &
      AR0132 &
      N/A &
      \begin{tabular}[c]{@{}c@{}}OV\\ OS05A10\end{tabular} &
      IMX519 &
      OV7725 \\
    \begin{tabular}[c]{@{}c@{}}Refresh Rate (Hz)\end{tabular} &
      45.458 &
      20 &
      30 &
      30.012 &
      N/A \\
    Resolution &
      \begin{tabular}[c]{@{}c@{}}1280*\\ 964\end{tabular} &
      \begin{tabular}[c]{@{}c@{}}2560*\\ 1600\end{tabular} &
      \begin{tabular}[c]{@{}c@{}}1920*\\ 1080\end{tabular} &
      \begin{tabular}[c]{@{}c@{}}4608*\\ 3456\end{tabular} &
      \begin{tabular}[c]{@{}c@{}}320*\\ 240\end{tabular} \\
    Lens &
      \begin{tabular}[c]{@{}c@{}}Sunex\\ DSL945D\end{tabular} &
      N/A &
      N/A &
      N/A &
      N/A \\
    \begin{tabular}[c]{@{}c@{}}Field of View ($^{\circ}$)\end{tabular} &
      70 &
      130 &
      130 &
      N/A &
      70.8 \\
    F &
      2.5 &
      1.6 &
      1.8 &
      1.8 &
      2.0 \\ \bottomrule
    \end{tabular}
    \label{tab: spec sheet}
    \vspace{-0.1in}
    \end{table}

\begin{figure}[h!]
    \centering
    \vspace{-0.2in}
    \includegraphics[width=0.42\textwidth]{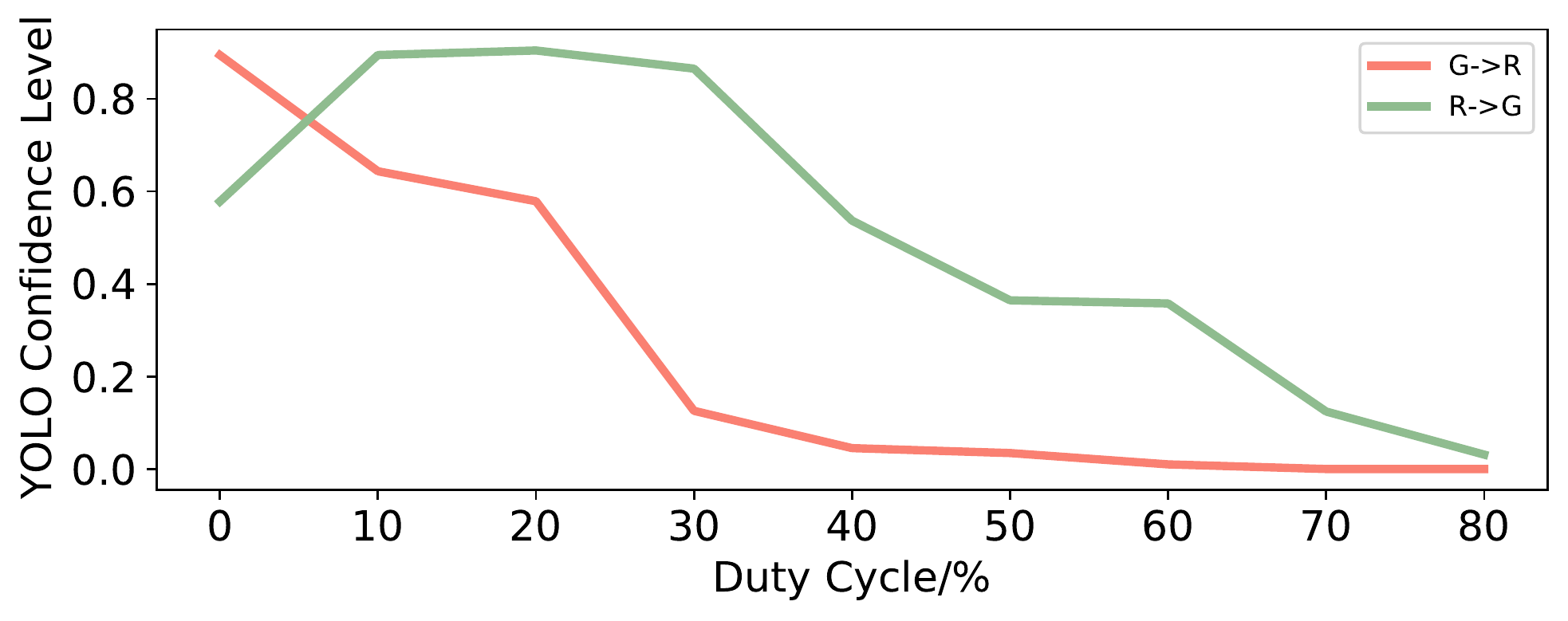}
    \vspace{-0.2in}
    \caption{The confidence score of traffic light detection while injecting stripes of various sizes. A larger duty cycle indicates a wider stripe. It shows that a wide stripe can significantly disrupt traffic light detection, and thus shall be avoided.}
    \vspace{-0.2in}
    \label{fig:stripe-width}
\end{figure}




\begin{figure}[t]
    \centering
    \subfigure[Random starting row]{
        \includegraphics[width=0.225\textwidth]{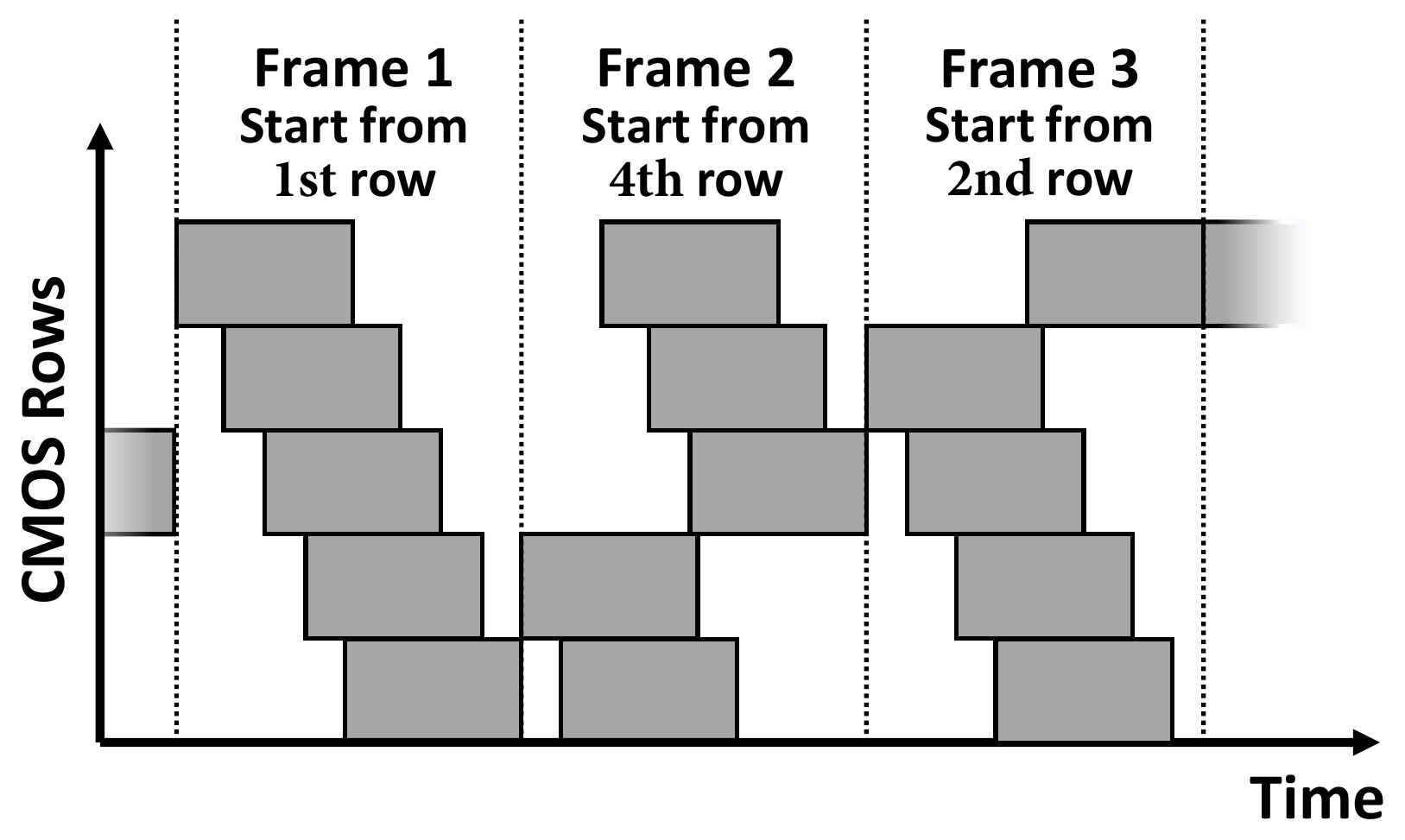}
    }
    \subfigure[Random rolling sequence]{
        \includegraphics[width=0.225\textwidth]{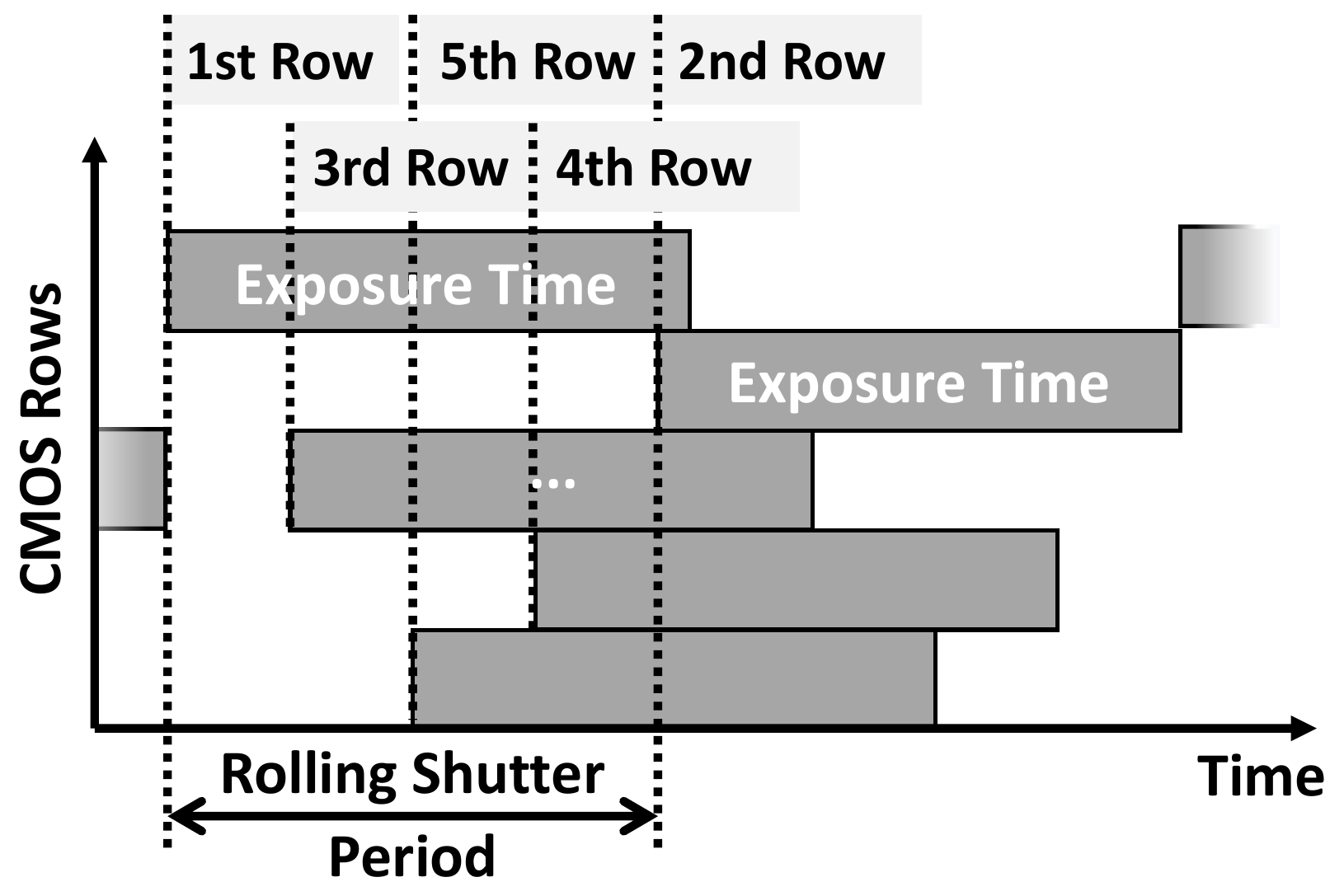}
    }
    \vspace{-0.2in}
    \caption{Random rolling shutter mechanisms that can destabilize or atomize the color stripe as a defense.}
    \label{fig:defense2}
    \vspace{-0.2in}
\end{figure}

\begin{figure}[h!]
    \centering
    \subfigure[Before defense]{
        \includegraphics[width=0.215\textwidth]{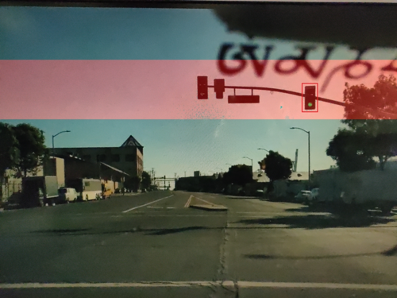}
        \vspace{-0.3in}
    }
    \subfigure[After defense]{
        \includegraphics[width=0.215\textwidth]{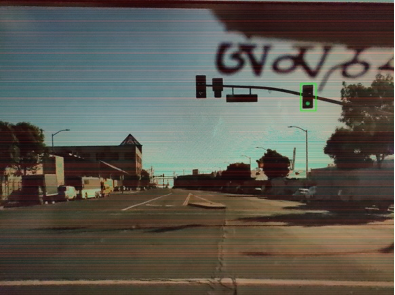}
    }
    \vspace{-0.15in}
    \caption{Images under attack before and after applying the random rolling sequence countermeasure.}
    \label{fig:defense-image}
    \vspace{-0.2in}
\end{figure}

\end{document}